%% file: main_paper.tex
\documentclass[lettersize,journal]{IEEEtran}
\usepackage{amsmath,amsfonts}
\usepackage{algorithmic}
\usepackage{algorithm}
\usepackage{array}
\usepackage[caption=false,font=normalsize,labelfont=sf,textfont=sf]{subfig}
\usepackage{textcomp}
\usepackage{stfloats}
\usepackage{url}
\usepackage{verbatim}
\usepackage{graphicx}
\usepackage{cite}
\usepackage{multirow}
\usepackage{multicol}
\usepackage{xcolor}
\usepackage{graphicx}
\usepackage{bbding}
\usepackage{pifont}
\usepackage{wasysym}
\usepackage{array}
\usepackage{colortbl}
\usepackage{booktabs}
\usepackage[capitalize]{cleveref}

\begin{document}

\title{Sparse-DeRF: Deblurred Neural Radiance Fields from Sparse View}

\author{
  Dogyoon~Lee,
  Donghyeong~Kim,
  Jungho~Lee,
  Minhyeok~Lee,
  Seunghoon~Lee,
  and~Sangyoun~Lee$^{\dagger}$,~\IEEEmembership{Member,~IEEE}
  \IEEEcompsocitemizethanks{
  \IEEEcompsocthanksitem D. Lee, D. Kim, J. Lee, M. Lee, S. Lee, and S. Lee are with the School of Electrical and Electronic Engineering, Yonsei University, Seoul 03722, Korea.
  Email: {nemotio, 2donghyung87, 2015142131, hydragon516, shlee423, syleee}@yonsei.ac.kr. 
  \IEEEcompsocthanksitem ${\dagger}$: corresponding author.
}}

%
%


\markboth{Journal of \LaTeX\ Class Files,~Vol.~14, No.~8, August~2021}%
{Shell \MakeLowercase{\textit{et al.}}: A Sample Article Using IEEEtran.cls for IEEE Journals}


\definecolor{third}{rgb}{1, 1, 0}
\definecolor{second}{rgb}{1, 0.5, 0}
\definecolor{best}{rgb}{1, 0, 0}

\maketitle

\begin{abstract}
Recent studies construct deblurred neural radiance fields~(DeRF) using dozens of blurry images, which are not practical scenarios if only a limited number of blurry images are available.
This paper focuses on constructing DeRF from sparse-view for more pragmatic real-world scenarios.
As observed in our experiments, establishing DeRF from sparse views proves to be a more challenging problem due to the inherent complexity arising from the simultaneous optimization of blur kernels and NeRF from sparse view.
Sparse-DeRF successfully regularizes the complicated joint optimization, presenting alleviated overfitting artifacts and enhanced quality on radiance fields.
The regularization consists of three key components:
Surface smoothness, helps the model accurately predict the scene structure utilizing unseen and additional hidden rays derived from the blur kernel based on statistical tendencies of real-world;
Modulated gradient scaling, helps the model adjust the amount of the backpropagated gradient according to the arrangements of scene objects;
Perceptual distillation improves the perceptual quality by overcoming the ill-posed multi-view inconsistency of image deblurring and distilling the pre-deblurred information, compensating for the lack of clean information in blurry images.
We demonstrate the effectiveness of the Sparse-DeRF with extensive quantitative and qualitative experimental results by training DeRF from 2-view, 4-view, and 6-view blurry images.
\end{abstract}


\input{sections/introduction}

\input{sections/related-work}

\input{sections/preliminary}

\input{sections/method}

\input{sections/experiments}

\input{sections/limitations_discussions}
\input{sections/conclusion}
\section*{Acknowledgments}
This work was supported by the National Research Foundation of Korea~(NRF) grant funded by the Korean government~(MSIT)~(RS-2024-00340745) and an Electronics and Telecommunications Research Institute~(ETRI) grant funded by the Korean government [24ZC1200, Research on hyper-realistic interaction technology for five senses and emotional experience]

%


\bibliographystyle{IEEEtran}
\bibliography{IEEEtran}

\begin{IEEEbiography}[{\includegraphics[width=1in,height=1.25in, clip,keepaspectratio]{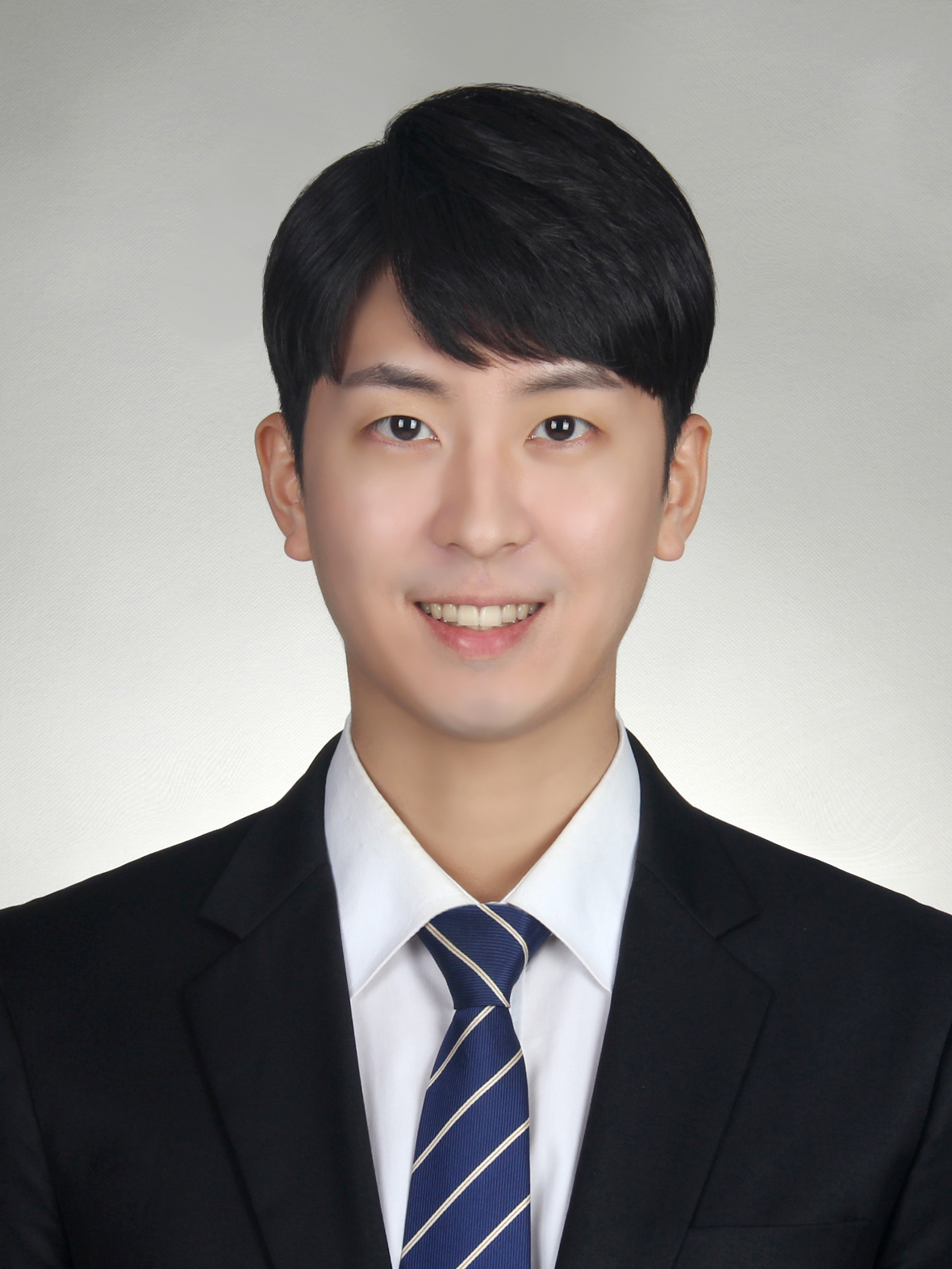}}]{Dogyoon Lee} is a Ph.D candidate at the School of Electrical and Electronic Engineering, Yonsei University.
He received his B.S. degree in Electrical and Electronic
Engineering from Yonsei University, Seoul, South Korea, in 2019. 
His current research interests focus on 3D computer vision including
Neural rendering and its applications in real-world scenarios, 3D from Images, 3D generative models, 3D reconstruction, and Image processing.
\end{IEEEbiography}

\begin{IEEEbiography}[{\includegraphics[width=1in,height=1.25in, clip,keepaspectratio]{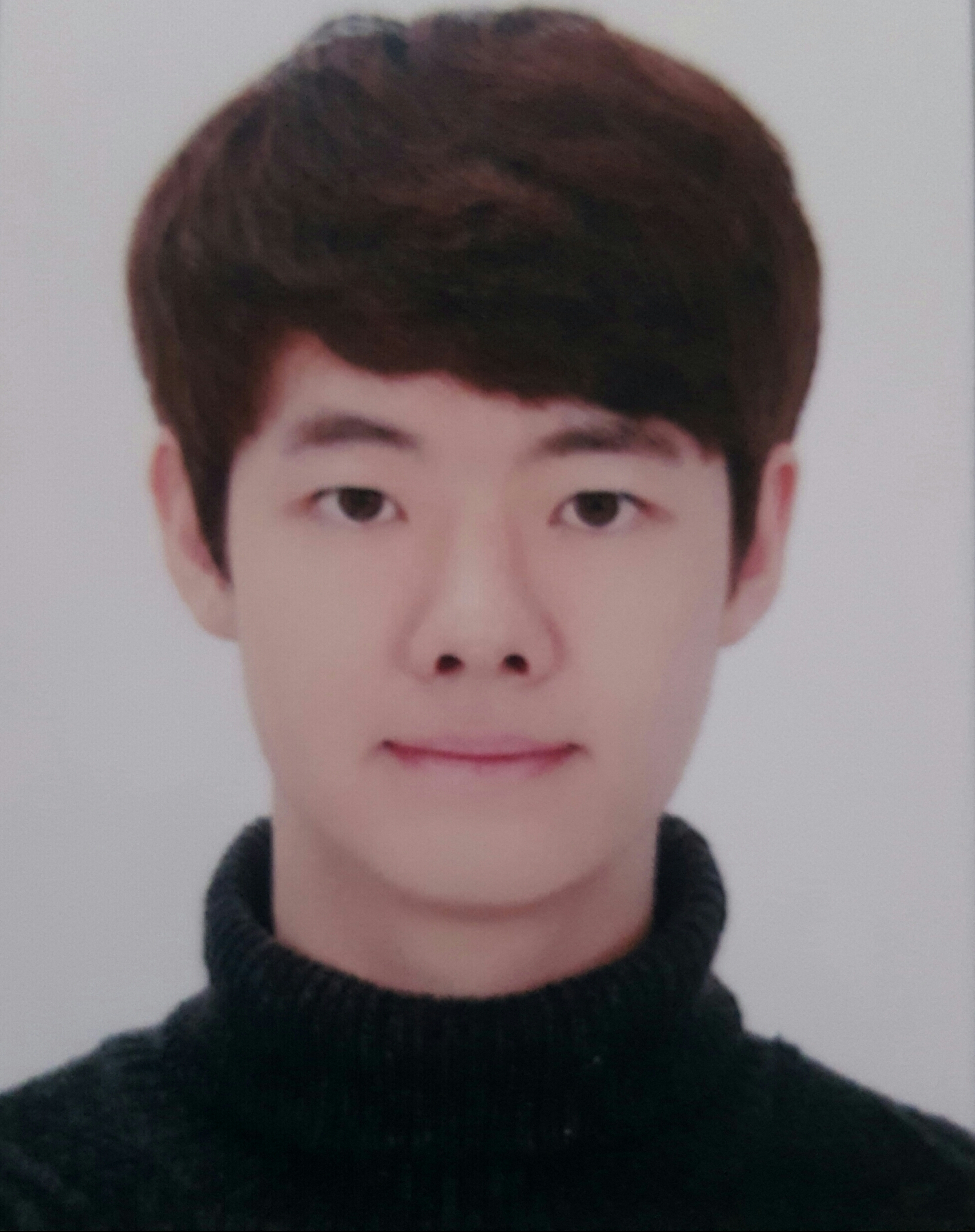}}]{Donghyeong Kim} is a Ph.D candidate at the School of Electrical and Electronic Engineering, Yonsei University. He received his B.S. degree in Electrical and Electronic
Engineering from Yonsei University, Seoul, South Korea, in 2021.  degree. His current research interests include
anomaly detection, 3D computer vision, generative models, 3D reconstruction, and Image processing.
\end{IEEEbiography}

\begin{IEEEbiography}[{\includegraphics[width=1in,height=1.25in, clip,keepaspectratio]{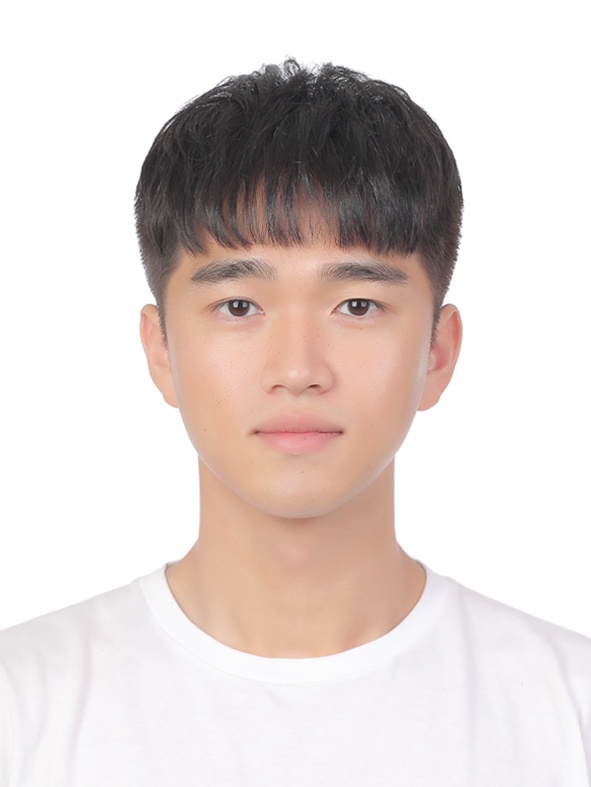}}]{Jungho Lee} is a Ph.D candidate at the School of Electrical and Electronic Engineering, Yonsei University. He received his B.S. degree in Electrical and Electronic Engineering from Yonsei University, Seoul, Korea, in 2021. His current research interests focus on neural rendering and human motion analysis in real-world conditions, with various mathematical machine learning tools such as neural ordinary differential equations.
\end{IEEEbiography}

\begin{IEEEbiography}[{\includegraphics[width=1in,height=1.25in, clip,keepaspectratio]{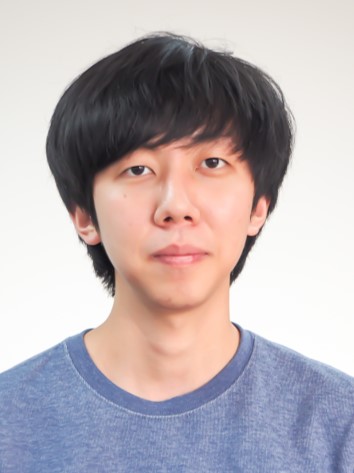}}]{Minhyeok Lee} is a dedicated Computer Vision and ML/DL Researcher with a focus on segmentation, autonomous driving, detection \& recognition, and novel view synthesis. Currently pursuing an Integrated M.S./Ph.D. in Electrical and Electronic Engineering at Yonsei University. His research spans various areas such as salient object detection, video object segmentation, camouflaged object detection, lane detection, and monocular depth estimation.
\end{IEEEbiography}

\begin{IEEEbiography}[{\includegraphics[width=1in,height=1.25in, clip,keepaspectratio]{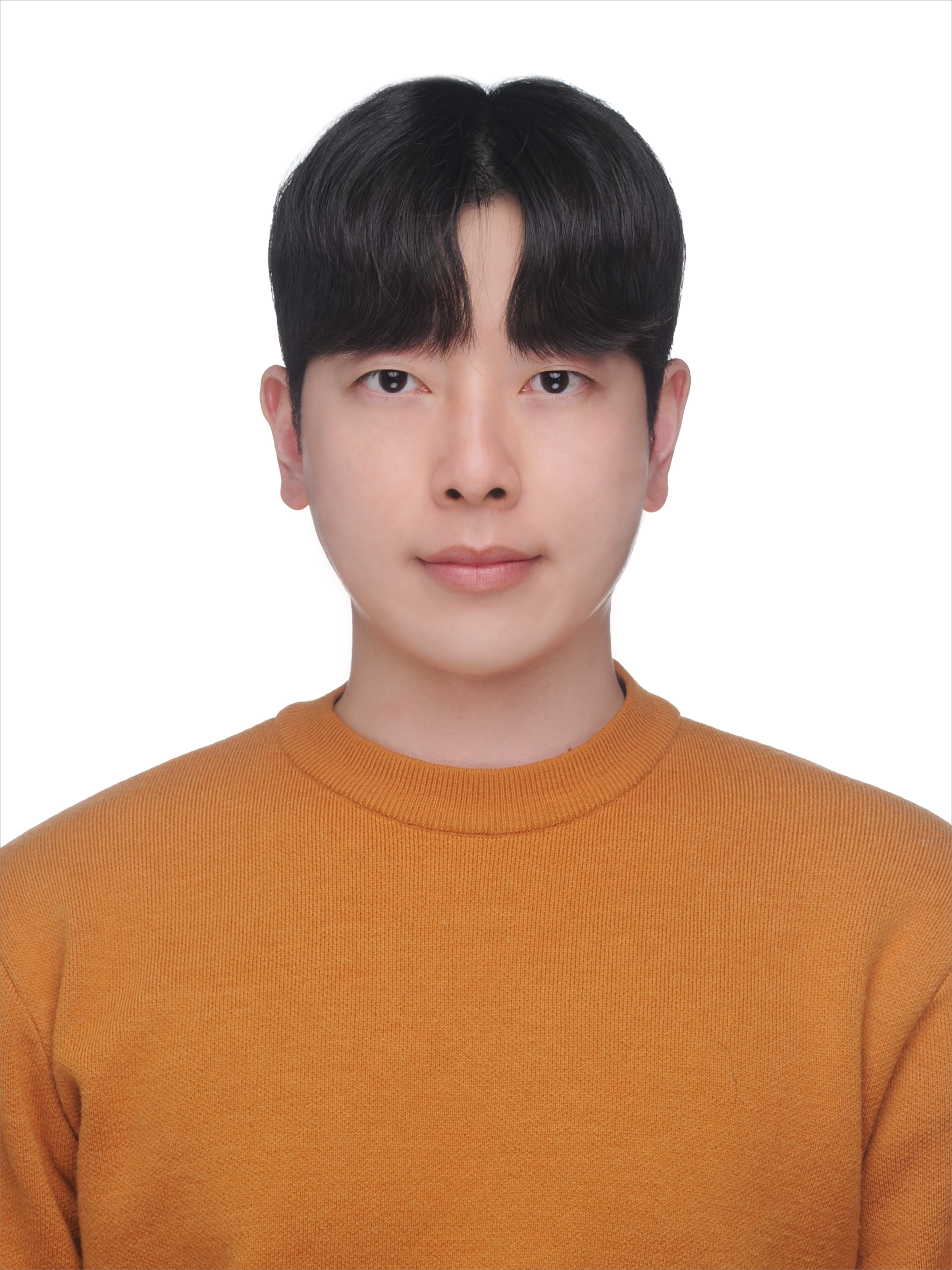}}]{Seunghoon Lee} received the B.S degree in Electronic Engineering of Inha University, Incheon, Korea in 2020.
He is currently an integrated MS/Ph.D degree student
in Electrical and Electronic Engineering, at Yonsei University. His research interests are video object segmentation, salient object detection, and super-resolution.
\end{IEEEbiography}

\begin{IEEEbiography}[{\includegraphics[width=1in,height=1.25in, clip,keepaspectratio]{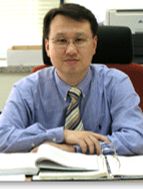}}]{Sangyoun Lee} received his Ph.D. degree in Electrical and Computer Engineering
from the Georgia Institute of Technology, Atlanta, Georgia, USA, in
1999. He is currently a professor at the School of Electrical and Electronic
Engineering. His research interests include all aspects of computer vision,
with a special focus on video codecs.
\end{IEEEbiography}

\clearpage
\newpage

{\appendices
\input{sections/appendix_sample_space}
\input{sections/appendix_inconsistency_of_estimated_depth}

\input{sections/appendix_additional_implementation_detail}

\input{sections/appendix_additional_results}

}

\vfill

\end{document}

%% file: sections/introduction.tex
\section{Introduction}
\label{sec:intro}
Representing 3-dimensional~(3D) space from multi-view images has rapidly grown after the emergence of the neural radiance fields (NeRF), which maps continuous spatial coordinates to volume density and radiance fields. 
Its realistic rendering quality and simple architecture have led to widespread applications and collaborations with various research fields in computer vision and graphics.
As practical applications of NeRF continue to attract attention, research in real-world scenarios has emerged as a promising research direction such as NeRF from noisy images or sparse view.

In real-world scenarios, tackling the blurry images from camera motion is regarded to be important since users often encounter degraded images when capturing photos with their own devices due to the unintentional camera movement during exposure time.
To solve this problem, several NeRF studies~\cite{ma2022deblurnerf, lee2023dpnerf, wang2023badnerf} have attempted to construct deblurred neural radiance fields~(hereafter, DeRF) from blurry images using joint optimization of internal implicit blur kernel and radiance fields, but they use dozens of blurry images to train, which is actually not practical scenarios.
The assumed experimental environments, where radiance fields are trained from about 20 to 30 blurry images, seem unlikely to occur in reality.
Hence we delve into the practical consideration for situations where only blurry images are utilized.
We reasoned that situations requiring the use of only blurry images would arise when the available images for reconstructing the desired 3D space are both very limited and blurry.
Following this rationale, we propose a novel pragmatic scenario for radiance fields from blurry images that establish the DeRF from sparse view settings.
Specifically, we set the 2, 4, and 6-view settings based on our consideration of the practical applications of research on generating radiance fields from blurry images.

Actually, the NeRF system already has an inherent drawback: it is prone to be overfitted to training views and struggles to grasp correct geometry when only sparse view inputs are available. 
Moreover, we experimentally find that blurred images lead to more severe overfitting in DeRF from sparse view because blur kernels introduce a more complex optimization process compared to standard NeRF.
Due to the increased complexity, DeRF training suffers more structural distortion than general NeRFs when trained from sparse view, exhibiting further overfitting with floating artifacts as shown in our experiments.
Although there are several works~\cite{yang2023freenerf, seo2023flipnerf, niemeyer2022regnerf} to regularize radiance fields in sparse view scenarios, existing regularization methods are not effective in addressing the complex optimization issue of DeRF as demonstrated in comparative experiments using existing representative regularization techniques in sparse view NeRF and blur kernel of the DeRF, namely RegNeRF~\cite{niemeyer2022regnerf} and DP-NeRF~\cite{lee2023dpnerf}.
Furthermore, in the DeRF system, it is challenging to use other data-deriven priors such as predicted depth supervision since available images do not ensure the confidence of the estimated values due to the inherent degradation of the given images.
The inherent degradations cause inconsistency across multiple views, which hinder accurate geometry prediction in NeRF, when it is used as additional prior. This tendency becomes more pronounced as the number of input images increases. We validated the tendency through experimental results.
Therefore, our goal is to regularize the complex joint optimization of blur kernel and radiance fields for DeRF to enhance the structural and perceptual quality of radiance fields from sparse blurry images, overcoming the aforementioned challenging issues without pretrained depth prior.

In this paper, we propose for the first time to ameliorate the spatial ambiguity and enhance the sharp texture of the DeRF from sparse view, which we refer to as Sparse-DeRF.
We introduce a novel regularization method for easing complex joint optimization, which consists of two geometric constraints and a perceptual prior.
Geometric constraints are proposed to predict the accurate structure in radiance fields from sparse view, which consists of surface smoothness~(SS) and modulated gradient scaling~(MGS).
First, SS rectifies the overall geometry based on classical depth smoothness on integrated unobserved rays as similar to RegNeRF~\cite{niemeyer2022regnerf}.
We utilize the novel hidden rays in camera motion cues derived from blur kernels as additional out-of-distribution unobserved rays to reflect the statistical flatness of real-world geometry as \cite{huang2000statistics, niemeyer2022regnerf} argued.
Second, MGS is designed to flexibly modulate the scaling function to compensate for the gradients based on the arrangement of the scene components, which cannot be handled by a single scaling function in non-parameterized coordinate systems such as normalized device coordinates~(NDC). 
It alleviates the spatial ambiguity arising from ray sampling and the disproportionate gradients of NeRF by introducing a parameterized sinusoidal function as a novel scaling function. 
These two geometric constraints improve the structural scene geometry of radiance fields even without explicit depth supervision in a sparse view setting.

In addition to geometric constraints, we introduce the perceptual distillation~(PD) as a perceptual prior to enhance the detailed texture of the radiance fields by taking advantage of the established image deblurring algorithm.
Traditional image deblurring has shown significant performance alongside the advancement of deep learning, demonstrating more enhanced details and textures.
We believe that the sharp texture information from such deblurred images can be used as complementary information to achieve high fidelity in the Sparse-DeRF environment, where only a few degraded images are available to reconstruct the scene.
However, while we can take the pre-deblurred images with a pre-trained deep learning-based image deblurring model, the independence of image deblurring poses challenges in directly utilizing deblurred images as pixel-wise color supervision, due to inconsistency across the given images.
This inconsistency comes from the inherent ill-posed property of the image deblurring that breaks the geometric and appearance consistency across the multi-view images of the single 3D scene.
Hence, we impart the perceptual information of pre-deblurred images to the radiance fields by distilling the features extracted from the deep learning-based image feature extractor.
Extracted features enable the radiance fields to enhance perceptual quality by utilizing pre-deblurred textures.

Our results illustrate that the Sparse-DeRF produces high-quality rendered images from sparse blurry images, with improved perceptual texture quality and well-structured scene geometry.
Additionally, we demonstrate the effectiveness of the proposed constraints and a prior through experimental results and analysis.
Furthermore, we conduct comprehensive experiments to investigate ablations using two types of representative blur kernels from Deblur-NeRF~\cite{ma2022deblurnerf} and DP-NeRF~\cite{lee2023dpnerf}.
These experiments aim to show the superiority of the proposed regularization method and analyze its effects depending on the type of kernel employed.

%% file: sections/related-work.tex
\section{Related Work}
\label{sec:re_work}

\subsection{Neural Rendering and Radiance Fields}
\label{subsec:neral_rendering_rework}
Traditionally, researchers have been required to know the physical properties of a scene to simulate the rendering process for generating photorealistic images from 3D space. 
While rendering simulations facilitated the synthesis of controllable high-quality images across the 3D scene, the quality of the synthesized image significantly depends on the physical properties involved in the rendering process.
For real-world scenes, estimation of the properties which is referred to as "inverse rendering" is required, but it is difficult to predict them accurately solely depending on 2D observations like images and videos.
Although several approaches have been attempted to overcome the challenges, "neural rendering" has recently emerged as a superior approach integrating deep learning methodologies and graphics rendering approaches, leveraging the outstanding representation capability of deep neural networks.

According to a comprehensive survey~\cite{tewari2020state}, which well summarizes the history of early neural rendering, this research area has been regarded as the intersection of generative adversarial networks~(GANs)~\cite{goodfellow2014generative} and graphical controllable image synthesis.
With the adoption of GANs, neural rendering has been considered as an image-to-image translation problem utilizing given scene parameters and several 3D scene representations, leveraging the insights of conditional GANs similar to Pix2Pix~\cite{isola2017pix2pix}.
For example, \cite{eslami2018neural, meka2019deep, sun2019single} generate high-quality images with particular scene conditions by transferring scene parameters to the deep neural network.
In addition, other works incorporate the intuition of classical graphics modules into GANs to synthesize and control the image outputs utilizing non-differentiable or differentiable modules such as usage of rendered images with dense input conditioning~\cite{kim2018deepvideoportraits, eslami2018neural, liu2019neuralrerendering}, computer graphics renderer~\cite{lombardi2019neuralvolumes, sitzmann2019srn}, and illumination model~\cite{shu2017neuralfaceediting}.

Although these researches present realistic neural rendering techniques over the past few years, there has been great transition in paradigm in neural rendering after emergence of the neural radiance fields~(NeRF)~\cite{mildenhall2021nerf}, which directly map 3D spatial location and viewing direction to irradiance solely relying on multi-view images through multi-layer perceptron~(MLP) and classical volume rendering method~\cite{kajiya1984ray}.
NeRF implicitly represents the 3D scene with the classical ray tracing methods and shows photorealistic novel view synthesis, but there is still room for improvement in various aspects.
NeRF has widely spread to other computer vision and graphics tasks thanks to its simple and intuitive architecture, which attracts huge attention and expands the research fields of neural rendering.
To enhance the performance of neural representation itself, several works have represented 3D scenes using another representation to improve training or rendering speed, such as voxel-grid~\cite{liu2020nsvf}, plenoctree~\cite{yu2021plenoctrees}, decomposed tensorial fields~\cite{chen2022tensorf}, hashgrid~\cite{muller2022instantngp}, plenoxels~\cite{fridovich2022plenoxels}, light fields~\cite{attal2022neurallightfield}, and 3D gaussians~\cite{kerbl202333dgaussiansplatting}.
In addition, its implicit representation capability leads to explosive development of other graphical tasks such as modeling dynamic scenes~\cite{li2021nsff, park2021nerfies, pumarola2021dnerf, li2022n3dv}, relighting~\cite{bi2020neuralreflectance, srinivasan2021nerv}, 3D reconstructions~\cite{wang2021neus, rudnev2022nerfosr}, and human avatar~\cite{peng2021animatablenerf}.

\subsection{Radiance Fields from Degraded Images}
\label{subsec:radiance_degarded_images}
Establishing NeRF from degraded images is recently emerging since the ideal training condition in images for NeRF often breaks in real-world scenarios.
RawNeRF~\cite{mildenhall2022rawnerf} denoises the internal noise of the camera sensor to construct high dynamic range~(HDR) radiance fields from dark raw images and controls the camera exposure.
Similarly, NaN~\cite{pearl2022nan} deals with burst noise in images, generating denoised images based on IBRNet~\cite{wang2021ibrnet}, which is another image-based rendering approach.
For more practical applications of NeRF in real-world, DeblurNeRF~\cite{ma2022deblurnerf} firstly attempts to deal with two types of blur degradation in images, blur from camera motion and defocus, constructing deblurred neural radiance fields~(DeRF) from only blurry images.
They imitate the blurring process integrating the concept of blind deblurring in image deblurring with the NeRF system, modeling the blur kernel as pixel-wise independent ray transformation and composition weights to approximate the blurring process.
Another representative approach is DP-NeRF~\cite{lee2023dpnerf}, which is a key prior work that provided strong motivation for our current research and serves as the main baseline in our experiments.
DP-NeRF~\cite{lee2023dpnerf} imposes physical consistency across the images by modeling the blur kernel as the 3D rigid transformation of rays depending on each view, to approximate the actual blurring process in the camera more precisely.
Several approaches~\cite{lee2023exblurf,peng2023pdrf, wang2023badnerf} are also proposed, attempting to improve the quality of the constructed DeRF with simple motion interpolation and progressive blur kernel estimation.
With the emergence of the Gaussian splatting~\cite{kerbl20233d}, researches such as~\cite{zhao2025badgaussian,lee2025deblurringgs,peng2025bags} have also been published in this field, demonstrating that research remains active and ongoing.

One of the most actively researched areas among those mentioned earlier is NeRF from blurry images, which often occurs when users take pictures with their own devices.
However, as we mentioned in Section~\ref{sec:intro}, the experimental setup of using only 20$\sim$30 blurry images, as in previous studies, is not practical.
If we assume a scenario where users only have access to blurry images, it is more realistic to consider that only a few images are available for a specific scene and all of those images are blurry.
Therefore, we propose a more practical scenario by combining DeRF and the sparse view setting, thereby enhancing real-world applicability.

\subsection{Radiance Fields from Sparse View}
\label{subsec:radiance_practical_rework}
There have been a lot of works to apply the neural representation in more pragmatic scenarios as the importance of VR and AR technologies increased, such as fast rendering, efficient sampling on rays, scene editing, denoising, and training from sparse view.
Fast rendering, efficient sampling on rays, and scene editing aim to increase the inference speed, enable surface sampling, and deform the trained mesh through various approaches, such as baking~\cite{hedman2021baking}, depth-guided sampling~\cite{neff2021donerf}, and surface deformation~\cite{yuan2022nerfediting}, respectively.

Another dominant area is constructing the NeRF from sparse view images, which is a practical environment considering real-world scenarios.
Sparse view images incur the inherent drawback of neural networks in that the network is more likely to be overfitted to the given data distribution.
This leads to lack of the information for dense scene geometry in the mapped representations, typically manifested as incorrectly predicted structure, such as elongated density artifacts in the rendered color and depth images from novel views.
Several approaches have mitigated this issue involving additional prior knowledge or out-of-distribution data.
InfoNeRF~\cite{kim2022infonerf} adopts entropy minimization to probability density function~(PDF) of density value along the ray density to make the shape of the PDF sharper.
RegNeRF~\cite{niemeyer2022regnerf} utilizes the statistical depth smoothness of real-world geometry~\cite{huang2000statistics} on unobserved ray patches to reduce the artifact.
Recently, FlipNeRF~\cite{seo2023flipnerf} considers flipped rays on the surface as supplement unseen rays to regularize the scene geometry.
In other approaches, some works, such as PixelNeRF~\cite{yu2021pixelnerf}, and DietNeRF~\cite{jain2021dietnerf}, exploit the semantic information extracted from deep image feature extractors to utilize the representative power of neural networks in feature level.
FreeNeRF~\cite{yang2023freenerf} tries to alleviate the overfitting problem based on an optimization perspective, imposing some restrictions on the frequency level.

There have been various studies attempting to address the challenge of accurate geometry prediction in sparse-view settings by utilizing depth priors obtained from pretrained depth estimation networks or point clouds generated by structure-from-motion (SfM).
For example, SparseNeRF~\cite{wang2023sparsenerf} tackled this issue by distilling depth ranking information, while SPARF~\cite{truong2023sparf} incorporated geometric warping to refine the depth priors for unseen views. 
DSNeRF~\cite{deng2022dsnerf} leveraged point clouds obtained from SfM as depth priors and Dense-Depth-Prior~\cite{roessle2022densedepthpriornerf} further employed depth completion techniques to reduce the sparsity of the SfM point cloud, resulting in denser depth priors. 
StructNeRF~\cite{chen2023structnerf} incorporate superpixel segmentation to refine the depth priors even more effectively.
While these approaches attempt to address the lack of information by using depth priors, our proposed practical experimental settings that use the sparse blurry images reveal significant limitations. 
As mentioned earlier, the presence of blur in each image makes predicted depth inaccurate and the inconsistent across the multiple views, which hinders their usability as a prior.

%% file: sections/preliminary.tex
\section{preliminary}
\label{sec:prelim}
\subsection{Deblurred Neural Radiance Fields}
\label{subsec:DeRF_prelim}
Neural radiance fields~(NeRF) is parameterized MLPs for mapping continuous 3D location $\textbf{x}=(x,y,z)$ to volumetric density $\sigma$ and view-dependent radiance color $\textbf{c}=(r,g,b)$.
It is formulated as an approximated universal function $\textbf{F}_{\Theta}: (\gamma_{\textbf{x}}(\textbf{x}), \gamma_{\textbf{d}}(\textbf{d})) \rightarrow (\textbf{c}, \sigma)$, where $\Theta$ and $\textbf{d}=(\phi, \theta)$ denote the parameters of the NeRF MLPs and viewing direction of ray, respectively.
The function $\gamma$ is a positional encoding function that maps each input $\textbf{x}$ and $\textbf{d}$ to a high dimensional encoded feature, which is generally defined as a concatenation of frequency-adjusted sinusoidal function as Eq.~\ref{eq:pos_enc_func}.
\begin{equation}
	\label{eq:pos_enc_func}
	\gamma(\textbf{x}) = [\textbf{x}, \sin(\textbf{x}), \cos(\textbf{x}), ..., \sin(2^{f}\pi\textbf{x}), \cos(2^{f}\pi\textbf{x})],
\end{equation}
where $f=\{0, ..., m-1\}$ denotes frequency band with maximum frequency value $m$.
Hereafter, we abbreviate the encoding function and represent the function of the NeRF as
\begin{equation}
	\label{eq:NeRF_forward}
	F_{\Theta}(\textbf{x}, \textbf{d}) \rightarrow (\textbf{c}, \sigma).
\end{equation}
NeRF is trained with pixel-wise color supervision from multi-view input images to optimize the MLPs by predicting each pixel color $\hat{C}$ based on volumetric rendering~\cite{max1995opticalvolumerendering} with the samples along the generated ray $\textbf{r}$ from paired camera parameters.
For given ray origin $\textbf{o}$ and viewing direction $\textbf{d}$ along a pixel $p$, the samples along the ray $\textbf{r}$ are evenly divided to $N$ intervals to generate coarse samples with stratified sampling.
The samples are defined as $\textbf{r}_{i}=\textbf{o} + t_i\textbf{d}$ in near-to-far bounded partitions $[t_n, t_f]$ as shown in Eq.\ref{eq:nerf_sampling}, where $i$ indicates i-th sample and $t$ denotes the distance from ray origin.
\begin{equation} 
	\label{eq:nerf_sampling}
	t_{i}\sim\mathcal{U}\left[t_{n}+\frac{i-1}{N}\left(t_{f}-t_{n}\right),t_{n}+\frac{i}{N}\left(t_{f}-t_{n}\right)\right].
\end{equation}
Following the~\cite{max1995opticalvolumerendering}, the coarse pixel color~$\hat{C}_c$ is rendered from estimated color $\textbf{c}_i$ and density $\sigma_i$ of each sample $\textbf{r}_i$ as
\begin{equation} 
	\label{eq:vol_ren}
	\hat{C}(\textbf{r})=\sum_{i=1}^N w_{i}\textbf{c}_{i}=\sum_{i=1}^N T_{i}\left(1-exp(-\sigma_{i}\delta_{i})\right)\textbf{c}_{i},
\end{equation}
where $T_{i}=exp(-\sum_{j=1}^{i-1}\sigma_{j}\delta_{j})$ and $\delta_{i}=t_{i+1}-t_{i}$ indicate transmittance of each sample along the ray and distance between adjacent samples, respectively.
Hierarchical volume sampling is conducted again utilizing normalized weights as probability density function~(PDF) from $w_i$ as $\hat{w}_i=w_i/\sum{w_j}^{N_c}_{j=1}$ and fine rendered pixel color $\hat{C}_f$ is produced through above rendering process again.
Coarse- and fine-rendered color is supervised from the true pixel colors from input images through L2-norm as
\begin{equation} 
	\label{eq:nerf_reconloss}
	\mathcal{L}_{recon} = \sum_{\textbf{r}\in\mathcal{R}}\left[ \lVert \hat{C}_{c}(\textbf{r}) - C(\textbf{r}) \rVert^{2}_{2} + \lVert \hat{C}_{f}(\textbf{r}) - C(\textbf{r}) \rVert^{2}_{2} \right],
\end{equation}
where $\mathcal{R}$ is the set of rays in each batch and C(r) is ground truth RGB colors for ray $\textbf{r}$.

However, the above loss can not be applied to train the DeRF, since there is no true pixel color for training the NeRF in the DeRF environments.
To solve this problem and construct DeRF, \cite{ma2022deblurnerf, lee2023dpnerf} build additional MLPs for predicting the blur kernel in front of the NeRF to imitate the traditional blind blurring process, which is shown as Eq.\ref{eq:blurring_process}.
\begin{equation} 
	\label{eq:blurring_process}
	\hat{B}_{p}=\hat{C}_{p}*h_{p},
\end{equation}
where $p$, $\hat{B}$, $*$, and $h$ indicate the target pixel, expected blurred color, convolution operator, and blur kernel, respectively.
Hereafter, we abbreviate the $p$ for clarity.
The expected blurred color $\hat{B}$ is composited from $n$ rendered pixel colors $\hat{C}_{q}$ induced from modeled rays that approximate the blurring process as Eq.\ref{eq:blurring_composition}.
\begin{equation} 
	\label{eq:blurring_composition}
	\hat{B}=\sum_{q\in \mathcal{B}(p)}m_{q}\hat{C}_{q},\indent where~\mathcal{B}(p)=\{1,\dots,n\},
\end{equation}
where $m$ and $\mathcal{B}$ denote composition weights and the set of indices of the approximated blurring rays with respect to pixel $p$, respectively.
Note that, the number of $\mathcal{B}(p)$ is $n$, which is a hyper-parameter that decides the approximation quality of discrete transformation for blur process.
Finally, DeRF is trained with the color reconstruction loss on blurred colors as 
\begin{equation} 
	\label{eq:derf_reconloss}
	\mathcal{L}^{B}_{recon} = \sum_{\textbf{r}\in\mathcal{R}}\left[ \lVert \hat{B}_{c}(\textbf{r}) - B(\textbf{r}) \rVert^{2}_{2} + \lVert \hat{B}_{f}(\textbf{r}) - B(\textbf{r}) \rVert^{2}_{2} \right],
\end{equation}
where $\hat{B}_{c}$, $\hat{B}_{f}$, and $B$ are expected coarse, fine, and ground truth blurred color of the ray $\textbf{r}$, respectively.
The blur kernels are representatively modeled as a different type of transformation in each paper,~\cite{ma2022deblurnerf,lee2023dpnerf}.

\begin{figure}[h]
  \centering
  \includegraphics[width=\linewidth]{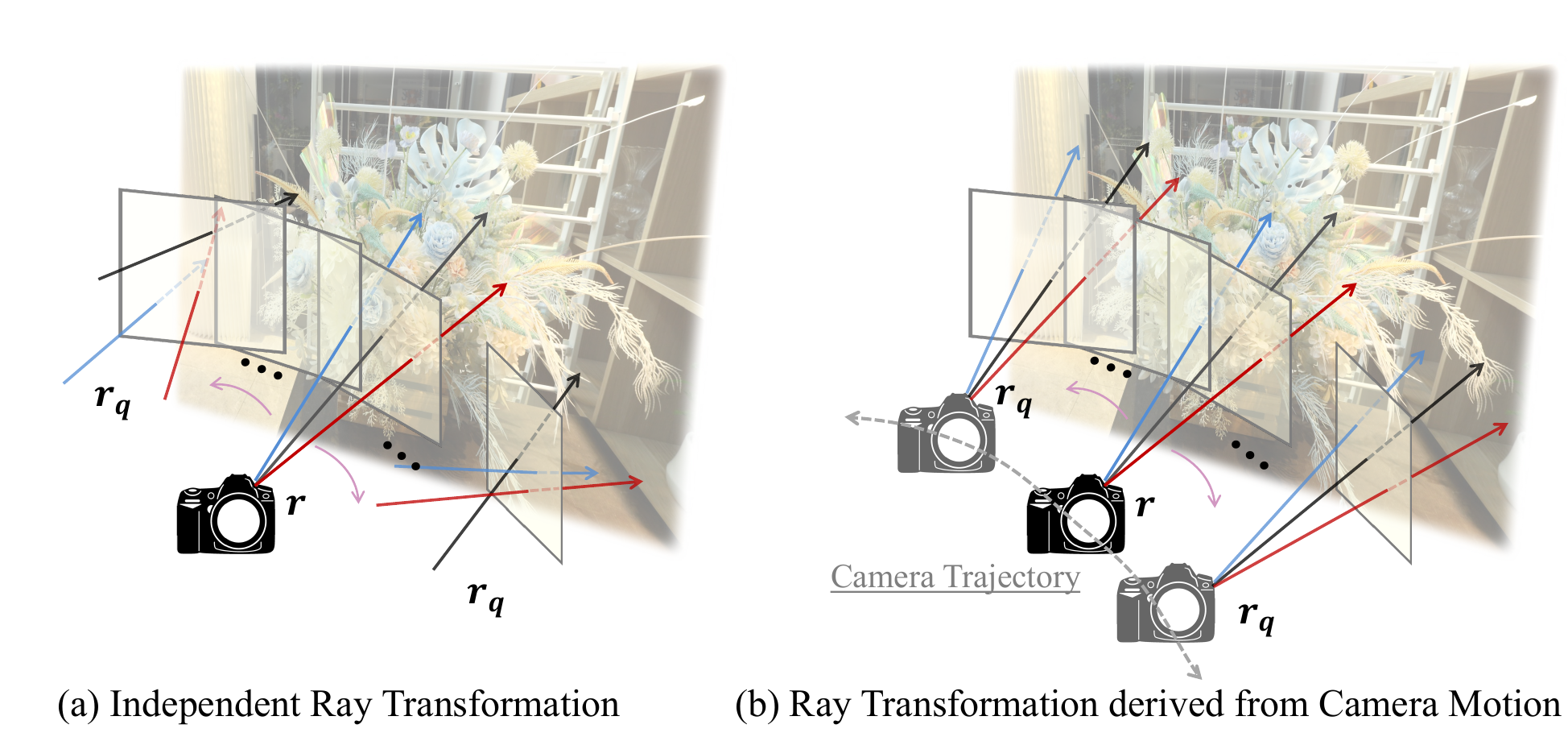}
  \vspace{-0.5cm}
  \caption{Simple illustration for different blur kernel modeling of (a)~Deblur-NeRF~\cite{ma2022deblurnerf} and (b)~DP-NeRF~\cite{lee2023dpnerf}. The main difference between the two kernels is the consistency between transformed rays induced from the blur kernel.}
  \label{fig:blur_kernel_modeling_difference}
\end{figure}

\subsection{Blur Kernels in Deblurred Neural Radiance Fields}
\label{subsec:blur_kernel_prelim}
The core difference of the blur kernels between the two papers~\cite{ma2022deblurnerf,lee2023dpnerf} is a consideration of 3D consistency across the entire pixels in each image as we illustrate in \figurename~\ref{fig:blur_kernel_modeling_difference}.
Deblur-NeRF~\cite{ma2022deblurnerf} designs the blur kernel as a deformable sparse kernel~(DSK), which consists of the $n$ number of transformations depending on embedded latent features from each view and location of image pixels.
The transformation is formulated as the 3D vector of ray origin and 2D vector of pixel coordinate, which is initialized within $N\times N$ window on the image plane as
\begin{equation} 
	\label{eq:blur_kernel_deblurnerf}
	(\Delta \textbf{v}_{o}, \Delta \textbf{v}_{T}, m)_q = G_{\Phi}(\chi, \textbf{l}_s),\indent where~q \in \{1,...,n\}.
\end{equation}
For $G_{\Phi}$, MLP with parameter $\Phi$, inputs are $\chi$ and $\textbf{l}_s$, which are latent embedded pixel coordinates and scene information, respectively. 
Here, $\chi$ is defined as randomly initialized canonical coordinates within a specific small range and $s$ indicates the specific scene.
The 3D vector $\Delta \textbf{v}_{\textbf{o}}$ and $\Delta \textbf{v}_{T}$ transform given ray $\textbf{r}=\textbf{o} + t\textbf{d}$ to generate transformed rays imitating blurring process as Eq.\ref{eq:blur_kernel_transform_deblurnerf}.
\begin{equation} 
	\label{eq:blur_kernel_transform_deblurnerf}
	\textbf{r}_{q} = (\textbf{o}+\Delta \textbf{v}_{\textbf{o};q})+t\textbf{d}_{q},
\end{equation}
where $\textbf{d}_{q}$ is transformed ray direction by applying $\Delta \textbf{v}_{T;q}$ to pixel coordinates to move the endpoint of the ray.
Then blurred color $\hat{B}$ of the target pixel is composited by weighted summation of each rendered color $\hat{C}_{q}$ and $m_{q}$ as Eq.\ref{eq:blurring_composition}.

However, DP-NeRF~\cite{lee2023dpnerf} argues that the pixel-wise independent optimization of the blur kernel in \cite{ma2022deblurnerf} incurs inconsistency in 3D geometry and appearance.
They utilize the physical intuition of actual blurred image acquisition in the camera process as an additional prior for the DeRF, to impose the constraints for constructing radiance fields while preserving 3D consistency.
To ensure 3D consistency, unlike Deblur-NeRF~\cite{ma2022deblurnerf}, they made the blur process shared within a single image and defined it as a learnable set of camera motions.
To directly model the actual camera motion as a 3D rigid transformation, they introduce scene-wise $SE(3)$ fields inspired by \cite{park2021nerfies}, \cite{lynch2017modernrobotics} and Rodrigues' formula~\cite{rodrigues1816}.
Scene-wise rigid transformation of the camera is formulated by estimated screw axis $S_s\in \mathbb{R}^6$ through MLPs depending on only scene information as Eq.\ref{eq:mlp_dpnerf}.
\begin{equation} 
	\label{eq:mlp_dpnerf}
	(S_{s;q}, m_s)=(r_s,v_s,m_s)_q=T_{\Psi}(\textbf{l}_s),\indent where~q\in\{1,...,n\},
\end{equation}
where $T_{\Psi}$ and $\textbf{l}_s$ denote MLP with parameter $\Psi$ and latent embedded scene information, respectively.
The predicted $r_q$ and $v_q$ of $S_q$ are converted to rotation matrix $e^{r_q}$ and translation matrix $\textbf{p}_q$ by formulas of  \cite{rodrigues1816} and \cite{lynch2017modernrobotics}, respectively.
Note that, we abbreviate specific scene indicator $s$ for clarity.
Hence, transformed rays are formulated as the rigid transformation of the rays as Eq.\ref{eq:blur_kernel_dpnerf}.
\begin{equation} 
	\label{eq:blur_kernel_dpnerf}
	\textbf{r}_q = e^{S_q}\textbf{r}=e^{r_q}\textbf{r} + \textbf{p}_q.
\end{equation}
The blurred color $\hat{B}$ of the target pixel is also composited by weighted summation of each rendered color $\hat{C}_{q}$ and $m_{q}$ as same as Eq.\ref{eq:blurring_composition}.
In addition to modeling the blurring process with rigid blur kernel~(RBK), \cite{lee2023dpnerf} proposes an adaptive weight proposal network~(AWP) based on the internal correlation between transformed rays and motion axis to predict the adaptive composition weights $\tilde{m}_q$ for each pixel, which complements the effect on the blur derived from the depth difference.

In this paper, we propose a novel approach that consists of geometric constraints and a perceptual prior through extensive experiments with these two types of blur kernels from Deblur-NeRF~\cite{ma2022deblurnerf} and~DP-NeRF~\cite{lee2023dpnerf}.
The conducted experiments demonstrate the effectiveness of Sparse-DeRF on both kernels.

%% file: sections/method.tex
\section{Sparse-DeRF}
\label{sec:sparse_derf_overview}
\begin{figure*}[h]
  \centering
  \includegraphics[width=\linewidth]{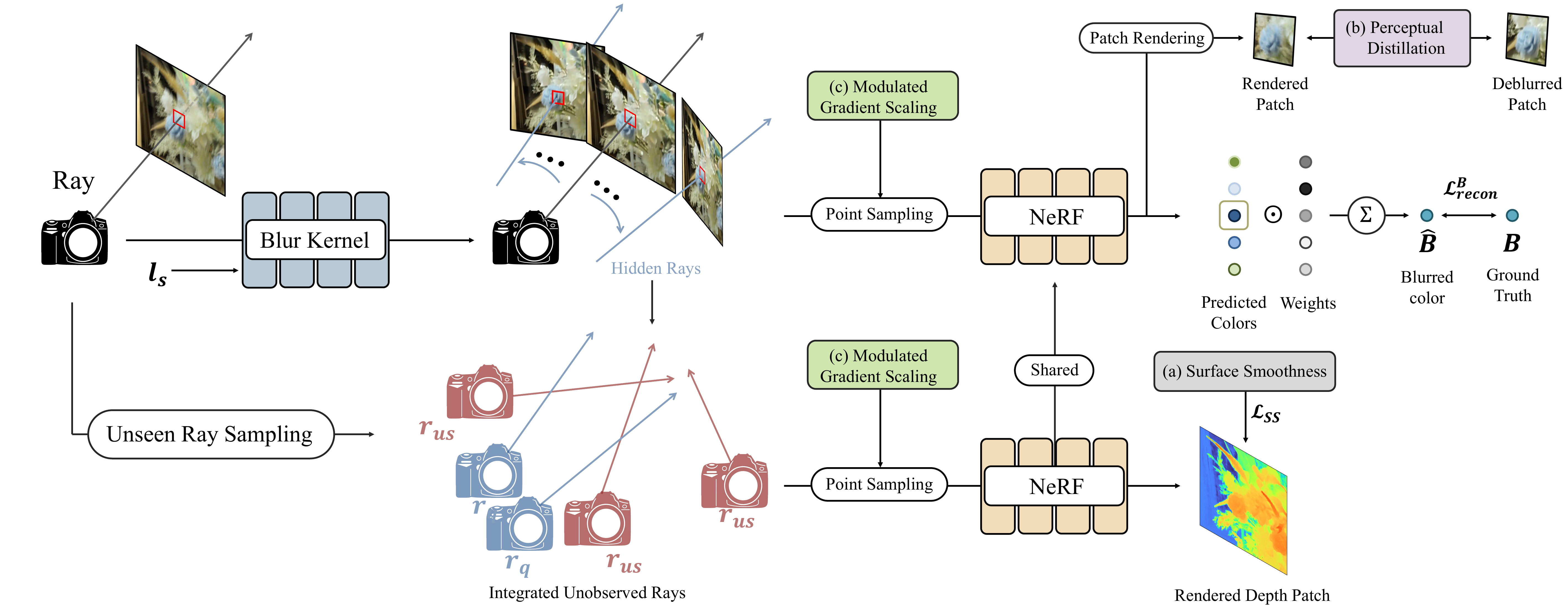}
  \vspace{-0.7cm}
  \caption{Overall architecture of the Sparse-DeRF. The main component of Spare-DeRF consists of three components, Surface Smoothness~(SS), Modulated Gradient Scaling~(MGS), and Perceptual Distillation~(PD), which are denoted as (a), (b), (c) in the figure. Note that, NeRF network is shared to predict the color of each sampled points along the ray in hidden rays, integrated unobserved rays and rays for patch rendering.}
  \vspace{-0.4cm}
  \label{fig:overall_architecture}
\end{figure*}

We find that it is not effective to construct the radiance fields based on naive NeRF, Deblur-NeRF~\cite{ma2022deblurnerf}, or DP-NeRF~\cite{lee2023dpnerf} from sparse view setting as we present in Section~\ref{subsec:evals}.
Although the DeRF usually recovers the high-frequency detail better than the naive NeRF in given views, it has become more fragmented in novel view synthesis, generating inaccurate scene geometry due to the complex joint optimization and overfitting.
The geometric error is represented as mapped RGB textures resembling painted walls in near or far depth and elongated density artifacts, which reveal challenges associated with accurate depth value prediction.
However, existing representative regularization for the NeRF from sparse view~\cite{niemeyer2022regnerf} can not regularize complex joint optimization of the DeRF.
We present experimental results in Section~\ref{subsubsec:abls_complex_optimization}, which shows the difficulty of the previous regularization technique on the DeRF from sparse view.
Hence, here we describe our method for regularizing optimization of the DeRF from sparse view to alleviate spatial ambiguity, which consists of two geometric constraints and a perceptual prior.
\figurename~\ref{fig:overall_architecture} illustrates the overall architecture of the Sparse-DeRF in detail, where (a),(b), and (c) of the \figurename~\ref{fig:overall_architecture} indicate each main component of the Sparse-DeRF, respectively.
Geometric constraints consist of surface smoothness~(SS) and modulated gradient scaling~(MGS), and a perceptual prior consists of a perceptual distillation~(PD).

\begin{figure}[h]
  \centering
  \includegraphics[width=\linewidth]{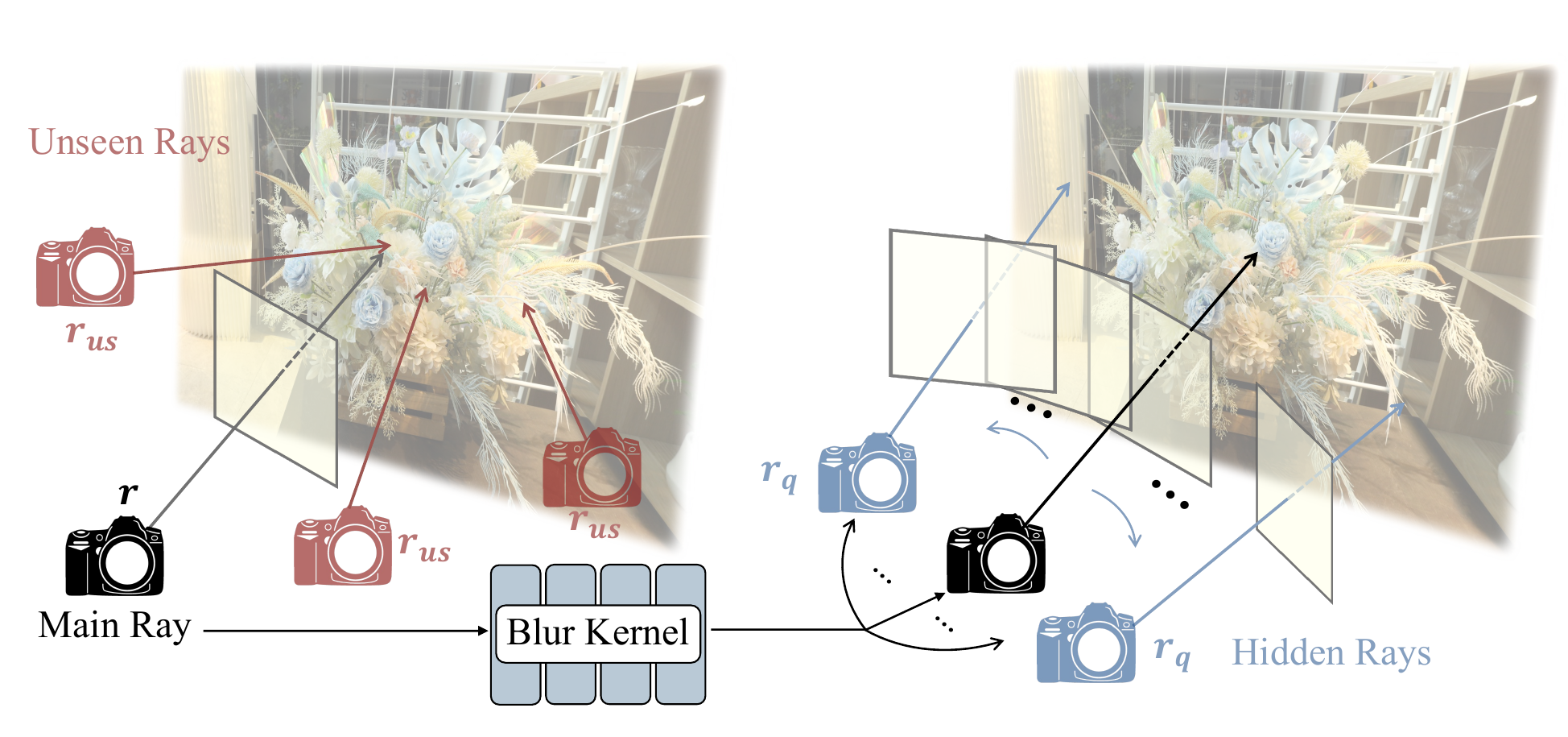}
  \vspace{-0.6cm}
  \caption{Simple illustration of unobserved rays in our method, which consists of unseen rays and hidden rays. Two types of rays are independently defined. Unseen rays remain unchanged during training, but hidden rays change during the training since it is derived from the learned blur kernel. Note that, hidden rays can be derived from both types of blur kernel we utilized.}
  \vspace{-0.4cm}
  \label{fig:difference_hidden_rays}
\end{figure}

\subsection{Surface Smoothness on Integrated Unobserved Rays}
\label{subsec:ss_method}

Inspired by the statistical tendencies of real-world geometry, piece-wise smoothness is adopted as a depth smoothness regularization on small rendered patches in RegNeRF~\cite{niemeyer2022regnerf}. 
This regularization can also be interpreted as imposing surface smoothness constraint, which is applied to the rendered depth obtained from unseen rays that are defined as rays not observed in the training inputs. 
The unseen rays are generated from possible camera locations sampled within a limited sample space, constrained by target poses for rendering during test time.
Similar to the method introduced in~\cite{niemeyer2022regnerf}, we adopted an approach to alleviate spatial ambiguity by utilizing information from unobserved rays. 
However, we propose additionally leveraging new unobserved ray information that can only be derived from the blur kernel to stabilize the simultaneous optimization of blur kernel and radiance fields.
Firstly, we utilize unseen rays as one of the integrated unobserved rays to ameliorate spatial ambiguity following the~\cite{niemeyer2022regnerf}.
For known set of camera poses $\{\textbf{P}^{i}_{target}\}_{i}$, where $\textbf{P}^{i}_{target}=[\textbf{R}^{i}_{target}|\textbf{t}^{i}_{target}]\in SE(3)$, sampled camera pose $S_{\textbf{P}}$ for unseen rays is formulated from camera location $S_{\textbf{t}}$ and rotation $S_{\textbf{R}}$ in limited sample space as
\begin{align} 
	\label{eq:unseenrays_sample_space_rt}
	\begin{split}
		S_{\textbf{t}} &= \{\textbf{t}\in \mathbb{R}^{3} | \textbf{t}_{min} \leq \textbf{t} \leq \textbf{t}_{max}\},\\
		S_{\textbf{R}}|\textbf{t} &= \{\textbf{R}(\bar{\textbf{p}}_{u}, \bar{\textbf{p}}_{f} + \epsilon, \textbf{t})| \epsilon\sim\mathcal{N}(0,0.125)\},
	\end{split}
\end{align}
where $\textbf{t}_{min}$, $\textbf{t}_{max}$, $\bar{\textbf{p}}_{u}$, and $\bar{\textbf{p}}_{f}$ indicate $min(\{\textbf{t}^{i}_{target}\})$, $max(\{\textbf{t}^{i}_{target}\})$, the normalized mean over the up axes of all target poses, and the mean focus point by solving a least squares problem, respectively.
$\textbf{R}(\cdot,\cdot,\cdot)$ and $\epsilon$ indicate camera rotation matrix to make the sampled camera roughly focus on a central point of a scene and a small jitter value added to the focus point, respectively.
To the end, sampled camera poses $S_{\textbf{p}}$ is formulated as
\begin{equation} 
	\label{eq:unseenrays_sample_space}
	S_{\textbf{P}} = \{[\textbf{R}|\textbf{t}] |~\textbf{R}\sim S_{\textbf{R}}|\textbf{t}, \textbf{t}\sim S_{\textbf{t}}\}.
\end{equation}

In addition to leveraging previously unseen rays, we employ hidden rays derived from the characteristics of the estimated blur kernel, harnessing supplementary information exclusively presented in blurry inputs. 
The blur kernel, denoted as $h$ in Eq. \ref{eq:blurring_process}, generates the ray transformation to approximate the color composition process of the blur, regardless of the kernel types, as demonstrated in Deblur-NeRF~\cite{ma2022deblurnerf} and DP-NeRF~\cite{lee2023dpnerf}. 
Motivated by the commonality in the color composition process across various blur kernels, these kernel-induced transformed rays $\textbf{r}_{q}$ are introduced as additional unobserved rays, which are referred to as hidden rays, to enforce depth smoothness constraint to the blur kernel. 
As the hidden rays $\textbf{r}_{q}$ are not directly presented in the training data for the DeRF system, they serve as supplementary data for imposing depth smoothness regularization effect to the blur kernel and stabilize the complex optimization of blur kernel and radiance fields.
In addition, the incorporation of depth smoothness on hidden rays effectively help alleviate the geometric inconsistency across the 3D space of blurry images within the specified training view.
Therefore, our integrated unobserved rays for depth smoothness are defined as an integrated set of unseen rays and hidden rays as
\begin{equation} 
	\label{eq:integrated_unobserved_rays}
	\textbf{r}_{iu} = \{\textbf{r}_{q}, \textbf{r}_{us}\},~\indent where~ \textbf{r}_{q}\sim h(\textbf{r}),~\textbf{r}_{us}\sim S_{\textbf{p}},
\end{equation}
where $r_{iu}$ and $r_{us}$ denote integrated unobserved rays and unseen rays, respectively.
For more intuitive understanding, the $\textbf{r}_{q}$ and $\textbf{r}_{us}$ are illustrated in~\figurename~\ref{fig:difference_hidden_rays}.

For applying depth smoothness constraint on $r_{iu}$, the expected depth of $r_{iu}$ is computed following Eq.\ref{eq:depth_smoothness_loss} as same as previous NeRF works.
\begin{equation} 
	\label{eq:depth_smoothness_loss}
	\hat{d} =  \int_{t_{n}}^{t_{f}}T(t)\sigma(\textbf{r}(t))tdt.
\end{equation}
Then the depth smoothness loss is reformulated by adding color-dependent weighted depth smoothness from~\cite{niemeyer2022regnerf} as
\begin{align} 
	\label{eq:depth_smoothness_patch}
	\mathcal{L}_{ss}\{\textbf{r}_{iu}\} &= \sum_{\textbf{r}\in \textbf{r}_{iu}}\sum^{S_{ptc}-1}_{i,j=1}\Bigl[\Delta\hat{d}_{x}(\textbf{r}_{i,j})+\Delta\hat{d}_{y}(\textbf{r}_{i,j})\Bigl],
\end{align}
where $\Delta\hat{d}_{x}(\textbf{r}_{i,j})$ and $\Delta\hat{d}_{y}(\textbf{r}_{i,j})$ indicates horizontal and vertial weighted depth difference as Eq.\ref{eq:weighted_depth_smoothness}, respectively.
\begin{align} 
	\label{eq:weighted_depth_smoothness}
	\begin{split}
	\Delta\hat{d}_{x}(\textbf{r}_{i,j}) &= \omega_{i+1,j}(\hat{d}(\textbf{r}_{i,j})-\hat{d}(\textbf{r}_{i+1,j}))^{2} \\
	\Delta\hat{d}_{y}(\textbf{r}_{i,j}) &= \omega_{i,j+1}(\hat{d}(\textbf{r}_{i,j})-\hat{d}(\textbf{r}_{i,j+1}))^{2},
	\end{split}
\end{align}
where $\omega_{i+k, j+1}=exp\big( -\big(\hat{C}(\textbf{r}_{i,j})-\hat{C}(\textbf{r}_{i+k,j+l})\big)^{2}\big)$ indicates pixel-wise color difference weight.

\begin{figure*}[t]
  \centering
  \includegraphics[width=\linewidth]{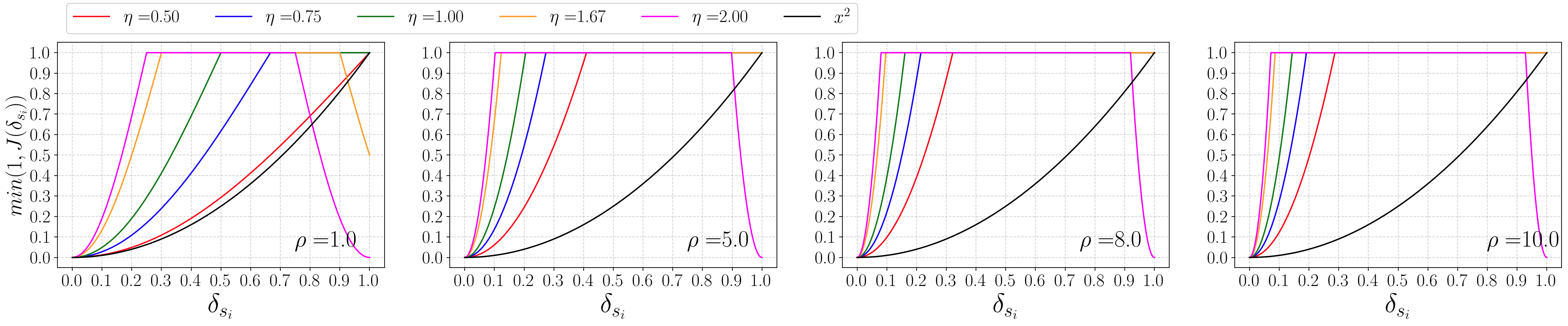}
  \vspace{-0.5cm}
  \caption{The comparison between ours modulated gradient scaling function with $\hat{J}(\delta_{\textbf{s}_{i}})$ and the previous function $J(\delta_{\textbf{s}_{i}})$ proposed by \cite{philip2023floatersnomore}, with respect to the ray distance $\delta_{\textbf{s}_{i}}$. In the table, the values of $\delta_{\textbf{s}_{i}}$ and $min(1, J(\delta_{\textbf{s}_{i}}))$ on the x-axis and y-axis represent the ray distance from the camera origin and gradient scaling value, respectively. The $J(\delta_{\textbf{s}_{i}})$ of~\cite{philip2023floatersnomore} is represented as $x^2$ with the black colored line for clarity. The graphs illustrate that our modulated function $\hat{J}(\delta_{\textbf{s}_{i}})$ can cover the diversity in the arrangement of scene components, exhibiting various shapes of the function depending on magnitude $\rho$ and period $\eta$.}
  \vspace{-0.2cm}
  \label{fig:modulated_gradient_scaling}
\end{figure*}

\subsection{Modulated Gradient Scaling}
\label{subsec:mgs_method}
Although previous surface smoothness alleviates the spatial ambiguity in the 3D scene, it is still hard to grasp accurate geometry due to the inherent drawback of the NeRF sampling strategy and casted volume occupancy as~\cite{philip2023floatersnomore} argued.
Following~\cite{philip2023floatersnomore}, the optimization of the NeRF often fails, generating the density artifact in the near-depth region due to the disproportionate gradient backpropagation induced from the imbalanced volumetric occupancy of the samples on the ray.
\cite{philip2023floatersnomore} alleviates the limitation introducing gradient scaling that reduces the propagated gradient $\nabla \textbf{s}_{i}$ of each $i$-th sample $\textbf{s}_{i}=(s_x, s_y, s_z)$ on ray $\textbf{r}$ according to the distance $\delta_{\textbf{s}_{i}}$ from ray origin $\textbf{o}$ as
\begin{equation} 
	\label{eq:gradient_scaling}
	\nabla \hat{\textbf{s}}_{i} = \nabla \textbf{s}_{i} \times min(1, J(\delta_{\textbf{s}_{i}})),~where~\delta_{\textbf{s}_{i}}=|\textbf{s}_{i}-\textbf{o}|,
\end{equation}
where $\nabla \hat{\textbf{s}_{i}}$ and $J$ indicate scaled gradient value for sample $s_{i}$ and scaling function.
The scaling function $J$ is strictly formulated as a squared function as
\begin{equation} 
	\label{eq:gradient_scaling_function}
	J(\delta_{\textbf{s}_{i}}) = (\delta_{\textbf{s}_{i}})^2.
\end{equation}
Note that, $\nabla \textbf{s}_{i}$ indicates the gradients of per-point characteristics such as RGB color $\textbf{c}$ and density $\sigma$.

However, we experimentally found that the limitation is more prominent in sparse view settings since there is less available diversity of viewing direction, which means the projective geometry and epipolar geometry do not properly work for NeRF optimization.
Therefore, the gradient scaling seems to be more necessary to the radiance fields from the sparse view setting.
In addition, we believe the proper gradient scaling can be an implicit geometric guidance in sparse view setting.
Although we tried to apply the technique to the Sparse-DeRF, it does not properly works as demonstrated in the appendix.
The reason is that the fixed square function of $J$, which is lower bounded as $1$, does not cover the non-linear parametrized space such as normalized device coordinates~(NDC), which is commonly used as well as our work.
Another reason is that a strictly fixed shape of the function cannot cover the arrangements of the scene components, which are usually different across the scene even in the same dataset.
\cite{philip2023floatersnomore} briefly mentioned the determinant of the jacobian as an additional scaling factor for the value of $J$, but they did not experiment on it.
Moreover, even if the additional scaling factor were applied, the shape of the scaling function would remain unchanged and simply in the form of the square function, which does not allow for flexible adaptation to the arrangement of scene components or guide the rough geometry of the radiance fields.
Therefore, we modulate the shape of the scaling function $J$ to adaptively reflect the scene arrangements and guide the rough geometry while suitable for NDC, since we use NDC for our experiments.

Our novel gradient scaling function $\hat{J}$ is designed based on three conditions.
First, the function should increase from zero at the camera origin, which is a critical condition to avoid the local minima in the initial training phase we mentioned before.
Second, the function should not be zero in the far distance, which is set to $1$ in our NDC environment, to ensure the NeRF training.
Second, the function should not be zero in the far distance, which is set to $1$, to ensure the NeRF training.
Finally, the function is designed to increase and decrease only once within a given depth range, which makes the function not fluctuate, since it is an intuitively reasonable scenario considering the goal of MGS that alleviates incorrect density mapping in near distance.
In addition to the above conditions for proper shape of the scaling function, we further consider designing the function shape when the location of the main objects is focused on the center of the scene and density mapping error that is represented as a painted wall of near or far depth.
Following the conditions, the proposed modulated gradient scaling~(MGS) function $\hat{J}$ is formulated as
\begin{equation} 
	\label{eq:modulated_gradient_scaling_function}
	\hat{J}(\delta_{\textbf{s}_{i}}) = \rho\Big(\sin\big(\eta\pi(\delta_{\textbf{s}_{i}}+\frac{3}{2\eta})\big)+1\Big),
\end{equation}
where $(1\leq\rho\leq10)$ and $(\frac{1}{2}\leq\eta<2)$ denote magnitude and period of sinusoidal function, respectively.
However, in contrast to~\cite{philip2023floatersnomore}, the distance range of $\delta_{\textbf{s}_{i}}$ is restricted to $\delta_{\textbf{s}_{i}}\in[0,1]$ since we use NDC for our dataset.

To apply gradient scaling both in the near and far regions while minimizing the scaling effect in the center of the scene, we adopt the sinusoidal function shape as the foundation for our MGS.
Furthermore, the second condition determines the maximum value of $\eta$ as 2 to ensure the scaling value of the proposed function in the far region does not fall to or below 0.
The function shape in the left top image of \figurename~\ref{fig:modulated_gradient_scaling} shows the characteristics of the proposed scaling function that we described above.
We shows the difference between the scaling value from $min(1, J(\delta_{\textbf{s}_{i}}))$ and $min(1, \hat{J}(\delta_{\textbf{s}_{i}}))$, according to the various $\rho$ and $\eta$ values in \figurename~\ref{fig:modulated_gradient_scaling}.

\begin{figure}[t]
  \centering
  \includegraphics[width=\linewidth]{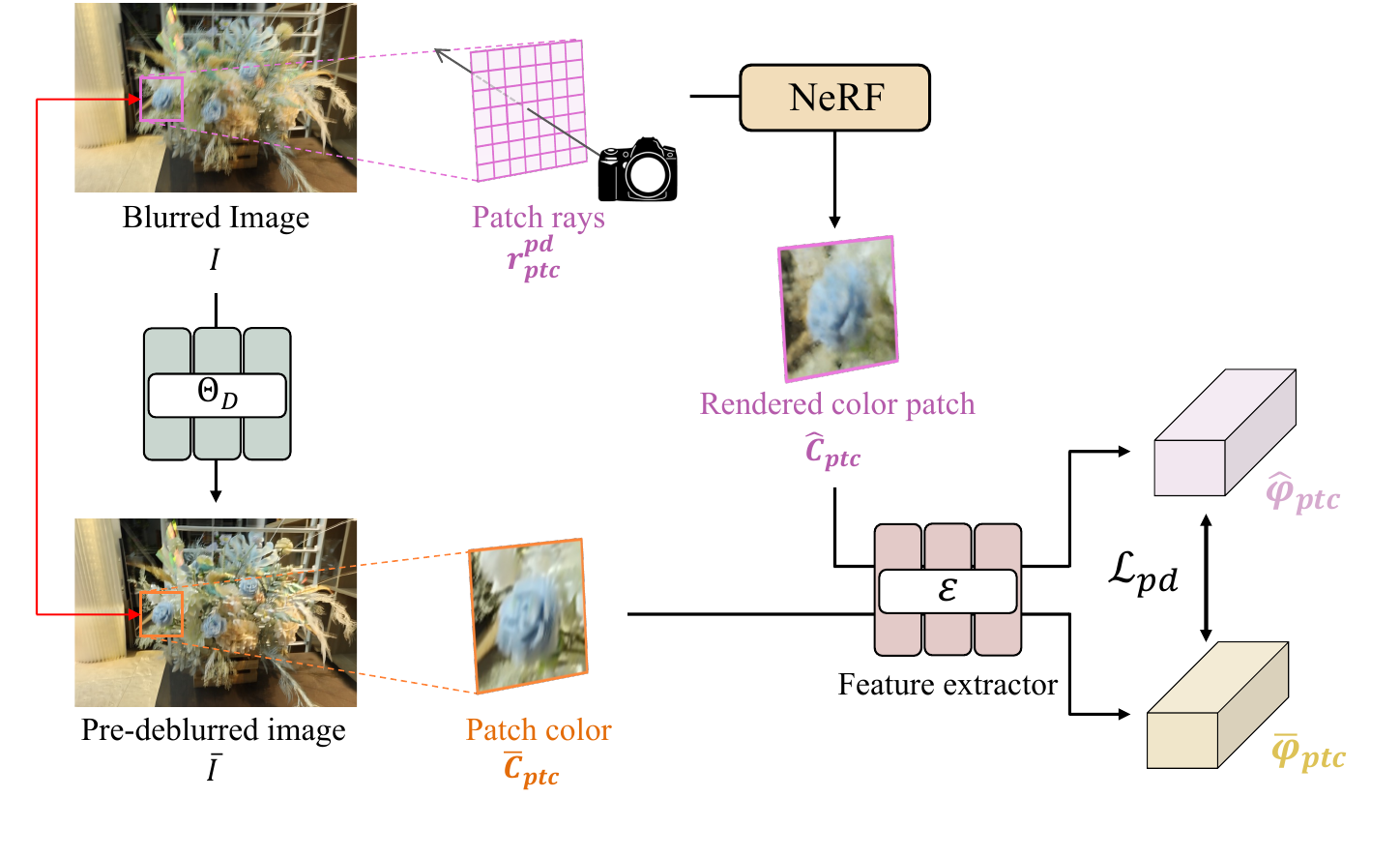}
  \vspace{-0.8cm}
  \caption{Illustration of our perceptual distillation. Perceptual distillation transfers the information of pre-deblurred texture by applying the perceptual loss to the pre-deblurred color patch $\bar{C}_{ptc}$ and rendered color patch $\hat{C}_{ptc}$, which is rendered from patch-wise sampled rays $\textbf{r}^{pd}_{ptc}$ in same pixel location. Note that $\Theta_{D}$ and $\mathcal{E}$ are pre-trained image deblurring network and a shared pre-trained image feature extractor, respectively.}
    \vspace{-0.1cm}
  \label{fig:perceptual_distillation}
\end{figure}

\subsection{Perceptual Distillation}
\label{subsec:pd_method}
In contrast to the previous NeRF-related works in sparse view settings, the Sparse-DeRF environments enable to use the off-the-shelf image processing algorithms, such as deep learning-based image deblurring networks, due to the degradation of the given images.
In addition to improvements from a geometric perspective, we aim to enhance the detailed textures of DeRF utilizing the advantages of existing image processing modules, thereby achieving high fidelity.

However, it is not possible to directly utilize the pre-deblurred images as additional pixel-wise color supervision, due to the lack of 3D consistency.
This inconsistency occurs due to the ill-posed property of image deblurring and independent deblurring processing across multi-view images, which generates incoherent deblurred results.
In Section~\ref{subsubsec:abls_inconssitency}, we additionally address this issue and present the qualitative comparison of pre-deblurred images and reference images, which are estimated from the DP-NeRF~\cite{lee2023dpnerf} trained with the full view, to reveal the geometric inconsistency issue.

To overcome the intrinsic geometric inconsistency of pre-deblurred images, we address the pre-deblurred images as a perceptual prior, which transfers the only deblurred texture information by extracting features from rendered patches and deblurred images with a pre-trained feature extractor.
\figurename~\ref{fig:perceptual_distillation} simply illustrates the pipeline of our perceptual distillation module.
To the best of our knowledge, utilizing pre-deblurred images as an additional prior in deblurring radiance fields is a first attempt.

Specifically, we first generate deblurred images $\bar{I}$ for blurry training images $I$ by exploiting a pre-trained image deblurring network $\Theta_{D}$ to prepare the feature extraction as 
\begin{equation} 
	\label{eq:pretrained_image_deblurring}
	\bar{I}=\Theta_{D}(I)
\end{equation}
Next, we additionally sample the $S^{pd}_{ptc}\times S^{pd}_{ptc}$ size of patch rays $\textbf{r}^{pd}_{ptc}$ and corresponding deblurred image patch $\bar{C}_{ptc}$ from $\bar{I}$ on training views.
Then we extract the abundant deblurred features from the rendered color patch $\hat{C}_{ptc}$, which is rendered from $\textbf{r}^{pd}_{ptc}$, and pre-deblurred images $\bar{I}$ using a shared pre-trained image feature extractor $\mathcal{E}$ as 
\begin{equation} 
	\label{eq:image_feature_extraction}
	\hat{\varphi}_{ptc}=\mathcal{E}(\hat{C}_{ptc}),\indent \bar{\varphi}_{ptc}=\mathcal{E}(\bar{C}_{ptc}), 
\end{equation}
where $\hat{\varphi}_{ptc}$ and $\bar{\varphi}_{ptc}$ indicate extracted features from each color patches, respectively.
Note that, the color patch $\hat{C}_{ptc}$ is rendered by forwarding the NeRF MLPs without blur kernel to perceptually transfer the pre-deblurred texture information to the implicit clean radiance fields, which is indicated as patch radiance in \figurename~\ref{fig:overall_architecture}. 
Then we apply the perceptual loss $\mathcal{L}_{pd}$~\cite{johnson2016perceptualloss} to distill the feature information as
\begin{equation}
	\label{eq:perceptual_loss}
	\mathcal{L}_{pd} = ||\hat{\varphi}_{ptc}-\bar{\varphi}_{ptc}||^{2}_{2} ~.
\end{equation}
 Our final loss function is a weighted composition of the proposed losses as
 \begin{equation}
	\label{eq:final_loss}
	\mathcal{L}_{final} = \mathcal{L}^{B}_{recon} + \lambda_{ss}\mathcal{L}_{ss} + \lambda_{pd}\mathcal{L}_{pd},
\end{equation}
where $\lambda_{ss}$ and $\lambda_{pd}$ denote weights for each loss, which are equally set to $0.01$ in our experiments.

%% file: sections/experiments.tex
\newcolumntype{L}[1]{>{\raggedright\let\newline\\\arraybackslash\hspace{0pt}}m{#1}}
\newcolumntype{C}[1]{>{\centering\let\newline\\\arraybackslash\hspace{0pt}}m{#1}}
\newcolumntype{R}[1]{>{\raggedleft\let\newline\\\arraybackslash\hspace{0pt}}m{#1}}
\definecolor{lightgray}{rgb}{0.83, 0.83, 0.83}

\section{Experiments}
\label{sec:exps}
\subsection{Dataset}
\label{subsec:imple_dataset_exps}
Sparse-DeRF has experimented with a forward-facing scene dataset proposed by Deblur-NeRF~\cite{ma2022deblurnerf}, which includes 5 synthetic and 10 real scenes. 
In particular, we use only the camera motion blur dataset since our goal is to alleviate the blur from camera motion in the DeRF from sparse view settings.
The dataset consists of multiple view images and paired camera poses calibrated by using COLMAP~\cite{schonberger2016pixelwise, schonberger2016structure}.
For the sparse view setting, we manually select the 2, 4, and 6 images as training datasets for all scenes, so that the entire space covered by each view is as wide as possible.
In addition, to ensure reasonable learning of radiance fields, we select views with visible spaces that overlap as little as possible, while avoiding excessively extreme blur magnitudes.
Note that, it is an inherent property of the dataset that blur magnitudes of selected views can be different according to each scene, which means that each scene has a different level of learning difficulty.
We attach the selected image indices of all scenes used in our experiments in the appendix for fair comparison in future research.

\subsection{Experimental Sparse View Setting}
\label{subsec:sparse_view_setting}
\begin{figure*}[t]
  \centering
  \includegraphics[width=\textwidth]{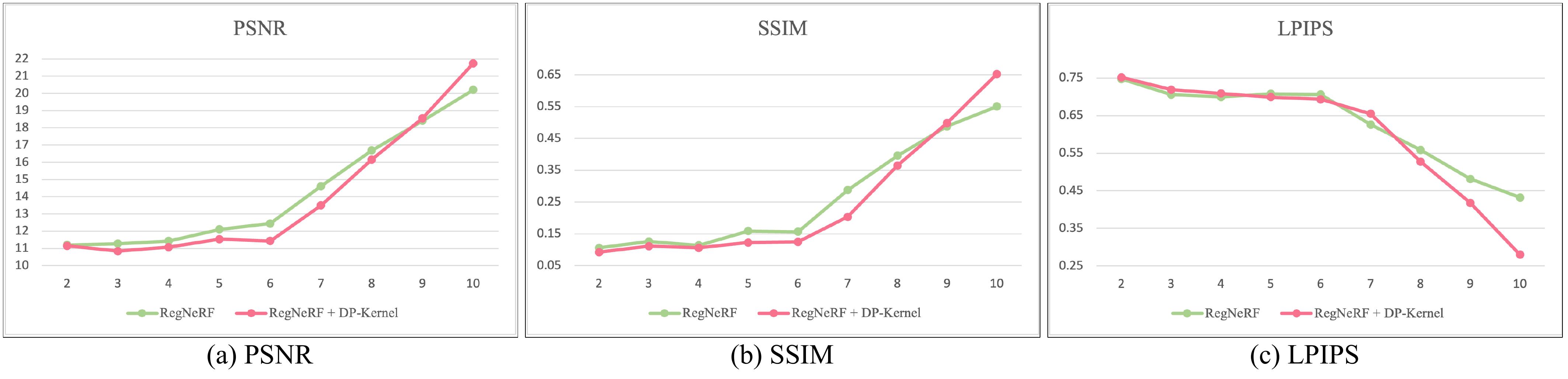}
  \vspace{-0.7cm}
  \caption{Graph of the experimental results of \textit{Decoration} scene from \textbf{2-view} to \textbf{10-view} settings. \figurename~(a), (b), and (c) indicate PSNR, SSIM, and LPIPS, respectively. The graph reveals that the network suffers joint optimization problems under the 9-view setting, which makes our experimental setting plausible.}
  \vspace{-0.2cm}
  \label{fig:appendix_regnerf_ablation_2view_to_10_view_decoration}
\end{figure*}

Scenarios for the Sparse-DeRF are assumed to be three kinds of settings, which are the DeRF from 2-view, 4-view, and 6-view settings unlike existing common sparse-view NeRFs, which usually use 3-view, 6-view, and 9-view settings.
The reason is that the joint optimization problem occurs more frequently under 9-view settings as shown in~\figurename~\ref{fig:appendix_regnerf_ablation_2view_to_10_view_decoration}.
The figure presents the graph of experiments in the \textit{Decoration} scene where we varied the number of input sparse views from 2-view to 10-view.
We also attach the quantitative results in the appendix.
Experimental results reveal that our sparse-view experimental settings are valid since the RegNeRF~\cite{niemeyer2022regnerf} with \textcolor[rgb]{0.13, 0.55, 0.13}{DP-kernel} shows poor performances under the 9 views.

\subsection{Implementation Details}
\label{subsec:imple_detail}
Spare-DeRF is implemented and modified based on the published official code of DP-NeRF~\cite{lee2023dpnerf} using the two types of blur kernels of DP-NeRF~\cite{lee2023dpnerf} and Deblur-NeRF~\cite{ma2022deblurnerf}.
For a fair comparison, the number of blurring rays is set to $5$ for the default setting as same as previous works~\cite{ma2022deblurnerf,lee2023dpnerf}.
We set the other settings for the blur kernels following the default parameters of each work.
For the NeRF optimization, we use $64$ coarse and $64$ fine samples per ray with a batch size of 1024 rays, exploiting Adam~\cite{kingma2014adam} optimizer with default parameters. 
In addition, exponential weight decay is applied from $5\times10^{-4}$ to $8\times10^{-5}$ for learning rate scheduling.
We train the DeRF for $20k$, $40k$, and $60k$ iterations in 2-view, 4-view, and 6-view settings, respectively.
For patch-wise sampling, we set the size of the patch $S_{ptc}$ and $S^{pd}_{ptc}$ as $8$ and $64$, respectively.
However, the generating method of unseen rays in surface smoothness constraint is especially considered for fair comparison across the extensive experiments.
In particular, we generate the unseen rays from fixed views, which are evenly selected from the rest of the training rays, to remove the randomness of the unseen ray generation for experimental analysis and fair comparison.
Note that, the rest of the training rays are not included in the training or test view.
We demonstrate that it is a reasonable choice since the selected views are still in the sample space we defined in Section~\ref{subsec:ss_method}.
In addition, we attach the visualization of this issue in the appendix to show the rationality.
The hyper-parameters of the modulated gradient scaling function for each scene, magnitude $\rho$ and period $\eta$, are attached in the appendix in detail.
For perceptual distillation, MPRNet~\cite{zamir2021MPR} is utilized as a pre-trained image deblurring network.
We select the VGG19~\cite{simonyan2014vggnet} as a pre-trained image feature extractor $\mathcal{E}$ since it is widely exploited as an image feature extractor in image-based computer vision.

\subsection{Evaluation Metrics}
\label{subsubsec:metric_evals}
Our experimental results for the synthetic and real datasets are evaluated in three quantitative metrics and qualitative comparisons between rendered images through a novel view synthesis task.
Consistent with prior research, we employ widely utilized evaluation metrics to compare the synthesized images with corresponding ground truth images: the peak signal-to-noise ratio~(PSNR), the structural similarity index measure~(SSIM), and learned perceptual image patch similarity~(LPIPS)~\cite{zhang2018lpips}.
These metrics assess the relative sharpness, structural similarity, and perceptual quality of the generated images, respectively.
In addition, we encourage readers to refer to the supplementary video for a more comprehensive and detailed presentation of the results.

\subsection{Evaluation}
\label{subsec:evals}

\begin{table*}
   \footnotesize
   \begin{center}
      \caption{\textbf{Average} results on both synthetic and real scenes obtained from 2-view, 4-view, and 6-view settings. Each color shading represents the \colorbox{best!35}{best}, \colorbox{second!35}{second best}, and \colorbox{third!35}{third best} result for each experimental setting, respectively.}
      \vspace{-0.4cm}
      \renewcommand\arraystretch{1.2}
      \resizebox{\linewidth}{!}{
		\setlength{\tabcolsep}{1pt}
         \begin{tabular}{c||ccc|ccc||ccc|ccc||ccc|ccc||}
	        & \multicolumn{6}{c}{[~2-view ]} & \multicolumn{6}{c}{[~4-view ]} & \multicolumn{6}{c}{[~6-view ]}\\
			& \multicolumn{3}{c}{[~Synthetic Scene~]} & \multicolumn{3}{c}{[~Real Scene~]} & \multicolumn{3}{c}{[~Synthetic Scene~]} & \multicolumn{3}{c}{[~Real Scene~]} & \multicolumn{3}{c}{[~Synthetic Scene~]} & \multicolumn{3}{c}{[~Real Scene~]}\\
			& ~PSNR($\uparrow$)~ & SSIM($\uparrow$)~ & LPIPS($\downarrow$)~& ~PSNR($\uparrow$)~ & SSIM($\uparrow$)~ & LPIPS($\downarrow$)~ & ~PSNR($\uparrow$)~ & SSIM($\uparrow$)~ & LPIPS($\downarrow$)~& ~PSNR($\uparrow$)~ & SSIM($\uparrow$)~ & LPIPS($\downarrow$)~ & ~PSNR($\uparrow$)~ & SSIM($\uparrow$)~ & LPIPS($\downarrow$)~& ~PSNR($\uparrow$)~ & SSIM($\uparrow$)~ & LPIPS($\downarrow$)~ \\
			\midrule
			Naive NeRF~\cite{mildenhall2021nerf}~ & 15.11 &\cellcolor{second!35}.2999 & .5578 & 14.38 & .2635 & .6004 & \cellcolor{third!35}20.02 & .5327 & .4000 & \cellcolor{third!35}18.98 & \cellcolor{third!35}.4860 & .4481 & 21.65 & .5985 & .3638 & 20.63 & .5513 & .4014  \\
			NeRF~+~MPR~\cite{zamir2021MPR} & \cellcolor{third!35}15.16 & \cellcolor{best!35}.3006 & .5595 & 14.38 & .2594 & .6019 & 20.00 & \cellcolor{third!35}.5381 & .3956 & 18.89 & .4829 & .4484 & 21.72 & .5999 & .3629 & 20.60 & .5513 & .4010 \\
			\midrule
			Deblur-NeRF~\cite{ma2022deblurnerf}~ & 15.14 & .2884 & \cellcolor{third!35}.5330  & 14.41 & .2506 & .5921 & 19.99 & .5199 & .3499 & 18.89 & .4761 & \cellcolor{third!35}.4151 & 23.12 & .6798 & .2386 & 21.36 & .6003 & .3163 \\	
			DP-NeRF~\cite{lee2023dpnerf} & 15.06 & .2827 & .5389 & 14.36 & .2506 & .5904 & 19.84 & .5336 & \cellcolor{second!35}.3075 & 18.77 & .4582 & .4175 & \cellcolor{third!35}23.68 & \cellcolor{second!35}.7036 & \cellcolor{best!25}.1998 & \cellcolor{third!35}21.68 & \cellcolor{third!35}.6137 & \cellcolor{second!35}.2992 \\	
			\midrule		
			RegNeRF~\cite{niemeyer2022regnerf}~(No kernel) & 14.60 & .2849 & .5869 & 15.49 & .2997 & .5888 & 18.39 & .4600 & .4704 & 18.44 & .4326 & .4852 & 19.69 & .5249 & .4165 & 19.65 & .4790 & .4583\\	
			RegNeRF~\cite{niemeyer2022regnerf}~(No kernel) + MPR~\cite{zamir2021MPR} & 14.99 & .2980 & .5690 & \cellcolor{best!35}15.84 & \cellcolor{third!35}.3040 & \cellcolor{third!35}.5818 & 18.54 & .4678 & .4593 & 18.66 & .4425 & .4804 & 19.89 & .5339 & .4071 & 19.64 & .4753 & .4597\\	
			RegNeRF~\cite{niemeyer2022regnerf}~(w/\textcolor[rgb]{0.13, 0.55, 0.13}{DP-kernel}) & 13.24 & .2162 & .6062 & 12.76 & .1836 & .6447 & 16.44 & .3657 & .4826 & 13.59 & .2221 & .5887 & 21.60 & .6162 & .3046 & 18.25 & .4439 & .4326 \\		
			\midrule
			Sparse-DeRF~(w/\textcolor[rgb]{0.25, 0.0, 1.0}{DN-kernel})~-~Ours & \cellcolor{best!35}15.52 & \cellcolor{third!35}.2966 & \cellcolor{second!35}.5291 & \cellcolor{third!35}15.53 & \cellcolor{second!35}.3112 & \cellcolor{second!35}.5515 & \cellcolor{second!35}20.57 & \cellcolor{second!35}.5565 & \cellcolor{third!35}.3354 & \cellcolor{second!35}19.98 & \cellcolor{best!35}.5231 & \cellcolor{second!35}.3871 & \cellcolor{second!35}23.32 & \cellcolor{third!35}.6903& \cellcolor{third!35}.2379 & \cellcolor{second!35}22.15 & \cellcolor{second!35}.6248 & \cellcolor{third!35}.3030 \\
			Sparse-DeRF(w/\textcolor[rgb]{0.13, 0.55, 0.13}{DP-kernel})~-~Ours & \cellcolor{second!35}15.35 & .2904 & \cellcolor{best!35}.5242 & \cellcolor{second!35}15.57 & \cellcolor{best!35}.3114 & \cellcolor{best!35}.5467  & \cellcolor{best!35}21.05 & \cellcolor{best!35}.5776 & \cellcolor{best!35}.2975  & \cellcolor{best!35}20.05 & \cellcolor{second!35}.5178 & \cellcolor{best!35}.3736 & \cellcolor{best!35}24.27 & \cellcolor{best!35}.7255 & \cellcolor{second!35}.2044 & \cellcolor{best!35}22.32 & \cellcolor{best!35}.6283 & \cellcolor{best!35}.2907 \\
			\midrule
      \end{tabular}
      }
   \vspace{-0.5cm}
   \label{tab:average_quantitative_results}
   \end{center}
\end{table*}

\begin{figure*}[t]
  \centering
  \includegraphics[width=0.99\textwidth]{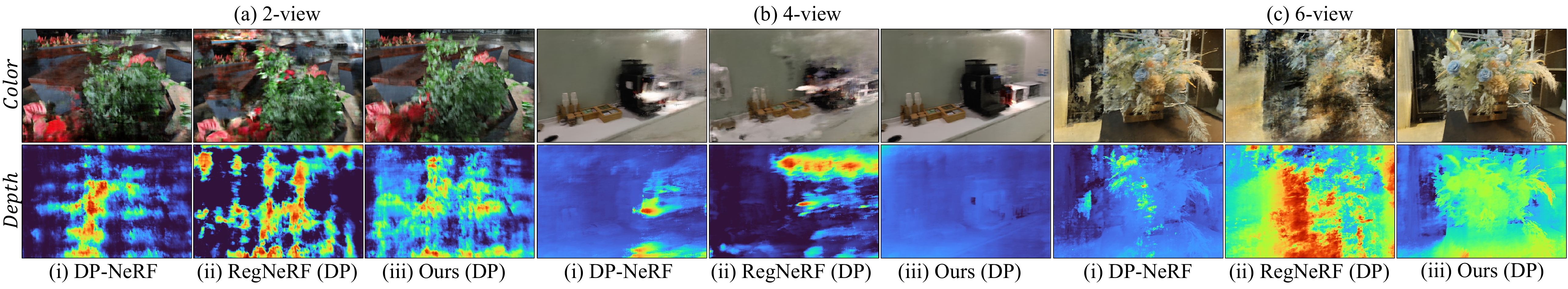}
  \vspace{-0.3cm}
  \caption{Qualitative results on \textit{Parterre},~\textit{Coffee}, and~\textit{Decoration} scene from (a) 2-view, (b) 4-view, and (3) 6-view respectively. (i), (ii), and (iii) denote rendered color and depth images from DP-NeRF~\cite{lee2023dpnerf}, RegNeRF~\cite{niemeyer2022regnerf} with \textcolor[rgb]{0.13, 0.55, 0.13}{DP-kernel}, and Ours(Sparse-DeRF) with \textcolor[rgb]{0.13, 0.55, 0.13}{DP-kernel}.}
  \label{fig:qualitative_results_all}
  \vspace{-0.2cm}
\end{figure*}

\subsubsection{Quantitative Evaluation}
\label{subsubsec:quant_res_evals}
We present the quantitative results of Sparse-DeRF for two different types of blur kernels from Deblur-NeRF~\cite{ma2022deblurnerf} and DP-NeRF~\cite{lee2023dpnerf}, comparing these results with established baseline methods.
The effectiveness of our approach is demonstrated in \tablename~\ref{tab:average_quantitative_results}, achieving outstanding performance across entire sparse view settings, regardless of the blur kernel employed.
In the Sparse-DeRF results, \textcolor[rgb]{0.25, 0.0, 1.0}{DN-kernel} and \textcolor[rgb]{0.13, 0.55, 0.13}{DP-kernel} denote the blur kernels proposed by the Deblur-NeRF~\cite{ma2022deblurnerf} and DP-NeRF~\cite{lee2023dpnerf}, respectively.
NeRF$+$MPR denotes the naive NeRF model trained solely on color supervision from deblurred images by MPRNet~\cite{zamir2021MPR}, utilizing the reconstruction loss $\mathcal{L}_{recon}$ of Eq.\ref{eq:nerf_reconloss}, which is samely utilized in RegNeRF+MPR experiment.
The results of the experiment with MPR~\cite{zamir2021MPR} reveal interesting observations and marginal improvements in radiance fields when only employing pre-deblurred images as direct color supervision for training.
This tendency even prominent when observing the results of applying MPR~\cite{zamir2021MPR} to both NeRF~(2 row in~\tablename~\ref{tab:average_quantitative_results}) and RegNeRF~(6 row in~\tablename~\ref{tab:average_quantitative_results}). 
Regardless of the presence or absence of a regularizer, the improvement is marginal and the performance of perceptual quality(LPIPS score) even decreases when a larger number of images is used due to the inconsistency as we mentioned above. 
In the RegNeRF+MPR experiments, we observe a slight increase in PSNR in the 2-view setting. 
In this case, we expect the somewhat cleaner image from the pre-deblurred image appears to be the advantage, since there are significant lack of information in the experimental environment.
Nonetheless, perceptual quality significantly deteriorates with poor LPIPS scores, which demonstrates the pre-deblurred images are not quite helpful to enhance the quality of the reconstructed radiance fields.
This tendency emphasizes the 3D inconsistency across the pre-deblurred images, as discussed in Section~\ref{subsec:pd_method}.

In addition, the poor results of the RegNeRF~\cite{niemeyer2022regnerf} with and without a blur kernel demonstrate that existing regularization faces difficulty in alleviating the complex joint optimization involving both the blur kernel and radiance fields.
We further present an analysis of this optimization issue in Section~\ref{subsubsec:abls_complex_optimization} with detailed experimental results.
In contrast, Sparse-DeRF demonstrates significant enhancements across all evaluation metrics for both types of blur kernels, indicating its superior ability to represent the DeRF with improved visual quality.
In particular, our results exhibit more prominent improvements in real-scene scenarios, although there is non-ideal blur degradation, which occurs due to various real environmental factors.
For a comprehensive understanding, we provide an extensive ablation study in Section~\ref{subsec:abls}, incorporating results with both types of blur kernels.
Additionally, detailed results for all scenes are appended in the appendix, including synthetic and real scenes.

\subsubsection{Qualitative Evaluation}
\label{subsubsec:qual_res_evals}
In~\figurename~\ref{fig:qualitative_results_all}, we present representative qualitative results on three scenes~(\textit{Parterre}, \textit{Coffee}, and \textit{Decoration}) from 2-view, 4-view, and 6-view settings, respectively.
The figure depicts the results of novel view synthesis, presenting rendered color and depth images.
The figures demonstrate that our model significantly enhances the visual quality of radiance fields in terms of geometric and perceptual fidelity.
In addition to the above quantitative results, qualitative results also demonstrate the inconsistency issue of pre-deblurred images and complex joint optimization of the DeRF from sparse view.
All the results (ii) of \figurename~\ref{fig:qualitative_results_all} demonstrate that existing representative regularization technique, RegNeRF~\cite{niemeyer2022regnerf}, can not effectively alleviate the optimization issue of the DeRF from sparse view.
Furthermore, we encourage readers to view the supplementary videos that emphasize 3D consistency through rendered videos from a spiral camera path.

\subsection{Ablations}
\label{subsec:abls}

\subsubsection{Problem Analysis and Motivation}
\label{subsubsec:abls_complex_optimization}
\begin{figure*}[t]
  \centering
  \includegraphics[width=\textwidth]{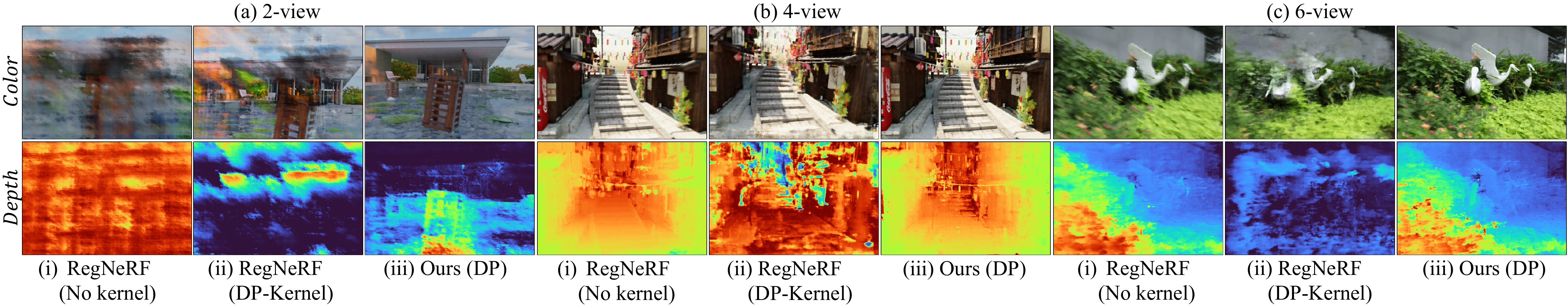}
  \vspace{-0.7cm}
  \caption{Qualitative results on \textit{Pool}, \textit{Tanabata}, and \textit{Heron} scene for complex optimization issue from 2-view, 4-view, and 6-view settings. (i), (ii), and (ii) denote the rendered images from the RegNeRF~\cite{niemeyer2022regnerf} with no blur kernel, RegNeRF with \textcolor[rgb]{0.13, 0.55, 0.13}{DP-kernel}, and Sparse-DeRF(Ours) with \textcolor[rgb]{0.13, 0.55, 0.13}{DP-kernel}, respectively.}
  \vspace{-0.2cm}
  \label{fig:ablation_complex_optimization_all}
\end{figure*}

we present the experimental results to describe that it is difficult to jointly optimize the DeRF and naively apply the previous regularization technique of the NeRF to the DeRF from sparse view setting, utilizing the representative existing regularization method, RegNeRF~\cite{niemeyer2022regnerf}.
We attach the experimental results of the RegNeRF~\cite{niemeyer2022regnerf} from 2-view, 4-view, and 6-view settings with and without the blur kernel to demonstrate the difficulty of the complex joint optimization problem as we mentioned.
The results of the RegNeRF w/ and w/o \textcolor[rgb]{0.13, 0.55, 0.13}{DP-kernel} in \tablename~\ref{tab:average_quantitative_results} (5, 6, 8 row) and \figurename~\ref{fig:ablation_complex_optimization_all} present quantitative and qualitative evidence of complex optimization from sparse view setting and shows the outstanding performance of our model.
To help the reader compare the plausible appearance and dense geometry with the proposed Sparse-DeRF, we attach the rendered color and depth images of our model.
The results of the figures demonstrate that the integration of the blur kernel involves a straightforward difficulty in optimizing the high-frequency details and the overall scene geometry simultaneously, as indicated by the visual quality of the rendered color and depth images.
Although the presence of a blur kernel enables radiance fields to capture high-frequency details, it causes a geometric distortion with the overall wrong density mapping that resembles the fragmented structure of objects or appearance like the painted wall of near- or far-depth regions.
These results and analysis demonstrate that naively applying the regularization of the NeRF from sparse view to the DeRF is not effective.
In addition, the images (i) and (ii) in~\figurename~\ref{fig:ablation_complex_optimization_all} demonstrate that although the overall performance is better without the blur kernels, the model still faces difficulty in modeling high-frequency details.
This difficulty makes the necessity of the blur kernel optimization still important, leading to a blurry visual quality across the entire scene.
Therefore, motivated by the experimental analysis, we propose Sparse-DeRF, the novel regularization method for optimization of the blur kernel and radiance fields simultaneously.
The proposed Sparse-DeRF presents high-quality rendered images with dense geometry and detailed high-frequency texture as shown in~\figurename~\ref{fig:ablation_complex_optimization_all}.

\begin{table*}[t]
   \renewcommand\arraystretch{1.2}
   \footnotesize
   \begin{center}
   \caption{\textbf{Average} results of the ablation study for the comparison of the proposed geometric constraints and perceptual prior, denoted in the table as surface smoothness~(SS), modulated gradient scaling~(MGS), and perceptual distillation~(PD). The experiement is conducted on novel view synthesis for both synthetic and real scenes obtained from 2-view, 4-view, and 6-view settings. Each color shading represents the \colorbox{best!35}{best}, \colorbox{second!35}{second best}, and \colorbox{third!35}{third best} result for each experimental setting, respectively.}
      \vspace{-0.2cm}
      \resizebox{2\columnwidth}{!}{
		\setlength{\tabcolsep}{1pt}
         \begin{tabular}{c|ccc||ccc|ccc||ccc|ccc||ccc|ccc||}
			& & & & \multicolumn{6}{c}{[~2-view ]} & \multicolumn{6}{c}{[~4-view ]} & \multicolumn{6}{c}{[~6-view ]}\\		
			& \multirow{2}{*}{\textit{SS}} & \multirow{2}{*}{\textit{MGS}} & \multirow{2}{*}{\textit{PD}} & \multicolumn{3}{c}{\textcolor[rgb]{0.13, 0.55, 0.13}{DP-kernel}} & \multicolumn{3}{c}{\textcolor[rgb]{0.25, 0.0, 1.0}{DN-kernel}} & \multicolumn{3}{c}{\textcolor[rgb]{0.13, 0.55, 0.13}{DP-kernel}} & \multicolumn{3}{c}{\textcolor[rgb]{0.25, 0.0, 1.0}{DN-kernel}} & \multicolumn{3}{c}{\textcolor[rgb]{0.13, 0.55, 0.13}{DP-kernel}} & \multicolumn{3}{c}{\textcolor[rgb]{0.25, 0.0, 1.0}{DN-kernel}}\\
			& & & & ~PSNR($\uparrow$)~ & SSIM($\uparrow$)~ & LPIPS($\downarrow$)~ & ~PSNR($\uparrow$)~ & SSIM($\uparrow$)~ & LPIPS($\downarrow$)~ & ~PSNR($\uparrow$)~ & SSIM($\uparrow$)~ & LPIPS($\downarrow$)~ & ~PSNR($\uparrow$)~ & SSIM($\uparrow$)~ & LPIPS($\downarrow$)~ & ~PSNR($\uparrow$)~ & SSIM($\uparrow$)~ & LPIPS($\downarrow$)~ & ~PSNR($\uparrow$)~ & SSIM($\uparrow$)~ & LPIPS($\downarrow$)~\\
			\midrule
			\multirow{7}{*}{\rotatebox{90}{Synthetic Scene}} & & & & 15.06 & .2827 & .5389 & \cellcolor{third!35}15.14 & .2884 & .5330 & 19.84 & .5336 & .3075 & \cellcolor{third!35}19.99 & .5199 & \cellcolor{third!35}.3499  & 23.68 & .7036 & \cellcolor{second!35}.1998 & 23.12 & .6798 & .2386 \\
			& \ding{51} & & & 14.88 & .2772 & .5388 & 15.25 & .2828 & .5418 & 19.93 & .5411 & .3107 & 19.89 & .5201 & .3558 & \cellcolor{third!35}24.15 & \cellcolor{second!35}.7263 & \cellcolor{best!35}.1974 & \cellcolor{second!35}23.59 & .6795 & .2429 \\		
			& & \ding{51} & & \cellcolor{second!35}15.40 & \cellcolor{third!35}.2857 & \cellcolor{second!35}.5289 & 15.12 & \cellcolor{best!35}.3025 & \cellcolor{best!35}.5243 & \cellcolor{third!35}20.16 & \cellcolor{third!35}.5425 & \cellcolor{third!35}.3024 & 19.44 & \cellcolor{second!35}.5295 & .3562 & 23.83 & .7147 & .2072 & \cellcolor{best!35}24.33 & .6832 & .2406\\	
			& & & \ding{51} & 15.03 & .2796 & .5391 & 14.84 & .2702 & .5454 & 19.65 & .5237 & .3311 & 19.72 & .5104 & .3630 & 23.89 & .7150 & .2052 & 23.35 & \cellcolor{best!35}.6941 & \cellcolor{second!35}.2341 \\		
			& \ding{51} & & \ding{51} & 14.52 & .2647 & .5481 & 15.11 & .2889 & .5389 & \cellcolor{second!35}20.91 & \cellcolor{best!35}.5812 & \cellcolor{best!35}.2948 & \cellcolor{second!35}20.03 & \cellcolor{third!35}.5294 & \cellcolor{second!35}.3450 & \cellcolor{best!35}24.28 & \cellcolor{best!35}.7279 & \cellcolor{third!35}.2021 & \cellcolor{third!35}23.41 & \cellcolor{best!35}.6941 & \cellcolor{best!35}.2324 \\
			& \ding{51} & \ding{51} & & \cellcolor{best!35}15.50 & \cellcolor{best!35}.2954 & \cellcolor{best!35}.5215 & \cellcolor{best!35}15.57 & \cellcolor{second!35}.2979 & \cellcolor{third!35}.5319 & 19.65 & .5164 & .3334 & 19.96 & .5222 & .3547 & 23.53 & .7088 & .2153 & 22.91 & .6654 & .2560 \\	
			& \ding{51} & \ding{51} & \ding{51} & \cellcolor{third!35}15.35 & \cellcolor{second!35}.2904 & \cellcolor{third!35}.5242 & \cellcolor{second!35}15.52 & \cellcolor{third!35}.2966 & \cellcolor{second!35}.5291 & \cellcolor{best!35}21.05 & \cellcolor{second!35}.5776 & \cellcolor{second!35}.2975 & \cellcolor{best!35}20.57 & \cellcolor{best!35}.5565 & \cellcolor{best!35}.3354 & \cellcolor{second!35}24.27 & \cellcolor{third!35}.7255 & .2044 & 23.32 & \cellcolor{third!35}.6903 & \cellcolor{third!35}.2379\\
			\midrule
			\midrule
			\multirow{7}{*}{\rotatebox{90}{Real Scene}} & & & & 14.36 & .2506 & .5904 & 14.41 & .2506 & .5921 & 18.77 & .4582 & .4175 & 18.89 & .4761 & .4151 & 21.68 & .6137 & .2992 & 21.36 & .6003 & .3163\\	
			& \ding{51} & & & 14.26 & .2414 & .5933 & 14.33 & .2495 & .5937 & 19.04 & .4760 & .4008 & 18.96 & .4839 & .4125 & \cellcolor{third!35}22.10 & \cellcolor{third!35}.6214 & .2951 & 21.53 & .6044 & .3178 \\		
			& & \ding{51} & & \cellcolor{second!35}15.46 & \cellcolor{second!35}.3035 & \cellcolor{best!35}.5465 & \cellcolor{third!35}15.44 & \cellcolor{third!35}.3037 & \cellcolor{third!35}.5558 & \cellcolor{second!35}19.75 & \cellcolor{third!35}.4980 & \cellcolor{second!35}.3737 & 19.06 & .4907 & .4105 & 22.07 & .6212 & \cellcolor{second!35}.2889 & \cellcolor{third!35}21.76 & .6101 & \cellcolor{second!35}.3097 \\		
			& & & \ding{51} & 14.28 & .2490 & .5896 & 14.47 & .2549 & .5939 & 19.00 & .4724 & .4080 & \cellcolor{third!35}19.84 & \cellcolor{third!35}.5118 & \cellcolor{third!35}.3937 & 21.73 & .6082 & .3082 & 21.69 & .6087 & .3142 \\		
			& \ding{51} & & \ding{51} & 14.00 & .2587 & .5830 & 14.47 & .2563 & .5881 & 19.00 & .4760 & .4037 & 18.94 & .4861 & .4145 & 21.84 & .6105 & .3084 & 21.70 & \cellcolor{third!35}.6129 & .3144 \\	
			& \ding{51} & \ding{51} & & \cellcolor{third!35}15.40 & \cellcolor{third!35}.3009 & \cellcolor{third!35}.5527 & \cellcolor{second!35}15.46 & \cellcolor{second!35}.3073 & \cellcolor{best!35}.5519 & \cellcolor{third!35}19.74 & \cellcolor{second!35}.4986 & \cellcolor{third!35}.3774 & \cellcolor{second!35}19.93 & \cellcolor{second!35}.5188 & \cellcolor{second!35}.3873 & \cellcolor{second!35}22.22 & \cellcolor{second!35}.6257 & \cellcolor{best!35}.2868 & \cellcolor{second!35}21.88 & \cellcolor{second!35}.6154 & \cellcolor{third!35}.3112 \\		
			& \ding{51} & \ding{51} & \ding{51} & \cellcolor{best!35}15.57 & \cellcolor{best!35}.3114 & \cellcolor{second!35}.5467 & \cellcolor{best!35}15.53 & \cellcolor{best!35}.3112 & \cellcolor{second!35}.5515 & \cellcolor{best!35}20.05 & \cellcolor{best!35}.5178 & \cellcolor{best!35}.3736 & \cellcolor{best!35}19.98 & \cellcolor{best!35}.5231 & \cellcolor{best!35}.3871 & \cellcolor{best!35}22.32 & \cellcolor{best!35}.6283 & \cellcolor{third!35}.2907 & \cellcolor{best!35}22.15 & \cellcolor{best!35}.6248 & \cellcolor{best!35}.3030 \\
			\midrule 
      \end{tabular}
      }
   \vspace{-0.5cm}
   \label{tab:abl_quantitative_comprehensive}
   \end{center}
\end{table*}

\begin{figure*}[t]
  \centering
  \includegraphics[width=\textwidth]{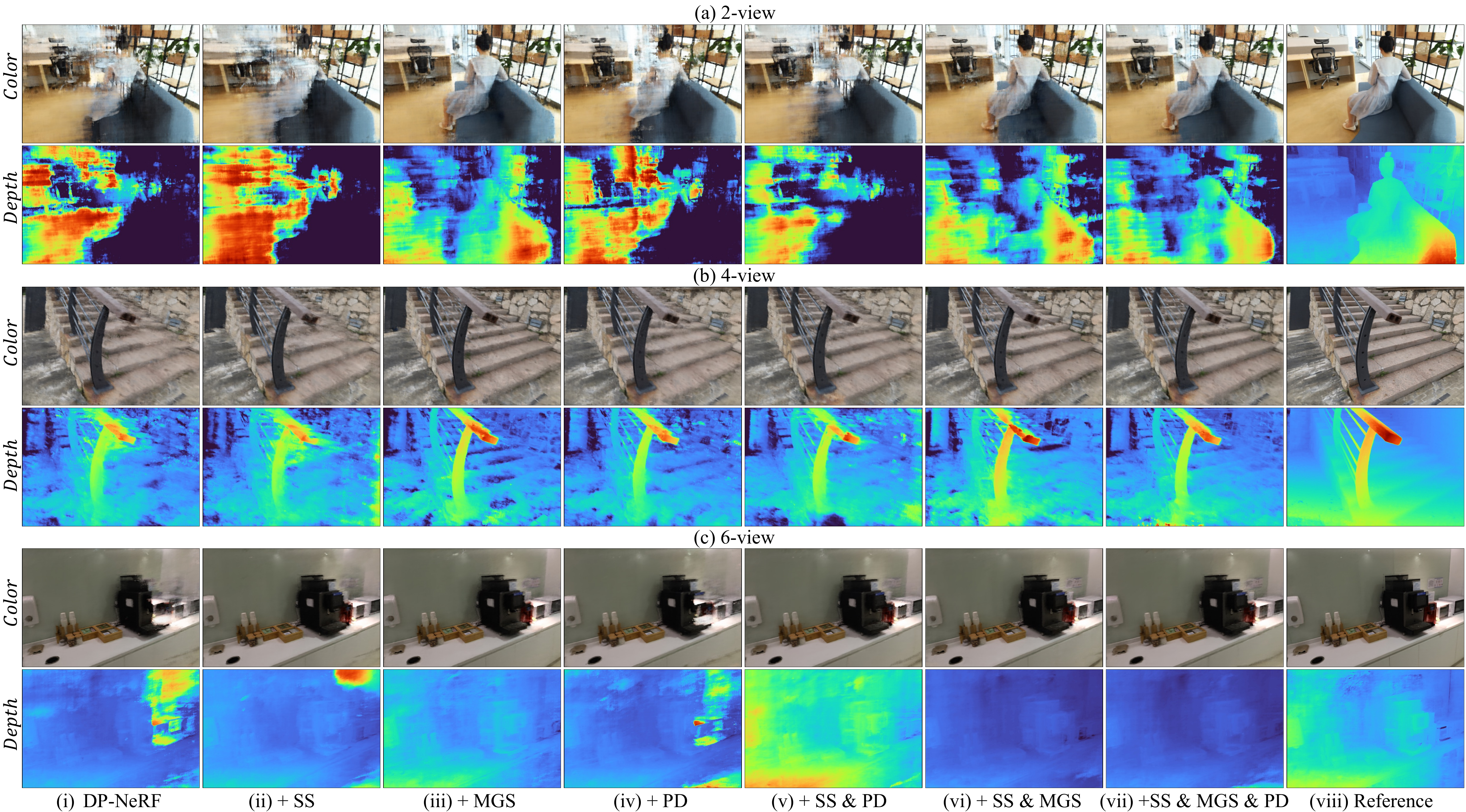}
  \vspace{-0.7cm}
  \caption{Qualitative ablation results of each component on \textit{Girl}, \textit{Stair}, and \textit{Coffee} scenes from 2-view, 4-view, and 6-view settings, respectively. We attach the rendered depth images of the DP-NeRF~\cite{lee2023dpnerf} trained from full view as reference depth images to help the reader compare the results.}
  \vspace{-0.4cm}
  \label{fig:abl_quantitative_comprehensive}
\end{figure*}

\subsubsection{Ablation of Each Component}
\label{subsubsec:abls_effectiveness_each_component}
We demonstrate the effectiveness of the Sparse-DeRF's each component through comprehensive ablation studies, presenting both quantitative and qualitative results.
In~\tablename~\ref{tab:abl_quantitative_comprehensive} and~\figurename~\ref{fig:abl_quantitative_comprehensive}, we present independent quantitative and qualitative results from SS, MGS, and PD, where each component is individually applied to observe the influence of each method.
The results include several combinations of the components to reveal the complement effect of each proposed method.
The experiments are conducted for entire 2-view, 4-view, and 6-view settings, providing a different performance depending on two types of blur kernels, \textcolor[rgb]{0.25, 0.0, 1.0}{DN-kernel}~\cite{ma2022deblurnerf} and \textcolor[rgb]{0.13, 0.55, 0.13}{DP-kernel}~\cite{lee2023dpnerf}.
Our model shows superior results in predicting both 3D geometric and appearance details precisely, demonstrating the enhanced evaluation results.

\subsubsection{Ablation Analysis}
\label{subsubsec:abls_analysis}
Quantitative ablation results demonstrate that our model with full components shows the best results in the real-scene dataset.
The results reveal that MGS plays an important role across the entire 2-view, 4-view, and 6-view settings among geometric constraints although each constraint enhances the geometric accuracy.
In~\figurename~\ref{fig:abl_quantitative_comprehensive}, the importance of MGS is more clearly prominent with qualitative results as color and depth images.
If MGS is not applied, we can see that the geometry of the scene is not accurately captured or there are many density artifacts.
On the other hand, PD seems to be not effective without geometric constraints in 2-view and 4-view settings as we can figure out in~\tablename~\ref{tab:abl_quantitative_comprehensive} and~\figurename~\ref{fig:abl_quantitative_comprehensive}.
The reason is that the NeRF has more difficulty in predicting the correct geometry with only pre-deblurred images due to the inherent 3D inconsistency, which makes perceptual prior not effective.
These difficulties become more severe as the number of views decreases.
We can demonstrate that a certain level of accurate geometry should be achieved before applying perceptual distillation.

\begin{table*}
   \footnotesize
   \begin{center}
      \caption{\textbf{Average} results between naive gradient scaling of~\cite{philip2023floatersnomore} and our MGS. The experiements are conducted on both synthetic and real scenes obtained from 2-view, 4-view, and 6-view settings. Color shading represents the \colorbox{best!35}{better} result.}
      \vspace{-0.4cm}
      \renewcommand\arraystretch{1.2}
      \resizebox{\linewidth}{!}{
		\setlength{\tabcolsep}{1pt}
         \begin{tabular}{c|c||ccc|ccc||ccc|ccc||ccc|ccc}
	        \multicolumn{2}{c}{} & \multicolumn{6}{c}{[~2-view ]} & \multicolumn{6}{c}{[~4-view ]} & \multicolumn{6}{c}{[~6-view ]}\\
			\multicolumn{2}{c}{} & \multicolumn{3}{c}{[~Synthetic Scene~]} & \multicolumn{3}{c}{[~Real Scene~]} & \multicolumn{3}{c}{[~Synthetic Scene~]} & \multicolumn{3}{c}{[~Real Scene~]} & \multicolumn{3}{c}{[~Synthetic Scene~]} & \multicolumn{3}{c}{[~Real Scene~]}\\
			Blur Kernel & Gradient Scaling & ~PSNR($\uparrow$)~ & SSIM($\uparrow$)~ & LPIPS($\downarrow$)~& ~PSNR($\uparrow$)~ & SSIM($\uparrow$)~ & LPIPS($\downarrow$)~ & ~PSNR($\uparrow$)~ & SSIM($\uparrow$)~ & LPIPS($\downarrow$)~& ~PSNR($\uparrow$)~ & SSIM($\uparrow$)~ & LPIPS($\downarrow$)~ & ~PSNR($\uparrow$)~ & SSIM($\uparrow$)~ & LPIPS($\downarrow$)~& ~PSNR($\uparrow$)~ & SSIM($\uparrow$)~ & LPIPS($\downarrow$)~ \\
			\midrule
			\textcolor[rgb]{0.25, 0.0, 1.0}{DN-kernel} & + Naive~\cite{philip2023floatersnomore} & 14.70 & .2483 & .5643 & 14.85 & .2860 & .5653 & 18.69 & .4321 & .4242 & 19.28 & .4909 & .4119 & 22.00 & .6002 & .3117 & 21.20 & .5877 & .3274 \\
			\textcolor[rgb]{0.25, 0.0, 1.0}{DN-kernel} & + MGS & \cellcolor{best!35}15.12 & \cellcolor{best!35}.3025 & \cellcolor{best!35}.5243 & \cellcolor{best!35}15.44 & \cellcolor{best!35}.3037 & \cellcolor{best!35}.5558 & \cellcolor{best!35}19.44 & \cellcolor{best!35}.5295 & \cellcolor{best!35}.3562 & \cellcolor{best!35}19.84 & \cellcolor{best!35}.5118 & \cellcolor{best!35}.3937 & \cellcolor{best!35}22.40 & \cellcolor{best!35}.6832 & \cellcolor{best!35}.2406 & \cellcolor{best!35}21.76 & \cellcolor{best!35}.6101 & \cellcolor{best!35}.3097 \\
			\midrule
			\textcolor[rgb]{0.13, 0.55, 0.13}{DP-kernel} & + Naive~\cite{philip2023floatersnomore} & 14.01 & .2377 & .5743 & 14.95 & .2871 & .5591 & 17.25 & .3933 & .4221 & 18.85 & .4670 & .4014 & 21.55 & .6097 & .2838 & 21.49 & .6010 & .3111 \\
			\textcolor[rgb]{0.13, 0.55, 0.13}{DP-kernel} & + MGS & \cellcolor{best!35}15.40 & \cellcolor{best!35}.2857 & \cellcolor{best!35}.5289 & \cellcolor{best!35}15.46 & \cellcolor{best!35}.3035 & \cellcolor{best!35}.5465 & \cellcolor{best!35}20.16 & \cellcolor{best!35}.5425 & \cellcolor{best!35}.3024 & \cellcolor{best!35}19.75 & \cellcolor{best!35}.5980 & \cellcolor{best!35}.3737 & \cellcolor{best!35}23.83 & \cellcolor{best!35}.7147 & \cellcolor{best!35}.2072 & \cellcolor{best!35}22.07 & \cellcolor{best!35}.6212 & \cellcolor{best!35}.2889 \\
			\midrule
      \end{tabular}
      }
   \vspace{-0.4cm}
   \label{tab:abls_comp_ngs_mgs}
   \end{center}
\end{table*}

\begin{figure*}[t]
  \centering
  \includegraphics[width=\textwidth]{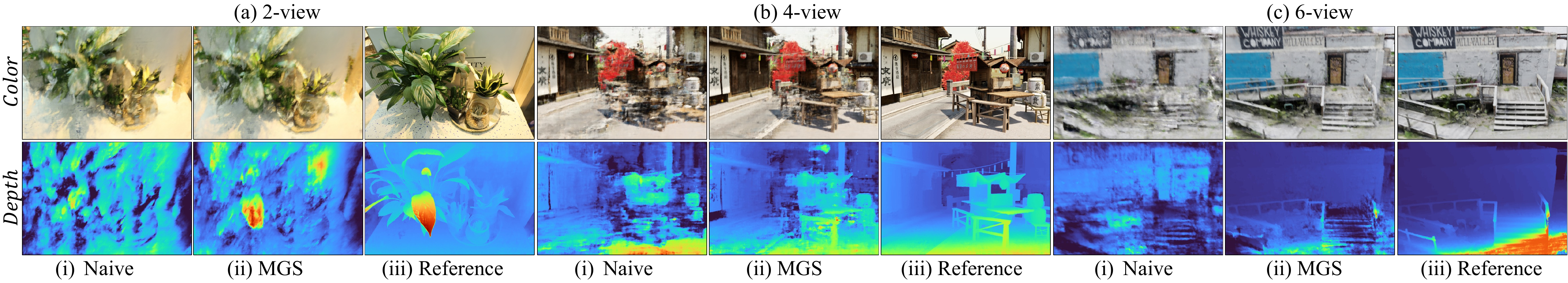}
  \vspace{-0.6cm}
  \caption{Qualitative results on \textit{Puppet}, \textit{Trolley}, and \textit{Factory} scene for comparison between naive gradient scaling~\cite{philip2023floatersnomore} and our proposed MGS from 2-view, 4-view, and 6-view settings. Note that we attach the rendered color and depth images from DP-NeRF~\cite{lee2023dpnerf} trained with full view as reference images.}
  \vspace{-0.3cm}
  \label{fig:abls_comp_ngs_mgs}
\end{figure*}

\begin{table*}
   \footnotesize
   \begin{center}
      \caption{\textbf{Average} results of the comparison with the model that uses depth-prior. The experiments are conducted on both synthetic and real scenes obtained from 2-view, 4-view, and 6-view settings. Each color shading represents the \colorbox{best!35}{best}, \colorbox{second!35}{second best} results.}
      \vspace{-0.4cm}
      \renewcommand\arraystretch{1.2}
      \resizebox{\linewidth}{!}{
		\setlength{\tabcolsep}{1pt}
         \begin{tabular}{c||ccc|ccc||ccc|ccc||ccc|ccc||}
	        & \multicolumn{6}{c}{[~2-view ]} & \multicolumn{6}{c}{[~4-view ]} & \multicolumn{6}{c}{[~6-view ]}\\
			& \multicolumn{3}{c}{[~Synthetic Scene~]} & \multicolumn{3}{c}{[~Real Scene~]} & \multicolumn{3}{c}{[~Synthetic Scene~]} & \multicolumn{3}{c}{[~Real Scene~]} & \multicolumn{3}{c}{[~Synthetic Scene~]} & \multicolumn{3}{c}{[~Real Scene~]}\\
			& ~PSNR($\uparrow$)~ & SSIM($\uparrow$)~ & LPIPS($\downarrow$)~& ~PSNR($\uparrow$)~ & SSIM($\uparrow$)~ & LPIPS($\downarrow$)~ & ~PSNR($\uparrow$)~ & SSIM($\uparrow$)~ & LPIPS($\downarrow$)~& ~PSNR($\uparrow$)~ & SSIM($\uparrow$)~ & LPIPS($\downarrow$)~ & ~PSNR($\uparrow$)~ & SSIM($\uparrow$)~ & LPIPS($\downarrow$)~& ~PSNR($\uparrow$)~ & SSIM($\uparrow$)~ & LPIPS($\downarrow$)~ \\
			\midrule	
			SparseNeRF~\cite{wang2023sparsenerf} & \cellcolor{best!35}18.12 & \cellcolor{best!35}.4729 & \cellcolor{best!35}.4884 & \cellcolor{best!35}18.01 & \cellcolor{best!35}.4683 & \cellcolor{best!35}.5245 & \cellcolor{best!35}21.08 & \cellcolor{best!35}.6105 & .4042 & \cellcolor{best!35}20.69 & \cellcolor{best!35}.5909 & .4450 & 21.77 & .6369 & .3903 & 21.28 & .6189 & .4250\\	
			Sparse-DeRF~(w/\textcolor[rgb]{0.25, 0.0, 1.0}{DN-kernel})~-~Ours & \cellcolor{second!35}15.52 & \cellcolor{second!35}.2966 & .5291 & 15.53 & .3112 & .5515 & 20.57 & .5565 & \cellcolor{second!35}.3354 & 19.98 & .5231 & \cellcolor{second!35}.3871 & \cellcolor{second!35}23.32 & \cellcolor{second!35}.6903& \cellcolor{second!35}.2379 & \cellcolor{second!35}22.15 & \cellcolor{second!35}.6248 & \cellcolor{second!35}.3030 \\
			Sparse-DeRF(w/\textcolor[rgb]{0.13, 0.55, 0.13}{DP-kernel})~-~Ours & 15.35 & .2904 & \cellcolor{second!35}.5242 & \cellcolor{second!35}15.57 & \cellcolor{second!35}.3114 & \cellcolor{second!35}.5467  & \cellcolor{second!35}21.05 & \cellcolor{second!35}.5776 & \cellcolor{best!35}.2975  & \cellcolor{second!35}20.05 & \cellcolor{second!35}.5178 & \cellcolor{best!35}.3736 & \cellcolor{best!35}24.27 & \cellcolor{best!35}.7255 & \cellcolor{best!35}.2044 & \cellcolor{best!35}22.32 & \cellcolor{best!35}.6283 & \cellcolor{best!35}.2907 \\
			\midrule
      \end{tabular}
      }
   \vspace{-0.5cm}
   \label{tab:abls_depth_prior}
   \end{center}
\end{table*}

\begin{figure*}[t]
  \centering
  \includegraphics[width=\textwidth]{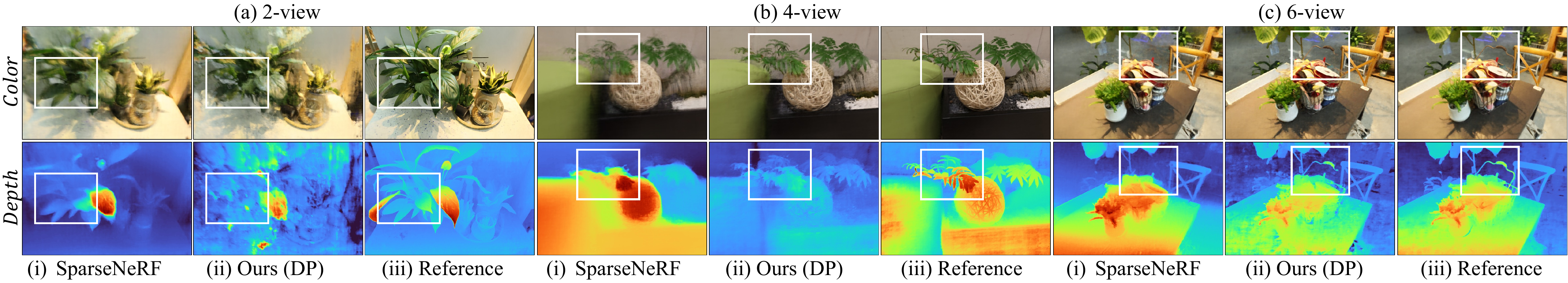}
  \vspace{-0.6cm}
  \caption{Qualitative results on \textit{Puppet}, \textit{Ball}, and \textit{Basket} scene from 2-view, 4-view, and 6-view settings with the SparseNeRF~\cite{wang2023sparsenerf} that use depth-prior. Even with inaccurate depth priors, it can be helpful in capturing rough geometry in extreme scenarios like a 2-view setting. However, when using 4 or 6 images, the multi-view inconsistency of the predicted depth prior leads to rather inaccurate geometry predictions, especially in thin or detailed structures. Note that, reference image is rendered from DP-NeRF~\cite{lee2023dpnerf} trained with full-view images for comparison.}
  \vspace{-0.2cm}
  \label{fig:abls_depth_prior}
\end{figure*}

\begin{table*}
   \footnotesize
   \begin{center}
      \caption{\textbf{Average} results of the comparison depending on usage of the pre-deblurred images on various models. The experiements are conducted on both synthetic and real scenes obtained from 2-view, 4-view, and 6-view settings. Each color shading represents the \colorbox{best!35}{best}, \colorbox{second!35}{second}, and \colorbox{third!35}{third} result for each experimental setting, respectively.}
      \vspace{-0.4cm}
      \renewcommand\arraystretch{1.2}
      \resizebox{\linewidth}{!}{
		\setlength{\tabcolsep}{1pt}
         \begin{tabular}{c||ccc|ccc||ccc|ccc||ccc|ccc||}
	        & \multicolumn{6}{c}{[~2-view ]} & \multicolumn{6}{c}{[~4-view ]} & \multicolumn{6}{c}{[~6-view ]}\\
			& \multicolumn{3}{c}{[~Synthetic Scene~]} & \multicolumn{3}{c}{[~Real Scene~]} & \multicolumn{3}{c}{[~Synthetic Scene~]} & \multicolumn{3}{c}{[~Real Scene~]} & \multicolumn{3}{c}{[~Synthetic Scene~]} & \multicolumn{3}{c}{[~Real Scene~]}\\
			& ~PSNR($\uparrow$)~ & SSIM($\uparrow$)~ & LPIPS($\downarrow$)~& ~PSNR($\uparrow$)~ & SSIM($\uparrow$)~ & LPIPS($\downarrow$)~ & ~PSNR($\uparrow$)~ & SSIM($\uparrow$)~ & LPIPS($\downarrow$)~& ~PSNR($\uparrow$)~ & SSIM($\uparrow$)~ & LPIPS($\downarrow$)~ & ~PSNR($\uparrow$)~ & SSIM($\uparrow$)~ & LPIPS($\downarrow$)~& ~PSNR($\uparrow$)~ & SSIM($\uparrow$)~ & LPIPS($\downarrow$)~ \\
			\midrule
			Naive NeRF~\cite{mildenhall2021nerf}~ & 15.11 & .2999 & .5578 &  14.38 & . 2635 &  .6004 &   20.02 & .5327 & .4000 &   18.98 &   .4860 &  .4481 & 21.65 & .5985 & .3638 &  20.63 &  .5513 & .4014  \\
			NeRF~+~MPR~\cite{zamir2021MPR} &   15.16 &   \cellcolor{third!35}.3006 &  .5595 &  14.38 & .2594 & .6019 & 20.00 &   .5381 &  .3956 & 18.89 & .4829 & .4484 &  21.72 &  .5999 &  .3629 & 20.60 &  .5513 &  .4010 \\		
			\midrule
			RegNeRF~\cite{niemeyer2022regnerf} & 14.60 & .2849 & .5869 & 15.49 & .2997 & .5888 & 18.39 & .4600 & .4704 & 18.44 & .4326 & .4852 & 19.69 & .5249 & .4165 &  19.65 &  .4790 &  .4583\\	
			RegNeRF~\cite{niemeyer2022regnerf} + MPR~\cite{zamir2021MPR} &  14.99 &  .2980 &  .5690 &  \cellcolor{third!35} 15.84 &   .3040 &   .5818 &  18.54 &  .4678 &  .4593 &  18.66 &  .4425 &  .4804 &  19.89 &  .5339 &  .4071 & 19.64 & .4753 & .4597\\	
			\midrule
			SparseNeRF~\cite{wang2023sparsenerf} &  \cellcolor{second!35}18.12 &  \cellcolor{second!35}.4729 &  \cellcolor{second!35}.4884 &   \cellcolor{best!35}18.01 &   \cellcolor{best!35}.4683 &\cellcolor{second!35}  .5245 &  \cellcolor{second!35}21.08 &  \cellcolor{second!35}.6105 & .4042 &  \cellcolor{second!35}20.69 &  \cellcolor{second!35}.5909 & .4450 & 21.77 & .6369 & .3903 & 21.28 & .6189 & .4250\\
			SparseNeRF~\cite{wang2023sparsenerf} + MPR~\cite{zamir2021MPR} &  \cellcolor{best!35}18.14 &  \cellcolor{best!35}.4850 &  \cellcolor{best!35}.4531 & \cellcolor{second!35}17.65 & \cellcolor{second!35}.4501 &  \cellcolor{best!35}.5126 &  \cellcolor{best!35}21.67 &  \cellcolor{best!35}.6623 &  \cellcolor{third!35}.3381 &  \cellcolor{best!35}20.88 &   \cellcolor{best!35}.6165 &  \cellcolor{third!35}.4132 &  \cellcolor{third!35}22.47 &  \cellcolor{second!35}.6995 &  \cellcolor{third!35}.3156 &  \cellcolor{third!35}21.53 &  \cellcolor{best!35}.6479 &  \cellcolor{third!35}.3935\\ 
			\midrule
			Sparse-DeRF~(w/\textcolor[rgb]{0.25, 0.0, 1.0}{DN-kernel})~-~Ours &  \cellcolor{third!35}15.52 &  .2966 &  .5291 &  15.53 &  .3112 &  .5515 &  20.57 &  .5565 &  \cellcolor{second!35}.3354 &  19.98 &  \cellcolor{third!35}.5231 &  \cellcolor{second!35}.3871 & \cellcolor{second!35} 23.32 &  \cellcolor{third!35}.6903&  \cellcolor{second!35}.2379 &  \cellcolor{second!35}22.15 &  \cellcolor{third!35}.6248 &  \cellcolor{second!35}.3030 \\
			Sparse-DeRF(w/\textcolor[rgb]{0.13, 0.55, 0.13}{DP-kernel})~-~Ours &  15.35 & .2904 &  \cellcolor{third!35}.5242 &  15.57 &  \cellcolor{third!35}.3114 &  \cellcolor{third!35}.5467  &  \cellcolor{third!35}21.05 &  \cellcolor{third!35}.5776 &  \cellcolor{best!35}.2975  & \cellcolor{third!35} 20.05 &  .5178 &  \cellcolor{best!35}.3736 &  \cellcolor{best!35}24.27 &  \cellcolor{best!35}.7255 &  \cellcolor{best!35}.2044 &  \cellcolor{best!35}22.32 &  \cellcolor{second!35}.6283 &  \cellcolor{best!35}.2907 \\
			\midrule
      \end{tabular}
      }
   \vspace{-0.5cm}
   \label{tab:abls_mpr_added}
   \end{center}
\end{table*}

\begin{figure*}[t]
  \centering
  \includegraphics[width=\textwidth]{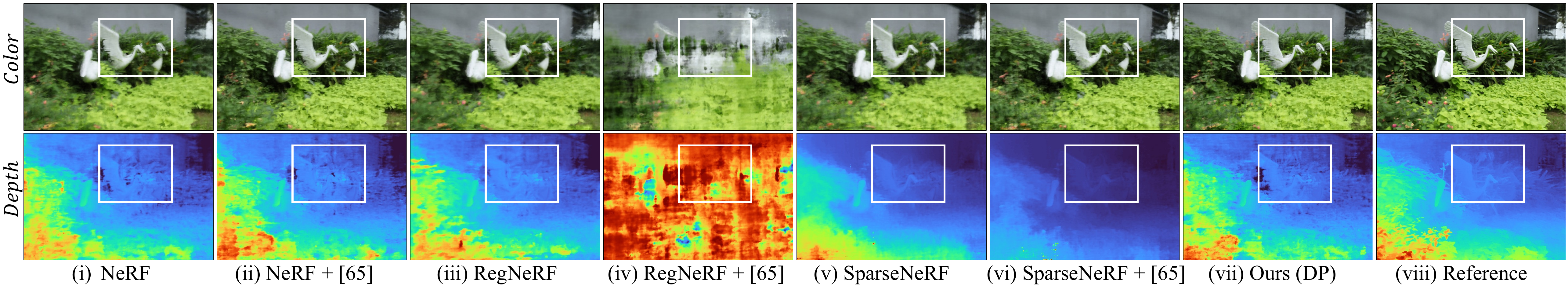}
  \vspace{-0.6cm}
  \caption{Qualitative results on \textit{Heron} scene from 4-view setting for the comparison of NeRF~\cite{mildenhall2021nerf}, RegNeRF~\cite{niemeyer2022regnerf} the SparseNeRF~\cite{wang2023sparsenerf}, and ours depending on the pre-deblurred images by using MPRNet~\cite{zamir2021MPR}. Note that, reference image is rendered from DP-NeRF~\cite{lee2023dpnerf} trained with full-view images for comparison.}
  \vspace{-0.2cm}
  \label{fig:ablation_inconsis_mpr}
\end{figure*}

\subsubsection{Comparison to Naive Gradient Scaling}
\label{subsubsec:abls_comp_ngs}
We present experimental results of quantitative and qualitative comparison between our proposed MGS and naive gradient scaling of~\cite{philip2023floatersnomore} from 2-view, 4-view, and 6-view settings in~\tablename~\ref{tab:abls_comp_ngs_mgs} and~\figurename~\ref{fig:abls_comp_ngs_mgs}.
We compare the effectiveness of our MGS for two types of kernels we utilize in our paper, which are the kernels of Deblur-NeRF~\cite{ma2022deblurnerf} and DP-NeRF~\cite{lee2023dpnerf}.
The results describe our MGS outperforms the naive gradient scaling in terms of both quantitative and qualitative performance across the entire experimental setting.
Specifically, depth images demonstrate that our MGS more effectively helps the model to predict the accurate geometry than naive gradient scaling.
Quantitative results for the entire scene are attached in the appendix.
Moreover, since experiments on the DeblurNeRF~\cite{ma2022deblurnerf} dataset were conducted in NDC by default, we adopted hyperparameters of MGS to be optimized for NDC. 
However, since MGS provides geometric guidance for the scene, it naturally works in other coordinate systems as well.
To validate this, we also included experimental results in the appendix demonstrating its effectiveness in the world coordinate system.

\subsubsection{Challenges in Utilizing Depth Priors}
\label{subsubsec:abls_inaccurate_depth}
We further address this inconsistency issue in relation to depth priors.
The different blur in each image lead to accurate and multi-view inconsistent depth estimation, which is not proper to be used as additional prior for predict radiance fields with clean geometric structure.
In~\tablename~\ref{tab:abls_depth_prior} and~\figurename~\ref{fig:abls_depth_prior}, we present the quantitative and qualitative results of the proposed Sparse-DeRF and SparseNeRF~\cite{wang2023sparsenerf}, which is the representative depth-prior based model.
The results show that as the number of images increases, the improvement diminishes and scene geometry is distorted, revealing the issue of inconsistent depth priors.
In extreme cases like a 2-view setting, where it is challenging to obtain even rough geometry information, even inaccurate depth maps can be somewhat helpful. 
In addition, since only two images are used, the inconsistency issue is less pronounced, resulting in relatively better structured geometry.
However, in 4- or 6-view settings, relying on inconsistent depth prior can make the prediction of a clean and accurate scene geometry even more difficult, particularly for thin or detailed structures as we can observe in the emphasized region in~\figurename~\ref{fig:abls_depth_prior}.
In particular, the qualitative experimental results with 4 views further support the inconsistency issue, as the significant difference in perceptual quality, reflected by LPIPS, provides additional evidence.
It is once again clear that qualitative evaluation and LPIPS performance are crucial in the deblurring task, while PSNR and SSIM are important, as mentioned in DP-NeRF~\cite{lee2023dpnerf}.
Furthermore, in the appendix, we included the experimental results of SparseNeRF~\cite{wang2023sparsenerf} with depth maps predicted by Depth-Anything V2~\cite{yang2024depthanythingv2}, one of the state-of-the-art monocular depth estimation models, which also exhibits the experimental tendency and multi-view inconsistency as well.
Therefore, to design a general sparse view model that works well across 2-, 4-, and 6-views, we adopted an experimental setup that does not utilize depth priors.
Quantitative results for the entire scene are attached in the appendix.

\subsubsection{Inconsistency of Pre-deblurred Images}
\label{subsubsec:abls_inconssitency}
\begin{figure*}[t]
  \centering
  \includegraphics[width=\linewidth]{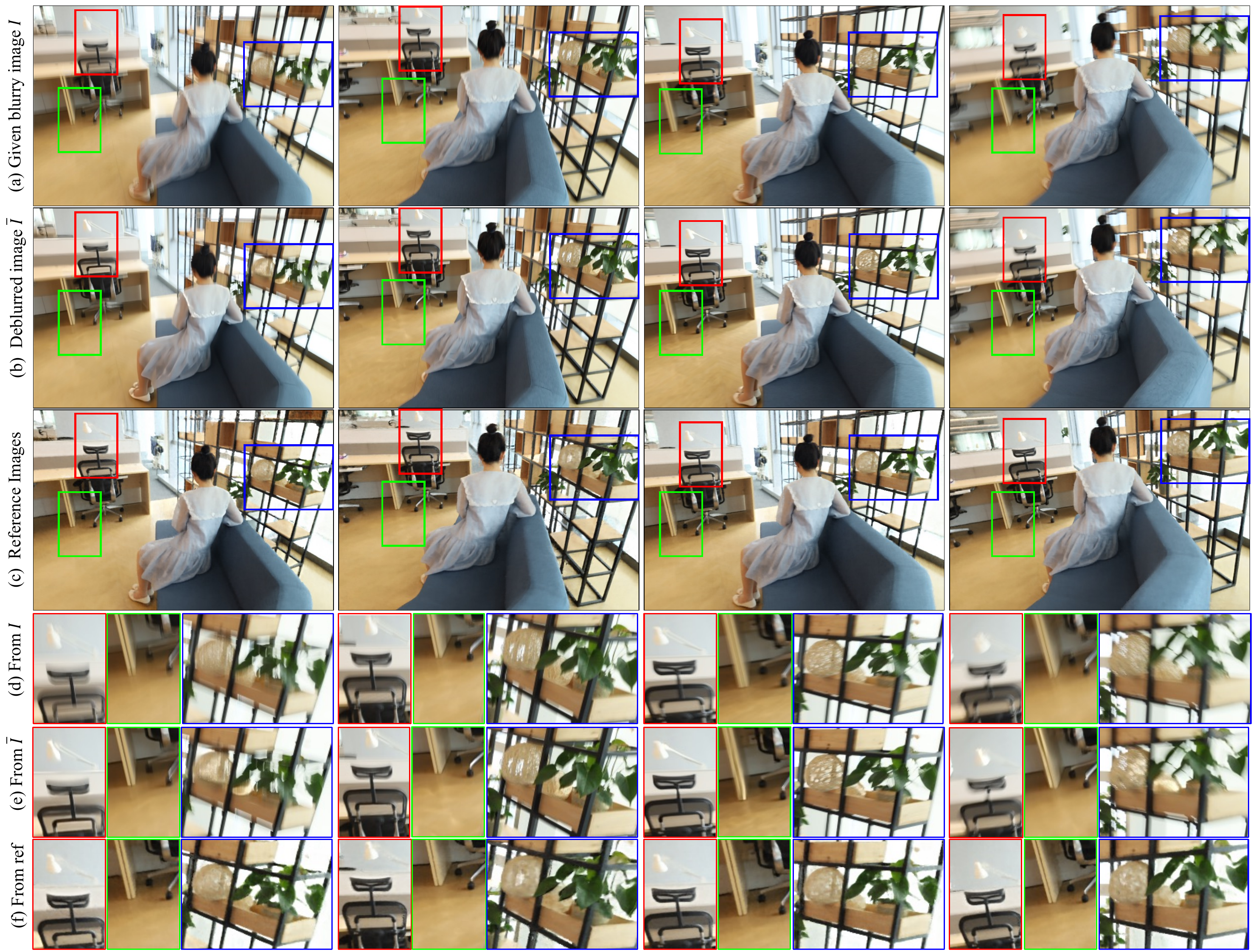}
  \vspace{-0.5cm}
  \caption{Qualitative comparison on \textit{Girl} scene that demonstrates the geometric inconsistency of pre-deblurred images.~\figurename~(a)$\sim$(c) presents input blurry images $I$ for training, pre-deblurred images $\bar{I}$ by MPRNet~\cite{zamir2021MPR}, and rendered images from DP-NeRF~\cite{lee2023dpnerf} trained with full view, respectively.~\figurename~(d)$\sim$(f) shows the emphasized regions of~\figurename~(a)$\sim$(c), which demonstrates the inconsistency issue in detail.}
  \vspace{-0.3cm}
  \label{fig:inconsis_predeblurre_images}
\end{figure*}

Image deblurring is conducted independently for each image and presents inconsistent geometry in various regions due to its ill-posed property. 
In \figurename~\ref{fig:inconsis_predeblurre_images}, we present the qualitative comparison to reveal the inconsistency of pre-deblurred images across the multi-view training images.
Pre-deblurred images are acquired by applying the MPRNet~\cite{zamir2021MPR}.
In addition, we attach the reference images, which are rendered from the DP-NeRF~\cite{lee2023dpnerf} trained with full view, to help readers better understand the inconsistency issue and compare the pre-deblurred geometry to approximated ground truth geometry.
Comparing emphasized regions from each image, the pre-deblurred image shows relatively well-restored textures in each image, but the geometry is inconsistently restored and distorted across the multiple views. 
Such inconsistency adversely affects the learning of radiance fields, which is crucial for perceptual quality, since it is trained by pixel-wise color reconstruction loss.
In~\tablename~\ref{tab:abls_mpr_added}, we present the comparison results of the NeRF~\cite{mildenhall2021nerf}, RegNeRF~\cite{niemeyer2022regnerf}, and SparseNeRF~\cite{wang2023sparsenerf} with and without the pre-deblurred images.
The results from 4-view and 6-view also demonstrate that, it is not enough to reconstruct the clean radiance fields with high perceptual quality only depending on the pre-deblurring and depth estimation although they provide the abundant additional priors.
To qualitatively demonstrate the observation, we included~\figurename~\ref{fig:ablation_inconsis_mpr}, which compares the results trained on 4 view setting. 
While SparseNeRF~\cite{wang2023sparsenerf} + MPR~\cite{zamir2021MPR}, which utilizes both pre-deblurred images and depth prior, achieved higher PSNR and SSIM scores numerically, the rendered results clearly show that our approach produces much sharper and more precise renderings.
Due to these reasons, directly utilizing pre-deblurred images for training radiance fields is difficult. 
Therefore, perceptual distillation is introduced to transfer only the perceptual texture of pre-deblurred images to the radiance fields.

%% file: sections/limitations_discussions.tex
\section{Limitations and Discussions}
\label{sec:limitations_discussion}
Despite the remarkable enhancement in terms of 3D geometry and appearance, there are several points for improvement.

The first one is derived from the blur kernel itself, especially the relationship between the type of the blur kernel and the properties of each scene.
For example, the performance is more improved with the rigid blur kernel of DP-NeRF~\cite{lee2023dpnerf} in some scenes, but in other scenes, the improvements are greater with the flexible blur kernel of Deblur-NeRF~\cite{ma2022deblurnerf}.
These kernel-dependent performances are different across the scenes. 
As we figure out, a flexible kernel leads to reduced space ambiguity but high scene distortion. 
In contrast, the rigid kernel leads to accurate geometry but suffers difficulty in optimizing the scene where the distances of the objects from the camera in the scene are diverse and some objects are located very close to the camera due to the inherent rigidity.
We tried to take advantage of both kernels and design the hybrid kernel to maximize the effectiveness of the Sparse-DeRF, but it didn't work as we imagined.
Constructing the hybrid blur kernel that has rigid and flexible properties can be a promising future research direction regardless of sparse view setting in the deblurred neural radiance fields~(DeRF).

The second one is that we have to set the proper hyper-parameter for MGS to find the most effective function shape although MGS greatly improves the 3D geometry in the DeRF from sparse view.
However, it is difficult to find an ideal function shape according to the arrangement of the object in the scene.
We handle these cases by setting the magnitude $\rho$ as a high value to only ignore the gradient in a very near distance region, which is attached to the appendix as detailed hyper-parameters per each scene.
In this sense, this manual setting of hyperparameters is regarded as one of our limitations.
We believe the limitation can be alleviated in future research through various methods such as the introduction of learnable parameters for gradient scaling function.

Finally, although sparse view setting of blurry inputs is an extremely practical scenario for blurry inputs, it is too hard to enhance the performance of the DeRF in the 2-view setting due to the lack of the scene information included in the input data.
The innate challenge of the 2-view setting is that sparse overlapped 3D space usually leads to inaccurate geometry, which is more likely to be mapped to be painted texture on the wall at the near or far depth regions.
There is still room to improve the visual quality of the Sparse-DeRF and solve these problems through state-of-the-art generative methods similar to the work like Diffusionerf~\cite{wynn2023diffusionerf}, which can be a great future direction for constructing the DeRF from sparse view.
On the other hand, designing models that account for inconsistency to utilize additional data-deriven priors, such as depth and normal priors from pretrained models, could be an interesting direction for future work. 
Furthermore, leveraging real-time rendering models, like the recently emerging 3D Gaussian Splatting~\cite{kerbl20233d}, could enhance practical applicability.

%% file: sections/conclusion.tex
\section{Conclusion}
\label{sec:conclusion}

In this work, we propose the Sparse-DeRF, a novel regularization method for high-quality deblurred neural radiance fields from sparse view settings, which considers more practical real-world scenarios for radiance fields from only blurry images.
We propose two geometric constraints that consist of surface smoothness and modulated gradient scaling, which reflect the real-world statistical geometry and alleviate elongated density artifacts in deblurred neural radiance fields system from sparse view.
In addition, we propose a perceptual distillation to utilize the pre-deblurred images as a perceptual prior, which enhances the sharp texture on deblurred neural radiance fields.
We demonstrate the effectiveness of the Sparse-DeRF that ameliorates the spatial ambiguity and structural distortion of deblurred neural radiance fields by presenting extensive experimental results in 2-view, 4-view, and 6-view settings.
As deblurred neural radiance fields have attracted attention across the research fields related to neural rendering, we believe our work presents a way for future research directions since we address the more practical scenarios for deblurred neural radiance fields from blurry images.

%% file: sections/appendix_sample_space.tex
\begin{figure*}[ht]
  \centering
  \includegraphics[width=\linewidth]{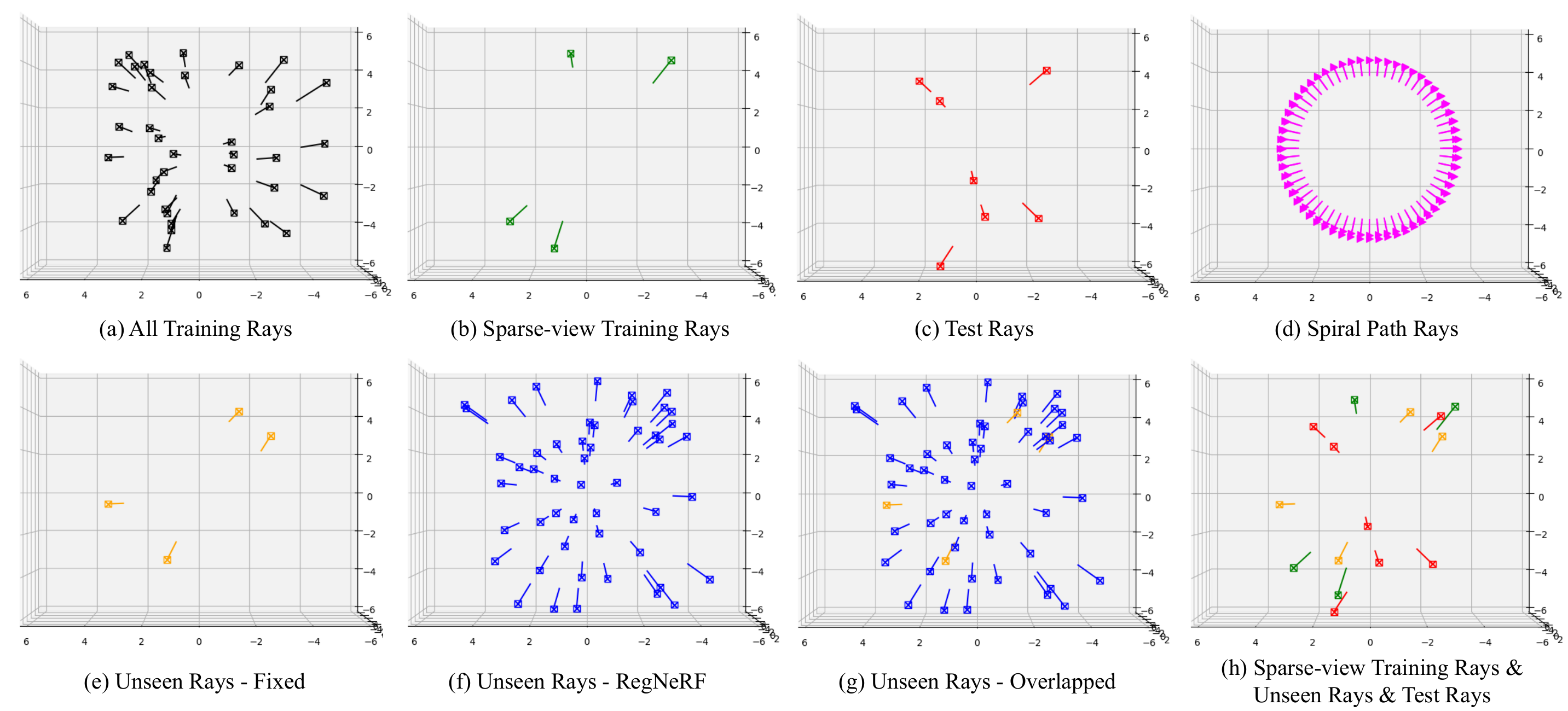}
  \vspace{-0.5cm}
  \caption{Sample space visualization for selection of unseen rays on \textit{Basket} scene from 4-view setting. \textcolor{orange}{Orange rays} and \textcolor{blue}{blue rays} indicate the unseen rays of two versions, which are fixed rays for the experiment in our paper and randomly extracted from sample space of the RegNeRF~\cite{niemeyer2022regnerf}, respectively. The number of unseen rays used in our paper is fixed and set to $4$ and we extract the $50$ rays from the sample space of the RegNeRF to represent the coverage of the sample space roughly. Black rays, \textcolor[rgb]{0.13, 0.55, 0.13}{green rays}, and \textcolor{red}{red rays} indicate all training rays, training rays from sparse-view, and test rays for novel view synthesis evaluation, respectively. \textcolor[rgb]{1, 0.41, 0.71}{Pink rays} indicate spiral path rays which are utilized as supplementary videos to evaluate the 3D consistency in most of the NeRF-related works.}
  \label{fig:sample_space_vis}
\end{figure*}

\section{Sample Space Visualization}
\label{appendix_sec:sample_space_vis}

In~\figurename~\ref{fig:sample_space_vis}, we present the visualization of the sample space of the unseen rays from the RegNeRF~\cite{niemeyer2022regnerf} and our fixed unseen rays, which is used for fair comparison in the paper.
Since they randomly sample the camera poses from the sample space in every training, we sample the $50$ unseen camera poses from the sample space to show the approximate coverage of the sample space.
As we can see in the~\figurename~\ref{fig:sample_space_vis}, the fixed unseen rays, which are used in our experiments still in the coverage of the sample space of the RegNeRF.
In addition, training camera poses and fixed unseen training rays do not significantly overlap with test rays, which also does not break the training and testing rule for novel view synthesis.
Hence, it is not a problem to use the fixed unseen rays as alternative unseen rays of the RegNeRF.
As we mentioned in Section~V-C in the main paper, we utilize the fixed unseen rays for training our model to fairly evaluate the performances across the extensive experiments since the randomness of unseen ray generation in the RegNeRF makes it hard to understand the effectiveness of each component.

%% file: sections/appendix_inconsistency_of_estimated_depth.tex
\newcolumntype{L}[1]{>{\raggedright\let\newline\\\arraybackslash\hspace{0pt}}m{#1}}
\newcolumntype{C}[1]{>{\centering\let\newline\\\arraybackslash\hspace{0pt}}m{#1}}
\newcolumntype{R}[1]{>{\raggedleft\let\newline\\\arraybackslash\hspace{0pt}}m{#1}}

\begin{table*}
   \footnotesize
   \begin{center}
      \caption{\textbf{Average} results of the comparison with the model that uses depth-prior. The experiments are conducted on both synthetic and real scenes obtained from 2-view, 4-view, and 6-view settings. Each color shading represents the \colorbox{best!35}{best}, \colorbox{second!35}{second best}, and \colorbox{third!35}{third best} results, respectively.}
      \vspace{-0.4cm}
      \renewcommand\arraystretch{1.2}
      \resizebox{\linewidth}{!}{
		\setlength{\tabcolsep}{1pt}
         \begin{tabular}{c||ccc|ccc||ccc|ccc||ccc|ccc||}
	        & \multicolumn{6}{c}{[~2-view ]} & \multicolumn{6}{c}{[~4-view ]} & \multicolumn{6}{c}{[~6-view ]}\\
			& \multicolumn{3}{c}{[~Synthetic Scene~]} & \multicolumn{3}{c}{[~Real Scene~]} & \multicolumn{3}{c}{[~Synthetic Scene~]} & \multicolumn{3}{c}{[~Real Scene~]} & \multicolumn{3}{c}{[~Synthetic Scene~]} & \multicolumn{3}{c}{[~Real Scene~]}\\
			& ~PSNR($\uparrow$)~ & SSIM($\uparrow$)~ & LPIPS($\downarrow$)~& ~PSNR($\uparrow$)~ & SSIM($\uparrow$)~ & LPIPS($\downarrow$)~ & ~PSNR($\uparrow$)~ & SSIM($\uparrow$)~ & LPIPS($\downarrow$)~& ~PSNR($\uparrow$)~ & SSIM($\uparrow$)~ & LPIPS($\downarrow$)~ & ~PSNR($\uparrow$)~ & SSIM($\uparrow$)~ & LPIPS($\downarrow$)~& ~PSNR($\uparrow$)~ & SSIM($\uparrow$)~ & LPIPS($\downarrow$)~ \\
			\midrule	
			SparseNeRF~\cite{wang2023sparsenerf} & \cellcolor{second!35}18.12 & \cellcolor{second!35}.4729 & \cellcolor{second!35}.4884 & \cellcolor{best!35}18.01 & \cellcolor{best!35}.4683 & \cellcolor{second!35}.5245 & \cellcolor{second!35}21.07 & \cellcolor{second!35}.6105 & .4042 & \cellcolor{best!35}20.69 & \cellcolor{best!35}.5909 & .4450 & 21.77 & .6369 & .3903 & \cellcolor{third!35}21.28 & \cellcolor{third!35}.6189 & .4250\\
			SparseNeRF~\cite{wang2023sparsenerf} + Depth-Any V2~\cite{yang2024depthanythingv2} & \cellcolor{best!35}18.24 & \cellcolor{best!35}.4909 & \cellcolor{best!35}.4479 & \cellcolor{second!35}16.35 & \cellcolor{second!35}.4262 & \cellcolor{best!35}.4768 & \cellcolor{best!35}21.78 & \cellcolor{best!35}.6679 & \cellcolor{second!35}.3354 & 18.87 & \cellcolor{second!35}.5406 & \cellcolor{third!35}.4016 & \cellcolor{third!35}22.56 & \cellcolor{second!35}.7034 & \cellcolor{third!35}.3141 & 19.44 & .5659 & \cellcolor{third!35}.3845 \\	
			Sparse-DeRF~(w/\textcolor[rgb]{0.25, 0.0, 1.0}{DN-kernel})~-~Ours & \cellcolor{third!35}15.52 & \cellcolor{third!35}.2966 & .5291 & 15.53 & .3112 & .5515 & 20.57 & .5565 & \cellcolor{second!35}.3354 & \cellcolor{third!35}19.98 & \cellcolor{third!35}.5231 & \cellcolor{second!35}.3871 & \cellcolor{second!35}23.32 & \cellcolor{third!35}.6903 & \cellcolor{second!35}.2379 & \cellcolor{second!35}22.15 & \cellcolor{second!35}.6248 & \cellcolor{second!35}.3030 \\
			Sparse-DeRF(w/\textcolor[rgb]{0.13, 0.55, 0.13}{DP-kernel})~-~Ours & 15.35 & .2904 & \cellcolor{third!35}.5242 & \cellcolor{third!35}15.57 & \cellcolor{third!35}.3114 & \cellcolor{third!35}.5467  & \cellcolor{third!35}21.05 & \cellcolor{third!35}.5776 & \cellcolor{best!35}.2975  & \cellcolor{second!35}20.05 & .5178 & \cellcolor{best!35}.3736 & \cellcolor{best!35}24.27 & \cellcolor{best!35}.7255 & \cellcolor{best!35}.2044 & \cellcolor{best!35}22.32 & \cellcolor{best!35}.6283 & \cellcolor{best!35}.2907 \\
			\midrule
      \end{tabular}
      }
   \vspace{-0.5cm}
   \label{tab:abls_depth_prior_depth_anything_avg}
   \end{center}
\end{table*}

\begin{table*}[t]
   \Large
   \begin{center}
   \caption{Quantitative results for the entire scenes of real scenes from \textbf{2-view} settings. Each color shading represents the \colorbox{best!35}{best}, \colorbox{second!35}{second best}, and \colorbox{third!35}{third best} result, respectively. Depth-Any V2 denotes the Depth-Anything V2~\cite{yang2024depthanythingv2}.}
      \resizebox{2\columnwidth}{!}{
		\centering
		\setlength{\tabcolsep}{1pt}
         \begin{tabular}{c||ccc|ccc|ccc|ccc|ccc||ccc}
			\midrule		
			\multirow{2}{*}{Synthetic Scene}& \multicolumn{3}{c}{Factory} & \multicolumn{3}{c}{Cozyroom} & \multicolumn{3}{c}{Pool} & \multicolumn{3}{c}{Tanabata} & \multicolumn{3}{c}{Trolley} & \multicolumn{3}{c}{Average} \\
			 & PSNR($\uparrow$) & SSIM($\uparrow$) & LPIPS($\downarrow$) & PSNR($\uparrow$) & SSIM($\uparrow$) & LPIPS($\downarrow$) & PSNR($\uparrow$) & SSIM($\uparrow$) & LPIPS($\downarrow$) & PSNR($\uparrow$) & SSIM($\uparrow$) & LPIPS($\downarrow$) & PSNR($\uparrow$) & SSIM($\uparrow$) & LPIPS($\downarrow$) & PSNR($\uparrow$) & SSIM($\uparrow$) & LPIPS($\downarrow$) \\
			\midrule
			SparseNeRF~\cite{wang2023sparsenerf} & \cellcolor{second!35} 15.80 &   \cellcolor{second!35}.3622 &   \cellcolor{second!35}.5988 & \cellcolor{second!35}  22.06 &   \cellcolor{second!35}.7001 &  \cellcolor{third!35} .3419 & \cellcolor{best!35}  23.90 &  \cellcolor{best!35} .6063 &  \cellcolor{best!35} .3824 &  \cellcolor{best!35} 14.53 & \cellcolor{best!35} .3454 &  \cellcolor{second!35} .5651 & \cellcolor{second!35}14.32 & \cellcolor{second!35}.3499 & \cellcolor{second!35}.5651 & \cellcolor{second!35}18.12 &   \cellcolor{second!35}.4728 &   \cellcolor{second!35}.4883  \\
			SparseNeRF~\cite{wang2023sparsenerf} + Depth-Any V2~\cite{yang2024depthanythingv2} &  \cellcolor{best!35}15.93 & \cellcolor{best!35}.3923 & \cellcolor{best!35}.5060 & \cellcolor{best!35}23.02 & \cellcolor{best!35}.7419 & \cellcolor{best!35}.2972 & \cellcolor{second!35}23.64 & \cellcolor{second!35}.5968 & \cellcolor{second!35}.3957 & \cellcolor{second!35}13.94 & \cellcolor{second!35}.3086 & \cellcolor{best!35}.5469 & \cellcolor{best!35}14.71 & \cellcolor{best!35}.4150 & \cellcolor{best!35}.4938 &\cellcolor{best!35} 18.24 & \cellcolor{best!35}.4909 & \cellcolor{best!35}.4479\\
			Sparse-DeRF~(w/\textcolor[rgb]{0.25, 0.0, 1.0}{DN-kernel})~-~Ours &    \cellcolor{third!35}14.27 &  \cellcolor{third!35} .2200 &  .6564 &   \cellcolor{third!35}20.57 &  \cellcolor{third!35} .5675 &  .3589 &   19.81 & .3330 &   .4058 &   11.67 &    .1827 &   .5947 &   11.28 &    \cellcolor{third!35}.1800 &   \cellcolor{third!35} .6298 &   \cellcolor{third!35} 15.52 &   \cellcolor{third!35} .2966 &    .5291 \\
			Sparse-DeRF~(w/\textcolor[rgb]{0.13, 0.55, 0.13}{DP-kernel})~-~Ours &  14.10 &  .2116 &   \cellcolor{third!35} .6413 &  18.97 &  .5206 &  \cellcolor{second!35} .3417 &   \cellcolor{third!35} 20.32 &   \cellcolor{third!35} .3488 &   \cellcolor{third!35} .4056 &   \cellcolor{third!35} 12.00 &   \cellcolor{third!35}.1943 &   \cellcolor{third!35} .5902 &   \cellcolor{third!35} 11.36 &   .1769 &   .6422 &   15.35 &  .2904 &  \cellcolor{third!35} .5242 \\
			\midrule\midrule
	
			\multirow{2}{*}{Real Scene}& \multicolumn{3}{c}{Ball} & \multicolumn{3}{c}{Basket} & \multicolumn{3}{c}{Buick} & \multicolumn{3}{c}{Coffee} & \multicolumn{3}{c}{Decoration} & & \\
			 & PSNR($\uparrow$) & SSIM($\uparrow$) & LPIPS($\downarrow$) & PSNR($\uparrow$) & SSIM($\uparrow$) & LPIPS($\downarrow$) & PSNR($\uparrow$) & SSIM($\uparrow$) & LPIPS($\downarrow$) & PSNR($\uparrow$) & SSIM($\uparrow$) & LPIPS($\downarrow$) & PSNR($\uparrow$) & SSIM($\uparrow$) & LPIPS($\downarrow$) &  & &\\
			\midrule
			SparseNeRF~\cite{wang2023sparsenerf} &  \cellcolor{best!35}21.62 &  \cellcolor{best!35} .5588 &  \cellcolor{best!35} .5092 &  \cellcolor{second!35} 16.32 &   \cellcolor{second!35}.4636 &  \cellcolor{second!35} .5488 &  \cellcolor{best!35} 18.32 &   \cellcolor{best!35}.5356 &  \cellcolor{second!35} .4754 &  \cellcolor{best!35} 24.82 &  \cellcolor{best!35}.8225 &  \cellcolor{best!35} .2958 &  \cellcolor{second!35} 14.85 &  \cellcolor{second!35} .3325 &   \cellcolor{second!35}.6020 &  &  & \\
			SparseNeRF~\cite{wang2023sparsenerf} + Depth-Any V2~\cite{yang2024depthanythingv2} & \cellcolor{second!35}21.58 & \cellcolor{second!35}.5573 & \cellcolor{second!35}.5107 & \cellcolor{best!35}16.44 & \cellcolor{best!35}.4698 & \cellcolor{best!35}.5462 & \cellcolor{second!35}18.28 & \cellcolor{second!35}.5352 & \cellcolor{best!35}.4752 & \cellcolor{second!35}24.65 & \cellcolor{second!35}.8195 & \cellcolor{second!35}.3003 & \cellcolor{best!35}15.19 & \cellcolor{best!35}.3432 & \cellcolor{best!35}.5941 \\
			Sparse-DeRF~(w/\textcolor[rgb]{0.25, 0.0, 1.0}{DN-kernel})~-~Ours &   20.07 &   \cellcolor{third!35} .4612 &   .5290 &   13.62 &   .2678 &   .5835 &  \cellcolor{third!35}  14.38 &   \cellcolor{third!35} .3182 &   \cellcolor{third!35} .4913 &   19.51 &   .6048 &   .4146 &   13.09 &   .2173 &   .6256 &   &   &  \\
			Sparse-DeRF~(w/\textcolor[rgb]{0.13, 0.55, 0.13}{DP-kernel})~-~Ours &   \cellcolor{third!35} 20.09 &   .4583 &    \cellcolor{third!35} .5244 &  \cellcolor{third!35}  14.05 &  \cellcolor{third!35}  .2890 &   \cellcolor{third!35} .5555 &  13.78 &   .2864 &   .5036 &    \cellcolor{third!35}19.67 &   \cellcolor{third!35} .6084 &   \cellcolor{third!35} .4096 &   \cellcolor{third!35} 13.11 &  \cellcolor{third!35}  .2192 &   \cellcolor{third!35} .6192 &   &   &  \\
			\midrule
			\multirow{2}{*}{Real Scene}& \multicolumn{3}{c}{Girl} & \multicolumn{3}{c}{Heron} & \multicolumn{3}{c}{Parterre} & \multicolumn{3}{c}{Puppet} & \multicolumn{3}{c}{Stair} & \multicolumn{3}{c}{Average} \\
			 & PSNR($\uparrow$) & SSIM($\uparrow$) & LPIPS($\downarrow$) & PSNR($\uparrow$) & SSIM($\uparrow$) & LPIPS($\downarrow$) & PSNR($\uparrow$) & SSIM($\uparrow$) & LPIPS($\downarrow$) & PSNR($\uparrow$) & SSIM($\uparrow$) & LPIPS($\downarrow$) & PSNR($\uparrow$) & SSIM($\uparrow$) & LPIPS($\downarrow$) & PSNR($\uparrow$) & SSIM($\uparrow$) & LPIPS($\downarrow$) \\
			\midrule
			SparseNeRF~\cite{wang2023sparsenerf} &  \cellcolor{second!35} 14.16 &  \cellcolor{second!35} .4380 &  .5641 & \cellcolor{second!35}  16.81 &  \cellcolor{second!35} .3257 &  \cellcolor{third!35}.5566 &  \cellcolor{best!35} 17.58 &  \cellcolor{best!35} .3601 &  \cellcolor{third!35}.5928 & \cellcolor{best!35}  16.89 &  \cellcolor{second!35} .4677 &  \cellcolor{second!35} .5261 & \cellcolor{best!35} 18.75 & \cellcolor{best!35} .3784 &  \cellcolor{best!35} .5738 & \cellcolor{best!35} 18.01 &  \cellcolor{best!35} .4683 &   \cellcolor{second!35}.5245 \\
			SparseNeRF~\cite{wang2023sparsenerf} + Depth-Any V2~\cite{yang2024depthanythingv2} & \cellcolor{best!35}14.65 & \cellcolor{best!35}.4600 & \cellcolor{second!35}.5513 & \cellcolor{best!35}16.90 & \cellcolor{best!35}.3258 & .5597 & \cellcolor{second!35}17.25 & \cellcolor{second!35}.3481 & .5983 & \cellcolor{second!35}16.66 & \cellcolor{best!35}.4684 & \cellcolor{best!35}.5234 &\cellcolor{second!35} 18.33 & \cellcolor{second!35}.3614 & \cellcolor{second!35}.5861 & \cellcolor{second!35}16.35 & \cellcolor{second!35}.4262 & \cellcolor{best!35}.4768 \\
			Sparse-DeRF~(w/\textcolor[rgb]{0.25, 0.0, 1.0}{DN-kernel})~-~Ours &   12.85 &   .3418 &    \cellcolor{third!35}.5585 &    14.30 &    .1706 & \cellcolor{best!35}.5197 &    16.61 &    .2903 &    \cellcolor{second!35} .5912 &   14.91 &   .3141 &   .5886 &   15.97 &   .1255 &   \cellcolor{third!35} .6125 &   15.53 &   .3112 &   .5515 \\
			Sparse-DeRF~(w/\textcolor[rgb]{0.13, 0.55, 0.13}{DP-kernel})~-~Ours &   \cellcolor{third!35} 13.11 &   \cellcolor{third!35} .3707 &    \cellcolor{best!35}.5486 &  \cellcolor{third!35} 14.17 &  \cellcolor{third!35} .1669 &   \cellcolor{second!35} .5285 &  \cellcolor{third!35} 16.19 &  \cellcolor{third!35}.2527 &  \cellcolor{best!35} .5908 &  \cellcolor{third!35}  15.25 & \cellcolor{third!35}   .3277 &   \cellcolor{third!35}  .5739 &   \cellcolor{third!35} 16.30 &   \cellcolor{third!35} .1343 &   .6133 &   \cellcolor{third!35} 15.57 &    \cellcolor{third!35}.3114 &    \cellcolor{third!35}.5467 \\
			\midrule
      \end{tabular}
      }
   \label{tab:appendix_quant_results_depth_prior_2view}
   \end{center}
\end{table*}

\begin{table*}[t]
   \Large
   \begin{center}
   \caption{Quantitative results for the entire scenes of real scenes from \textbf{4-view} settings. Each color shading represents the \colorbox{best!35}{best}, \colorbox{second!35}{second best}, and \colorbox{third!35}{third best} result, respectively. Depth-Any V2 denotes the Depth-Anything V2~\cite{yang2024depthanythingv2}.}
      \resizebox{2\columnwidth}{!}{
		\centering
		\setlength{\tabcolsep}{1pt}
         \begin{tabular}{c||ccc|ccc|ccc|ccc|ccc||ccc}
			\midrule
			\multirow{2}{*}{Synthetic Scene}& \multicolumn{3}{c}{Factory} & \multicolumn{3}{c}{Cozyroom} & \multicolumn{3}{c}{Pool} & \multicolumn{3}{c}{Tanabata} & \multicolumn{3}{c}{Trolley} & \multicolumn{3}{c}{Average} \\
			 & PSNR($\uparrow$) & SSIM($\uparrow$) & LPIPS($\downarrow$) & PSNR($\uparrow$) & SSIM($\uparrow$) & LPIPS($\downarrow$) & PSNR($\uparrow$) & SSIM($\uparrow$) & LPIPS($\downarrow$) & PSNR($\uparrow$) & SSIM($\uparrow$) & LPIPS($\downarrow$) & PSNR($\uparrow$) & SSIM($\uparrow$) & LPIPS($\downarrow$) & PSNR($\uparrow$) & SSIM($\uparrow$) & LPIPS($\downarrow$) \\
			\midrule
			SparseNeRF~\cite{wang2023sparsenerf} &  \cellcolor{third!35} 17.15 &   \cellcolor{third!35} .4271 &   .5516 &   23.50 &   .7531 &   .2994 &   \cellcolor{second!35}26.63 &  \cellcolor{best!35} .7091 &   .3114 &  \cellcolor{best!35} 18.93 &  \cellcolor{second!35}.5838 &   .4389 &  \cellcolor{second!35} 19.14 &   \cellcolor{second!35}.5865 &   .4195 & \cellcolor{second!35}21.07 & \cellcolor{second!35}.6105 & .4042 \\
			SparseNeRF~\cite{wang2023sparsenerf} + Depth-Any V2~\cite{yang2024depthanythingv2} & \cellcolor{second!35}18.93 & \cellcolor{best!35}.5813 & \cellcolor{best!35}.3959 & \cellcolor{third!35}24.86 & \cellcolor{second!35}.8044 & \cellcolor{third!35}.2335 & \cellcolor{best!35}26.74 & \cellcolor{second!35}.7047 & \cellcolor{third!35}.2988 & \cellcolor{second!35}18.90 & \cellcolor{best!35}.6111 & \cellcolor{third!35}.3955 & \cellcolor{best!35}19.46 & \cellcolor{best!35}.6379 & \cellcolor{third!35}.3482 & \cellcolor{best!35}21.78 & \cellcolor{best!35}.6679 & \cellcolor{second!35}.3354 \\
			Sparse-DeRF~(w/\textcolor[rgb]{0.25, 0.0, 1.0}{DN-kernel})~-~Ours &  16.91 &  .3693 &  \cellcolor{third!35} .5357 &  \cellcolor{second!35}  25.79 &  \cellcolor{third!35}  .7872 &   \cellcolor{second!35} .1801 &   \cellcolor{third!35} 25.48 &   \cellcolor{third!35} .6042 &   \cellcolor{best!35} .2405 &  16.58 &  .4705 &   \cellcolor{second!35} .3793 &  18.10 &   .5514 &    \cellcolor{second!35}.3414 &   20.57 &   .5565 &    \cellcolor{second!35}.3354 \\
			Sparse-DeRF~(w/\textcolor[rgb]{0.13, 0.55, 0.13}{DP-kernel})~-~Ours & \cellcolor{best!35}  18.99 &  \cellcolor{second!35} .4770 &   \cellcolor{second!35}.4034 &   \cellcolor{best!35}26.51 &  \cellcolor{best!35} .8051 &  \cellcolor{best!35} .1627 &  23.33 &  .4888 &  \cellcolor{second!35}.2733 &   \cellcolor{third!35} 17.51 &    \cellcolor{third!35}.5371 &  \cellcolor{best!35} .3356 &   \cellcolor{third!35} 18.92 &    \cellcolor{third!35}.5799 &  \cellcolor{best!35} .3126 &  \cellcolor{third!35}  21.05 &   \cellcolor{third!35} .5776 &  \cellcolor{best!35} .2975 \\
			\midrule\midrule			
			\multirow{2}{*}{Real Scene}& \multicolumn{3}{c}{Ball} & \multicolumn{3}{c}{Basket} & \multicolumn{3}{c}{Buick} & \multicolumn{3}{c}{Coffee} & \multicolumn{3}{c}{Decoration} & & \\
			 & PSNR($\uparrow$) & SSIM($\uparrow$) & LPIPS($\downarrow$) & PSNR($\uparrow$) & SSIM($\uparrow$) & LPIPS($\downarrow$) & PSNR($\uparrow$) & SSIM($\uparrow$) & LPIPS($\downarrow$) & PSNR($\uparrow$) & SSIM($\uparrow$) & LPIPS($\downarrow$) & PSNR($\uparrow$) & SSIM($\uparrow$) & LPIPS($\downarrow$) &  & &\\
			\midrule
			SparseNeRF~\cite{wang2023sparsenerf} & 21.56 & \cellcolor{third!35}.5539 & .5020 &  \cellcolor{second!35}20.45 &  \cellcolor{second!35}.5986 & .4620 & \cellcolor{second!35} 21.04 &  \cellcolor{second!35}.6392 & .4008 & \cellcolor{third!35}26.55 & \cellcolor{second!35} .8499 & .2739 &  \cellcolor{best!35}18.94 &  \cellcolor{best!35}.5128 & .5042 &  &  &  \\
			SparseNeRF~\cite{wang2023sparsenerf} + Depth-Any V2~\cite{yang2024depthanythingv2} &\cellcolor{third!35} 21.74 & \cellcolor{second!35}.5587 & \cellcolor{third!35}.4973 & \cellcolor{best!35}20.55 & \cellcolor{best!35}.6007 & \cellcolor{third!35}.4611 & \cellcolor{best!35}21.17 & \cellcolor{best!35}.6474 & \cellcolor{third!35}.3987 & 26.47 & \cellcolor{best!35}.8504 & \cellcolor{third!35}.2696 & \cellcolor{second!35}18.86 & \cellcolor{second!35}.5097 & \cellcolor{third!35}.5031 \\
			Sparse-DeRF~(w/\textcolor[rgb]{0.25, 0.0, 1.0}{DN-kernel})~-~Ours &    \cellcolor{second!35} 22.28 &     .5506 &   \cellcolor{second!35} .4334 &   18.69 &   .5058 &    \cellcolor{second!35} .3774 &  19.28 & .5414 &    \cellcolor{second!35} .3371 &  \cellcolor{second!35}  26.76 &   .8126 &     \cellcolor{second!35}.2419 &   \cellcolor{third!35} 17.12 &   \cellcolor{third!35} .4192 &   \cellcolor{best!35} .4761 &   &   &   \\
			Sparse-DeRF~(w/\textcolor[rgb]{0.13, 0.55, 0.13}{DP-kernel})~-~Ours &   \cellcolor{best!35} 23.39 &    \cellcolor{best!35}.6010 &    \cellcolor{best!35}.3806 &  \cellcolor{third!35}  20.41 &   \cellcolor{third!35} .5451 &   \cellcolor{best!35} .3305 &  \cellcolor{third!35}  19.48 &   \cellcolor{third!35} .5531 &   \cellcolor{best!35} .3251 &    \cellcolor{best!35}27.77 &   \cellcolor{third!35} .8364 &  \cellcolor{best!35}  .2049 &   16.33 &   .3914 &   \cellcolor{second!35}  .4950 &   &   &   \\
			\midrule
			\multirow{2}{*}{Real Scene}& \multicolumn{3}{c}{Girl} & \multicolumn{3}{c}{Heron} & \multicolumn{3}{c}{Parterre} & \multicolumn{3}{c}{Puppet} & \multicolumn{3}{c}{Stair} & \multicolumn{3}{c}{Average} \\
			 & PSNR($\uparrow$) & SSIM($\uparrow$) & LPIPS($\downarrow$) & PSNR($\uparrow$) & SSIM($\uparrow$) & LPIPS($\downarrow$) & PSNR($\uparrow$) & SSIM($\uparrow$) & LPIPS($\downarrow$) & PSNR($\uparrow$) & SSIM($\uparrow$) & LPIPS($\downarrow$) & PSNR($\uparrow$) & SSIM($\uparrow$) & LPIPS($\downarrow$) & PSNR($\uparrow$) & SSIM($\uparrow$) & LPIPS($\downarrow$) \\
			\midrule
			SparseNeRF~\cite{wang2023sparsenerf} &   \cellcolor{second!35}17.96 &   \cellcolor{second!35}.6579 &  .4445 &  \cellcolor{second!35}  19.08 &  \cellcolor{best!35} .4873 &  \cellcolor{third!35}.4793 &  \cellcolor{second!35} 20.91 &   \cellcolor{second!35}.5239 &  .4808 &  \cellcolor{second!35} 18.56 &  \cellcolor{second!35} .5409 &  .4444 &  \cellcolor{second!35} 21.83 &   \cellcolor{second!35}.5446 &  .4574 &   \cellcolor{best!35}20.69 &   \cellcolor{best!35}.5909 & .4450 \\
			SparseNeRF~\cite{wang2023sparsenerf} + Depth-Any V2~\cite{yang2024depthanythingv2} & \cellcolor{best!35}18.10 & \cellcolor{best!35}.6633 & \cellcolor{third!35}.4371 & 19.03 & \cellcolor{second!35}.4856 & .4807 & \cellcolor{best!35}21.07 & \cellcolor{best!35}.5276 & \cellcolor{third!35}.4777 & \cellcolor{best!35}18.67 & \cellcolor{best!35}.5506 & \cellcolor{third!35}.4378 & \cellcolor{best!35}21.99 & \cellcolor{best!35}.5530 & \cellcolor{third!35}.4545 & 18.87 & \cellcolor{second!35}.5406 & \cellcolor{third!35}.4016 \\
			Sparse-DeRF~(w/\textcolor[rgb]{0.25, 0.0, 1.0}{DN-kernel})~-~Ours &  \cellcolor{third!35}  16.95 &   \cellcolor{third!35} .5975 &  \cellcolor{best!35}  .3823 &  \cellcolor{third!35} 19.04 &   .4558 &    \cellcolor{second!35} .3315 &    \cellcolor{third!35} 20.74 &   \cellcolor{third!35} .4916 &   \cellcolor{best!35} .4323 &  \cellcolor{third!35}  18.11 &   \cellcolor{third!35} .4616 &    \cellcolor{best!35}.4202 &   20.80 &   .3948 &     \cellcolor{second!35}.4386 &  \cellcolor{third!35} 19.98 &   \cellcolor{third!35} .5231 &    \cellcolor{second!35} .3871 \\
			Sparse-DeRF~(w/\textcolor[rgb]{0.13, 0.55, 0.13}{DP-kernel})~-~Ours &   16.47 &   .5855 &    \cellcolor{second!35} .3835 &  \cellcolor{best!35}  19.09 &    \cellcolor{third!35}.4624 &  \cellcolor{best!35}  .3194 &  18.36 &  .3375 &   \cellcolor{second!35}  .4715 &  17.97 & .4612 &     \cellcolor{second!35}.4206 &   \cellcolor{third!35}  21.25 &   \cellcolor{third!35} .4045 &   \cellcolor{best!35} .4044 &  \cellcolor{second!35}  20.05 &   .5178 &    \cellcolor{best!35}.3736 \\
			\midrule
      \end{tabular}
      }
   \label{tab:appendix_quant_results_depth_prior_4view}
   \end{center}
\end{table*}

\begin{table*}[t]
   \Large
   \begin{center}
   \caption{Quantitative results for the entire scenes of real scenes from \textbf{6-view} settings. Each color shading represents the \colorbox{best!35}{best}, \colorbox{second!35}{second best}, and \colorbox{third!35}{third best} result, respectively. Depth-Any V2 denotes the Depth-Anything V2~\cite{yang2024depthanythingv2}.}
      \resizebox{2\columnwidth}{!}{
		\centering
		\setlength{\tabcolsep}{1pt}
         \begin{tabular}{c||ccc|ccc|ccc|ccc|ccc||ccc}
	        \midrule			
			\multirow{2}{*}{Synthetic Scene}& \multicolumn{3}{c}{Factory} & \multicolumn{3}{c}{Cozyroom} & \multicolumn{3}{c}{Pool} & \multicolumn{3}{c}{Tanabata} & \multicolumn{3}{c}{Trolley} & \multicolumn{3}{c}{Average} \\
			 & PSNR($\uparrow$) & SSIM($\uparrow$) & LPIPS($\downarrow$) & PSNR($\uparrow$) & SSIM($\uparrow$) & LPIPS($\downarrow$) & PSNR($\uparrow$) & SSIM($\uparrow$) & LPIPS($\downarrow$) & PSNR($\uparrow$) & SSIM($\uparrow$) & LPIPS($\downarrow$) & PSNR($\uparrow$) & SSIM($\uparrow$) & LPIPS($\downarrow$) & PSNR($\uparrow$) & SSIM($\uparrow$) & LPIPS($\downarrow$) \\
			\midrule
			SparseNeRF~\cite{wang2023sparsenerf} &  17.17 &   .4286 &   .5402 &   23.86 &   .7601 &   .3021 &   28.04 &   \cellcolor{second!35}.7612 &   .2772 &  \cellcolor{third!35} 19.99 &  .6216 &   .4212 &   19.78 &   .6131 &   .4106 & 21.77 & .6369 & .3903 \\
			SparseNeRF~\cite{wang2023sparsenerf} + Depth-Any V2~\cite{yang2024depthanythingv2} & \cellcolor{second!35}19.42 & \cellcolor{second!35}.6130 & \cellcolor{second!35}.3777 & \cellcolor{third!35}25.10 & \cellcolor{third!35}.8130 & \cellcolor{third!35}.2282 & \cellcolor{third!35}28.09 & \cellcolor{best!35}.7628 & \cellcolor{third!35}.2643 & 19.78 & \cellcolor{third!35}.6529 & \cellcolor{third!35}.3686 & \cellcolor{third!35}20.45 & \cellcolor{third!35}.6751 & \cellcolor{third!35}.3318 & \cellcolor{third!35}22.56 & \cellcolor{second!35}.7034 & \cellcolor{third!35}.3141 \\
			Sparse-DeRF~(w/\textcolor[rgb]{0.25, 0.0, 1.0}{DN-kernel})~-~Ours &  \cellcolor{third!35}18.33 &  \cellcolor{third!35}.5144 &  \cellcolor{third!35}.4004 &  \cellcolor{second!35}26.63 &  \cellcolor{second!35}.8184 &  \cellcolor{second!35}.1513 &  \cellcolor{second!35}28.16 & .7344 &   \cellcolor{second!35}.1755 &  \cellcolor{second!35} 22.06 &  \cellcolor{second!35}.6988 &  \cellcolor{second!35}.2245 &  \cellcolor{second!35}21.43 &  \cellcolor{second!35}.6855 &  \cellcolor{second!35}.2380 &  \cellcolor{second!35} 23.32 &   \cellcolor{third!35} .6903 &   \cellcolor{second!35} .2379 \\
			Sparse-DeRF~(w/\textcolor[rgb]{0.13, 0.55, 0.13}{DP-kernel})~-~Ours & \cellcolor{best!35} 21.29 & \cellcolor{best!35} .6179 &  \cellcolor{best!35}.3261 &  \cellcolor{best!35}27.34 &  \cellcolor{best!35}.8340 & \cellcolor{best!35} .1298 &  \cellcolor{best!35}28.19 &   \cellcolor{third!35}.7365 &  \cellcolor{best!35}.1606 &  \cellcolor{best!35}22.30 &  \cellcolor{best!35}.7233 &  \cellcolor{best!35}.2017 &  \cellcolor{best!35}22.21 &  \cellcolor{best!35}.7159 &  \cellcolor{best!35}.2036  &  \cellcolor{best!35}24.27 &   \cellcolor{best!35}.7255 &   \cellcolor{best!35}.2044 \\
			\midrule\midrule			
			\multirow{2}{*}{Real Scene}& \multicolumn{3}{c}{Ball} & \multicolumn{3}{c}{Basket} & \multicolumn{3}{c}{Buick} & \multicolumn{3}{c}{Coffee} & \multicolumn{3}{c}{Decoration} & & \\
			 & PSNR($\uparrow$) & SSIM($\uparrow$) & LPIPS($\downarrow$) & PSNR($\uparrow$) & SSIM($\uparrow$) & LPIPS($\downarrow$) & PSNR($\uparrow$) & SSIM($\uparrow$) & LPIPS($\downarrow$) & PSNR($\uparrow$) & SSIM($\uparrow$) & LPIPS($\downarrow$) & PSNR($\uparrow$) & SSIM($\uparrow$) & LPIPS($\downarrow$) &  & &\\
			\midrule
			SparseNeRF~\cite{wang2023sparsenerf} & \cellcolor{third!35}22.17 & .5654 & .4939 & 21.86 & .6733 & \cellcolor{third!35}.4081 & \cellcolor{third!35}  21.63 & \cellcolor{second!35} .6719 & .3747 & 26.17 & \cellcolor{third!35}.8382 & .2874 &  \cellcolor{third!35} 19.94 &  \cellcolor{second!35}.5696 & .4646\\
			SparseNeRF~\cite{wang2023sparsenerf} + Depth-Any V2~\cite{yang2024depthanythingv2} & 22.14 & \cellcolor{third!35}.5661 & \cellcolor{third!35}.4896 & \cellcolor{third!35}21.96 & \cellcolor{third!35}.6746 & .4099 & \cellcolor{best!35}21.66 & \cellcolor{best!35}.6754 & \cellcolor{third!35}.3736 & \cellcolor{third!35}26.21 & .8373 & \cellcolor{third!35}.2847 & \cellcolor{best!35}20.04 & \cellcolor{best!35}.5763 & \cellcolor{third!35}.4615 \\
			Sparse-DeRF~(w/\textcolor[rgb]{0.25, 0.0, 1.0}{DN-kernel})~-~Ours & \cellcolor{second!35}  23.76 &   \cellcolor{second!35}.6184 &   \cellcolor{second!35}.3607 &  \cellcolor{best!35} 23.39 &  \cellcolor{best!35}.7155 &   \cellcolor{best!35}.2329 &   \cellcolor{best!35}21.66 &   \cellcolor{third!35}.6405 &   \cellcolor{best!35}.2585 &  \cellcolor{second!35}  27.84 &  \cellcolor{second!35} .8388 &   \cellcolor{second!35} .2068 &   \cellcolor{second!35}19.98 &   .5553 &   \cellcolor{best!35}.3644 \\
			Sparse-DeRF~(w/\textcolor[rgb]{0.13, 0.55, 0.13}{DP-kernel})~-~Ours &   \cellcolor{best!35}24.80 &   \cellcolor{best!35}.6643 &   \cellcolor{best!35}.3200 &   \cellcolor{second!35} 23.32 &  \cellcolor{second!35} .6789 &   \cellcolor{second!35} .2455 & 21.27 & .6234 &  \cellcolor{second!35} .2662 &  \cellcolor{best!35} 28.91 &  \cellcolor{best!35} .8525 &  \cellcolor{best!35} .1902 &  19.69 &  .5440 &   \cellcolor{second!35} .3672 \\
			\midrule
			\multirow{2}{*}{Real Scene}& \multicolumn{3}{c}{Girl} & \multicolumn{3}{c}{Heron} & \multicolumn{3}{c}{Parterre} & \multicolumn{3}{c}{Puppet} & \multicolumn{3}{c}{Stair} & \multicolumn{3}{c}{Average} \\
			 & PSNR($\uparrow$) & SSIM($\uparrow$) & LPIPS($\downarrow$) & PSNR($\uparrow$) & SSIM($\uparrow$) & LPIPS($\downarrow$) & PSNR($\uparrow$) & SSIM($\uparrow$) & LPIPS($\downarrow$) & PSNR($\uparrow$) & SSIM($\uparrow$) & LPIPS($\downarrow$) & PSNR($\uparrow$) & SSIM($\uparrow$) & LPIPS($\downarrow$) & PSNR($\uparrow$) & SSIM($\uparrow$) & LPIPS($\downarrow$) \\
			\midrule
			SparseNeRF~\cite{wang2023sparsenerf} &  \cellcolor{third!35}18.92 &  \cellcolor{second!35}.7073 & .4091 & 19.09 & .4705 & \cellcolor{third!35}.4895 & 21.54 & .5630 & .4536 & 19.65 & \cellcolor{second!35} .5951 & .4017 & 21.79 &  \cellcolor{third!35} .5345 & .4673 & \cellcolor{third!35}21.28 & \cellcolor{third!35}.6189 & .4250 \\
			SparseNeRF~\cite{wang2023sparsenerf} + Depth-Any V2~\cite{yang2024depthanythingv2} & \cellcolor{best!35}19.40 & \cellcolor{best!35}.7156 & \cellcolor{third!35}.4050 & \cellcolor{third!35}19.17 & \cellcolor{third!35}.4731 & .4900 & \cellcolor{third!35}21.68 & \cellcolor{third!35}.5701 & \cellcolor{third!35}.4500 & \cellcolor{third!35}19.80 & \cellcolor{best!35}.5995 & \cellcolor{third!35}.4000 & \cellcolor{third!35}21.84 & \cellcolor{second!35}.5369 & \cellcolor{third!35}.4657 & 19.44 & .5659 & \cellcolor{third!35}.3845 \\
			Sparse-DeRF~(w/\textcolor[rgb]{0.25, 0.0, 1.0}{DN-kernel})~-~Ours &  18.86 &  .6860 &  \cellcolor{second!35} .2769 &  \cellcolor{best!35} 19.63 &   \cellcolor{second!35} .5011 &   \cellcolor{second!35} .2992 &  \cellcolor{second!35}  22.15 &  \cellcolor{second!35}  .5703 &  \cellcolor{second!35} .3787 & \cellcolor{best!35}  20.81 & \cellcolor{third!35}  .5882 &  \cellcolor{second!35} .3149 &   \cellcolor{second!35} 23.42 &  .5340 &  \cellcolor{second!35} .3374 &  \cellcolor{second!35}  22.15 &   \cellcolor{second!35} .6248 &   \cellcolor{second!35} .3030 \\
			Sparse-DeRF~(w/\textcolor[rgb]{0.13, 0.55, 0.13}{DP-kernel})~-~Ours & \cellcolor{second!35}  18.96 &   .6887 &   \cellcolor{best!35}.2744 &    \cellcolor{second!35}19.59 &   \cellcolor{best!35}.5051 &  \cellcolor{best!35} .2808 &  \cellcolor{best!35} 22.57 &   \cellcolor{best!35}.5994 &  \cellcolor{best!35} .3317 &   \cellcolor{second!35} 20.60 & .5756 &   \cellcolor{best!35}.3124 & \cellcolor{best!35}  23.48 &  \cellcolor{best!35} .5509 &  \cellcolor{best!35} .3189 & \cellcolor{best!35}  22.32 &  \cellcolor{best!35} .6283 &   \cellcolor{best!35}.2907 \\
			\midrule
      \end{tabular}
      }
   \label{tab:appendix_quant_results_depth_prior_6view}
   \end{center}
\end{table*}

\begin{figure*}[h]
  \centering
  \includegraphics[width=\textwidth]{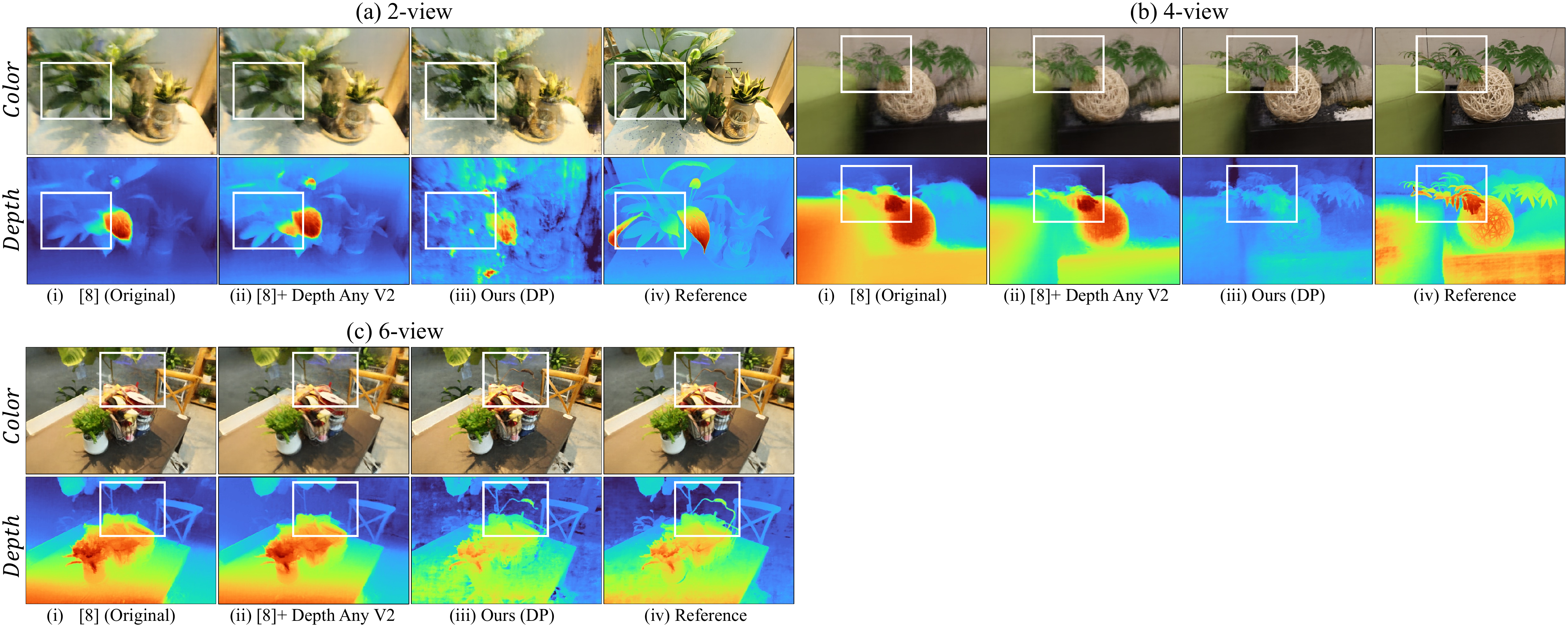}
  \caption{Additional qualitative ablation results on \textit{Puppet}, \textit{Ball}, and \textit{Basket} scene from 2-view, 4-view, and 6-view settings, which are the same scene in the main paper. The results show the comparison between SparseNeRF~\cite{wang2023sparsenerf} and SparseNeRF~\cite{wang2023sparsenerf} trained with the depth maps predicted from Depth-Anything V2~\cite{yang2024depthanythingv2}, which are indicated as reference number~\cite{wang2023sparsenerf} and Depth-Any V2, respectively.}
  \label{fig:abl_sparsenerf_depth_anything}
\end{figure*}

\section{Additional Analysis and Experimental Results}
\subsection{Models with Depth Prior}
\label{appendix_subsec:depth_quantitative}
We present the quantitative results of our model and SparseNeRF~\cite{wang2023sparsenerf} in~\tablename~\ref{tab:abls_depth_prior_depth_anything_avg},~\ref{tab:appendix_quant_results_depth_prior_2view},~\ref{tab:appendix_quant_results_depth_prior_4view}, and~\ref{tab:appendix_quant_results_depth_prior_6view}.
Additionally, we include the results of SparseNeRF~\cite{wang2023sparsenerf} trained with the more recent state-of-the-art monocular depth estimation model, Depth-Anything V2~\cite{yang2024depthanythingv2}, in these tables and present the qualitative comparison in~\figurename~\ref{fig:abl_sparsenerf_depth_anything}.
Interestingly, the results show that SparseNeRF~\cite{wang2023sparsenerf} using depth maps predicted by Depth-Anything V2~\cite{yang2024depthanythingv2} yields lower PNSR and SSIM scores but a higher LPIPS score in real-world scenes compared to the originally adopted depth maps from DPT~\cite{ranftl2021vision}. 
We believe that this degradation occurs because Depth-Anything V2~\cite{yang2024depthanythingv2} predicts more accurate depth maps even from blurred images in real-world data. 
However, this improvement in depth estimation unexpectedly exacerbates multi-view inconsistency among the predicted depth maps, which in turn negatively affects the optimization process of the radiance field in real scenes.
As a consequence, the multi-view inconsistency issue, which becomes particularly noticeable across four or more views, remains unresolved.
This degradation is also evident in the lower quality of the rendered images in~\figurename~\ref{fig:abl_sparsenerf_depth_anything}.

\begin{table*}
   \footnotesize
   \begin{center}
      \caption{\textbf{Average} results of MGS experiments on \textbf{world coordinate systems}. The experiments are conducted on both synthetic and real scenes obtained from 2-view, 4-view, and 6-view settings. Color shading represents the \colorbox{best!35}{best} and  \colorbox{second!35}{second} results, respectively.}
      \renewcommand\arraystretch{1.2}
      \resizebox{\linewidth}{!}{
		\setlength{\tabcolsep}{1pt}
         \begin{tabular}{c|c||ccc|ccc||ccc|ccc||ccc|ccc}
	        \multicolumn{2}{c}{} & \multicolumn{6}{c}{[~2-view ]} & \multicolumn{6}{c}{[~4-view ]} & \multicolumn{6}{c}{[~6-view ]}\\
			\multicolumn{2}{c}{} & \multicolumn{3}{c}{[~Synthetic Scene~]} & \multicolumn{3}{c}{[~Real Scene~]} & \multicolumn{3}{c}{[~Synthetic Scene~]} & \multicolumn{3}{c}{[~Real Scene~]} & \multicolumn{3}{c}{[~Synthetic Scene~]} & \multicolumn{3}{c}{[~Real Scene~]}\\
			Blur Kernel & Gradient Scaling & ~PSNR($\uparrow$)~ & SSIM($\uparrow$)~ & LPIPS($\downarrow$)~& ~PSNR($\uparrow$)~ & SSIM($\uparrow$)~ & LPIPS($\downarrow$)~ & ~PSNR($\uparrow$)~ & SSIM($\uparrow$)~ & LPIPS($\downarrow$)~& ~PSNR($\uparrow$)~ & SSIM($\uparrow$)~ & LPIPS($\downarrow$)~ & ~PSNR($\uparrow$)~ & SSIM($\uparrow$)~ & LPIPS($\downarrow$)~& ~PSNR($\uparrow$)~ & SSIM($\uparrow$)~ & LPIPS($\downarrow$)~ \\
			\midrule
			\textcolor[rgb]{0.13, 0.55, 0.13}{DP-kernel} & $\times$  & \cellcolor{second!35}14.46 & \cellcolor{best!35}.2614 & .6258 & \cellcolor{second!35}13.26 & \cellcolor{second!35}.1797 & .6732 & \cellcolor{second!35}18.46 & \cellcolor{best!35}.4554 & \cellcolor{second!35}.4691 & 16.92 & \cellcolor{second!35}.3679 & .4570 & \cellcolor{second!35}19.04 & \cellcolor{second!35}.4822 & \cellcolor{second!35}.4408 & \cellcolor{second!35}19.38 & \cellcolor{second!35}.4760 & \cellcolor{second!35}.3740 \\
			\textcolor[rgb]{0.13, 0.55, 0.13}{DP-kernel} & + Naive~\cite{philip2023floatersnomore} & 14.41 & \cellcolor{second!35}.2579 & \cellcolor{second!35}.6204 & 13.20 & .1789 & \cellcolor{second!35}.6680 & 18.40 & \cellcolor{second!35}.4506 & .4706 & \cellcolor{second!35}17.05 & .3654 & \cellcolor{second!35}.4544 & 19.01 & \cellcolor{best!35}.4830 & .4425 & 19.12 & .4575 & .3831 \\
			\textcolor[rgb]{0.13, 0.55, 0.13}{DP-kernel} & + MGS & \cellcolor{best!35}14.49 & .2564 & \cellcolor{best!35}.5974 & \cellcolor{best!35}15.22 & \cellcolor{best!35}.2880 & \cellcolor{best!35}.5362 & \cellcolor{best!35}18.54 & .4493 & \cellcolor{best!35}.4300 & \cellcolor{best!35}17.91 & \cellcolor{best!35}.4061 & \cellcolor{best!35}.4019 & \cellcolor{best!35}19.10 & .4781 & \cellcolor{best!35}.3963 & \cellcolor{best!35}19.79 & \cellcolor{best!35}.4923 & \cellcolor{best!35}.3470 \\
			\midrule
      \end{tabular}
      }
   \vspace{-0.4cm}
   \label{tab:abls_mgs_world_avg}
   \end{center}
\end{table*}

\begin{figure*}[h]
  \centering
  \includegraphics[width=\textwidth]{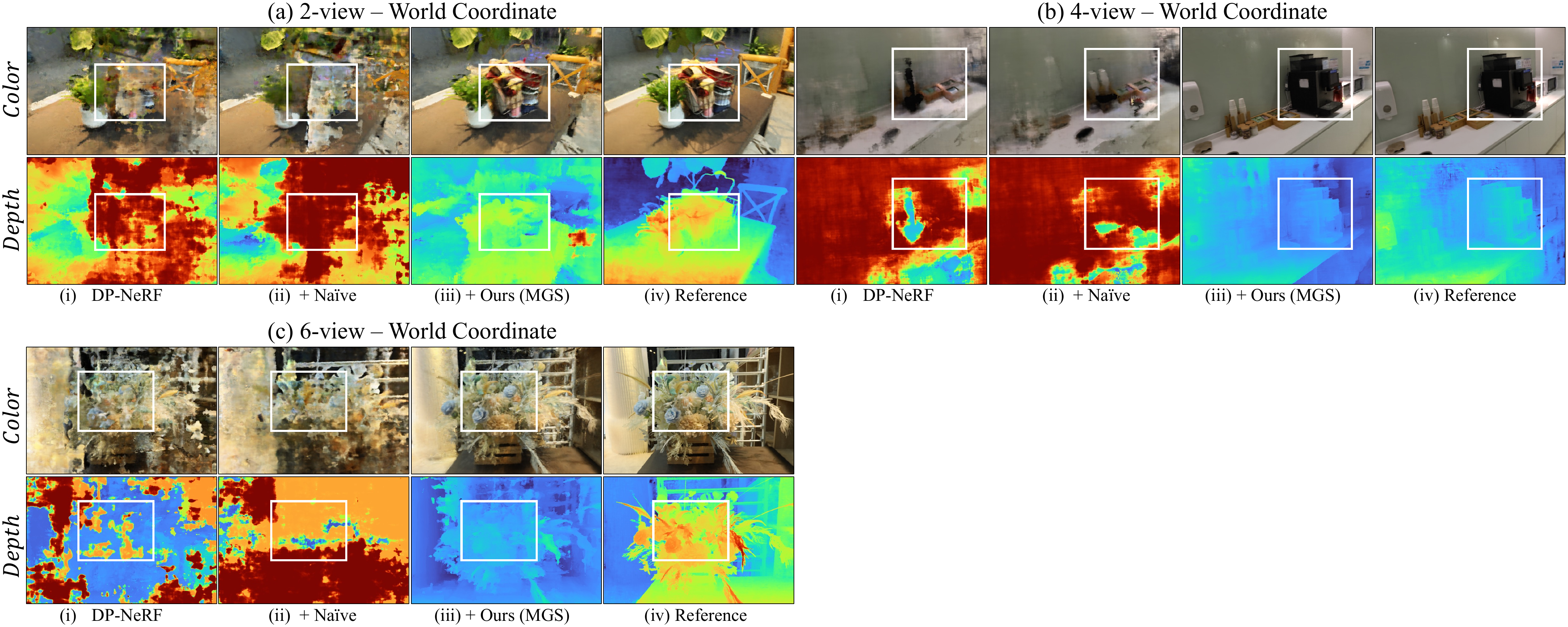}
  \caption{Additional qualitative ablation results on \textit{Basket}, \textit{Coffee}, and \textit{Decoration} scene from 2-view, 4-view, and 6-view settings, which are trained on world coordinate system. The results demonstrate that our proposed MGS guides the geometry of the scenes and works properly even in the world coordinate system.}
  \vspace{-0.4cm}
  \label{fig:mgs_world_average}
\end{figure*}

\subsection{Geometric Guidance of Modulated Gradient Scaling}
\label{appendix_subsec:geo_mgs}
We initially explored a gradient scaling function that could operate in normalized device coordinate~(NDC). However, inspired by the idea that gradient scaling could serve as geometric guidance for the scene, we proposed MGS.
To demonstrate that MGS can function as a structural clue regardless of the coordinate systems, we present~\tablename~\ref{tab:abls_mgs_world_avg} along with~\figurename~\ref{fig:mgs_world_average}. 
The experiments were conducted in the world coordinate system, using the same MGS parameters for each scene as presented  in~\tablename~\ref{tab:appendix_scene_indices} to match the scaling function shape in NDC for fair comparison.
We converted the distances of ray samples along the rays in the world coordinate system to the NDC depth scale and used them as $\delta_{\textbf{s}_{i}}$ in Eq.21 of the main paper.
Note that, other than this scaling adjustment, all settings for training the radiance fields remained in the world coordinate system.
The kernel employed for the experiments is \textcolor[rgb]{0.13, 0.55, 0.13}{DP-kernel}.
The results show that our proposed MGS achieves superior performance even in the world coordinate system.
These findings validate that MGS is not limited to a specific coordinate system but effectively captures the arrangement of scene components, facilitating the learning of radiance fields as intended.
We additionally attach the quantitative results for entire scenes in Section~\ref{appendix_subsec:quantitative_mgs_world_coord_entire}.

%% file: sections/appendix_additional_implementation_detail.tex
\newcolumntype{L}[1]{>{\raggedright\let\newline\\\arraybackslash\hspace{0pt}}m{#1}}
\newcolumntype{C}[1]{>{\centering\let\newline\\\arraybackslash\hspace{0pt}}m{#1}}
\newcolumntype{R}[1]{>{\raggedleft\let\newline\\\arraybackslash\hspace{0pt}}m{#1}}
\definecolor{lightgray}{rgb}{0.83, 0.83, 0.83}

\begin{table*}
   \begin{center}
   \caption{Image indices of each scene for training Sparse-DeRF.}
		\setlength{\tabcolsep}{1pt}
         \begin{tabular}{C{2.1cm}|C{1.5cm}|C{1.5cm}|C{1.5cm}|C{1.5cm}|C{1.5cm}|C{1.5cm}|C{1.5cm}|C{1.5cm}|C{1.5cm}|C{1.5cm}}
         	\midrule
			Real Scene~&~\textit{Ball}~ &~\textit{Basket}~&~\textit{Buick}~&~\textit{Coffee}~&~\textit{Decoration}~&~\textit{Girl}~&~\textit{Heron}~&~\textit{Parterre}~&~\textit{Puppet}~&~\textit{Stair}~\\
			\midrule
			$2$-view & 1, 12 & 12, 33 & 11, 39 & 3, 10 & 1, 19 & 9, 16 & 11, 35 & 8, 26 & 9, 31 & 13, 26 \\
			\cellcolor{lightgray}$4$-view & \cellcolor{lightgray}1, 12, 18, 22 & \cellcolor{lightgray}1, 12, 22, 33 & \cellcolor{lightgray}5, 11, 20, 39 & \cellcolor{lightgray}3, 10, 15, 26 & \cellcolor{lightgray}1, 19, 22, 39 & \cellcolor{lightgray}2, 9, 16, 32 & \cellcolor{lightgray}4, 11, 18, 35 & \cellcolor{lightgray}1, 8, 13, 26 & \cellcolor{lightgray}9, 13, 21, 31 & \cellcolor{lightgray}4, 13, 16, 26 \\
			$6$-view & 1, 5, 10, 12, 18, 22 & 1, 8, 12, 17, 22, 33 & 5, 11, 17, 20, 34, 39 & 3, 10, 11, 15, 21, 26 & 1, 14, 19, 22, 27, 39 & 2, 9, 16, 24, 32, 37 & 4, 11, 18, 23, 27, 35 & 1, 8, 13, 17, 26, 28 & 7, 9, 13, 21, 23, 31 & 2, 4, 13, 16, 26, 34 \\
			\midrule
			\midrule
			Synthetic Scene~&~\textit{Cozyroom}~ &~\textit{Factory}~&~\textit{Pool}~&~\textit{Tanabata}~&~\textit{Trolley}\\
			\midrule
			$2$-view & 2, 17 & 3, 19 & 10, 23 & 1, 7 & 13, 23 \\
			\cellcolor{lightgray}$4$-view & \cellcolor{lightgray}2, 17, 23, 29 & \cellcolor{lightgray}3, 14, 19, 33 & \cellcolor{lightgray}5, 10, 15, 23 & \cellcolor{lightgray}1, 7, 11, 22 & \cellcolor{lightgray}7, 13, 23, 31 \\
			$6$-view & 2, 14, 17, 21, 23, 29 & 1, 3, 14, 19, 28, 33 & 1, 5, 10, 15, 20, 23 & 1, 7, 11, 18, 22, 27 & 7, 13, 20, 23, 27, 31 \\
			\midrule
      \end{tabular}
   \vspace{-0.4cm}
   \label{tab:appendix_scene_indices}
   \end{center}
\end{table*}

\begin{table*}
   \begin{center}
   \caption{Hyper parameters of modulated gradient scaling~(MGS) for entire scenes of the Deblur-NeRF~\cite{ma2022deblurnerf} dataset. $\rho$ and $\eta$ denote the magnitude and period of sine function, respectively.}
		\setlength{\tabcolsep}{1pt}
         \begin{tabular}{C{2.5cm}|C{2cm}|C{1.3cm}|C{1cm}|C{2cm}|C{2cm}|C{1cm}|C{1cm}|C{1.5cm}|C{1.1cm}|C{1.1cm}}
         	\midrule
			Synthetic Scene~&~\textit{Cozyroom}~ &~\textit{Factory}~&~\textit{Pool}~&~\textit{Tanabata}~&~\textit{Trolley}\\
			\midrule
			$\rho$ & 10.0 & 10.0 & 10.0 & 1.0 & 10.0 \\
			$\eta$ & 1.75 & 1.75 & 1.75 & 1.5 & 1.75 \\
			\midrule
			\midrule
			Real Scene~&~\textit{Ball}~ &~\textit{Basket}~&~\textit{Buick}~&~\textit{Coffee}~&~\textit{Decoration}~&~\textit{Girl}~&~\textit{Heron}~&~\textit{Parterre}~&~\textit{Puppet}~&~\textit{Stair}~\\
			\midrule
			$\rho$ & 1.0 & 1.0 & 10.0 & 1.0 & 1.0 & 1.0 & 1.0 & 1.0 & 10.0 & 1.0 \\
			$\eta$ & 1.2 & 0.67 & 1.75 & 0.67 & 0.5 & 0.5 & 0.5 & 0.5 & 1.75 & 0.5 \\
			\midrule
      \end{tabular}
   \label{tab:appendix_mgs_parameters}
   \end{center}
\end{table*}

\section{Additional Implementation Details}
\label{appendix_sec:additional_implementation_details}

\subsection{Training Scene Indices}
\label{appendix_subsec:training_scene_indices}
For fair comparison in future research, we present the image indices of each scene for training the Sparse-DeRF in~\tablename~\ref{tab:appendix_scene_indices}.
The indices are manually selected from the training images of each scene as we mentioned in the main paper.
Note that, the indices of the 2-view and 4-view settings are subsets of the 6-view setting.

\subsection{Parameters for Entire Scenes}
\label{appendix_subsec:mgs_parameters}
In~\tablename~\ref{tab:appendix_mgs_parameters}, we present the hyper-parameters of Modulated Gradient Scaling~(MGS) for entire scenes, which consist of period $\eta$ and magnitude $\rho$ of the sine function.
As we indicate in Section~IV-B in the main paper, we set the magnitude $\rho$ as high value to only ignore the gradient in very near distance regions for scenes such as \textit{Buick}, \textit{Puppet}, \textit{Cozyroom}, \textit{Factory}, \textit{Pool}, and \textit{Trolley}.
Please refer to the figure of the function shape depending on the parameters in the main paper.

%% file: sections/appendix_additional_results.tex
\newcolumntype{L}[1]{>{\raggedright\let\newline\\\arraybackslash\hspace{0pt}}m{#1}}
\newcolumntype{C}[1]{>{\centering\let\newline\\\arraybackslash\hspace{0pt}}m{#1}}
\newcolumntype{R}[1]{>{\raggedleft\let\newline\\\arraybackslash\hspace{0pt}}m{#1}}

\section{Additional Quantitative Results}
\subsection{Quantitative Results for Entire Scenes}
\label{appendix_subsec:quantitative_results}
We present the comprehensive experimental results for entire scenes from 2-view, 4-view, and 6-view settings in~\tablename~\ref{tab:appendix_quantitative_results_2view},~\ref{tab:appendix_quantitative_results_4view}, and~\ref{tab:appendix_quantitative_results_6view}, respectively.
In the tables, \textcolor[rgb]{0.13, 0.55, 0.13}{DP-kernel} and \textcolor[rgb]{0.25, 0.0, 1.0}{DN-kernel} denote the kernel of~\cite{lee2023dpnerf} and~\cite{ma2022deblurnerf}, respectively.
The results of the Sparse-DeRF in the~\tablename~\ref{tab:appendix_quantitative_results_2view},~\ref{tab:appendix_quantitative_results_4view}, and~\ref{tab:appendix_quantitative_results_6view} are representative results as we presented in Section~V-E1 in the main paper.
In addition, we present more detailed ablation results for entire scenes from 2-view, 4-view, and 6-view settings in~\tablename~\ref{tab:appendix_ablation_quant_results_2view},~\ref{tab:appendix_ablation_quant_results_4view}, and~\ref{tab:appendix_ablation_quant_results_6view}, respectively.
As we mentioned before, we also present the separated ablation results according to the type of the kernel from DP-NeRF~\cite{lee2023dpnerf} and Deblur-NeRF~\cite{ma2022deblurnerf}.

\subsection{Complex Optimization Problem for Entire Scenes}
\label{appendix_subsec:quantitative_results_complex_optimization}
We present the comprehensive experimental results of the complex optimization problem for entire scenes from 2-view, 4-view, and 6-view settings in~\tablename~\ref{tab:appendix_regnerf_ablation_2view},~\ref{tab:appendix_regnerf_ablation_4view}, and~\ref{tab:appendix_regnerf_ablation_6view}, respectively.
In addition, we also present the quantitative results of the experiments on the \textit{Decoration} scene that varies the number of sparse views in~\tablename~\ref{tab:appendix_regnerf_ablation_2view_to_10_view_decoration} as we mentioned in the main paper.

\subsection{Comparison to Naive Gradient Scaling for Entire Scenes}
\label{appendix_subsec:quantitative_results_ngs_mgs_comparison}
We present the quantitative comparison results for entire scenes that compare our modulated gradient scaling~(MGS) and naive gradient scaling of~\cite{philip2023floatersnomore} from 2-view, 4-view, and 6-view settings in~\tablename\ref{tab:appendix_gradient_scaling_comparison_2view},~\ref{tab:appendix_gradient_scaling_comparison_4view}, and~\ref{tab:appendix_gradient_scaling_comparison_6view}, respectively.

\subsection{World Coordinate System Experiment for Entire Scenes}
\label{appendix_subsec:quantitative_mgs_world_coord_entire}
We present the quantitative comparison results for entire scenes that include MGS experiments on the world coordinate system from 2-view, 4-view, and 6-view settings in~\tablename\ref{tab:appendix_mgs_world_exps_2view},~\ref{tab:appendix_mgs_world_exps_4view}, and~\ref{tab:appendix_mgs_world_exps_6view}, respectively.
The results demonstrate the ability of our modulated gradient scaling~(MGS) as geometric guidance.

\begin{table*}[t]
   \Large
   \begin{center}
   \caption{Quantitative results for the entire scene of synthetic and real scenes from \textbf{2-view} settings. Each color shading represents the \colorbox{best!35}{best}, \colorbox{second!35}{second best} and \colorbox{third!35}{third best} result, respectively.}
      \resizebox{2\columnwidth}{!}{
		\centering
		\setlength{\tabcolsep}{1pt}
         \begin{tabular}{c||ccc|ccc|ccc|ccc|ccc||ccc}
			\midrule
			\multirow{2}{*}{Synthetic Scene}& \multicolumn{3}{c}{Factory} & \multicolumn{3}{c}{Cozyroom} & \multicolumn{3}{c}{Pool} & \multicolumn{3}{c}{Tanabata} & \multicolumn{3}{c}{Trolley} & \multicolumn{3}{c}{Average} \\
			 & PSNR($\uparrow$) & SSIM($\uparrow$) & LPIPS($\downarrow$) & PSNR($\uparrow$) & SSIM($\uparrow$) & LPIPS($\downarrow$) & PSNR($\uparrow$) & SSIM($\uparrow$) & LPIPS($\downarrow$) & PSNR($\uparrow$) & SSIM($\uparrow$) & LPIPS($\downarrow$) & PSNR($\uparrow$) & SSIM($\uparrow$) & LPIPS($\downarrow$) & PSNR($\uparrow$) & SSIM($\uparrow$) & LPIPS($\downarrow$) \\
			\midrule
			Naive NeRF~\cite{mildenhall2021nerf} & 14.24 & .2186 & .6867 & \cellcolor{second!35}21.07 & \cellcolor{best!35}.6127 & .3644 & 18.59 & \cellcolor{second!35}.3575 & .4288 & 10.84 & .1517 & .6500 & 10.82 & .1591 & .6593 & 15.11 & \cellcolor{second!35}.2999 & .5578 \\
			MPR~\cite{zamir2021MPR}+NeRF & \cellcolor{second!35}14.43 & \cellcolor{second!35}.2250 & .6858 & \cellcolor{best!35}21.08 & \cellcolor{second!35}.6111 & .3671 & \cellcolor{third!35}18.72 & \cellcolor{best!35}.3676 & .4228 & 11.03 & .1456 & .6553 & 10.56 & .1535 & .6663 & \cellcolor{third!35}15.16 & \cellcolor{best!35}.3006 & .5595 \\
			Deblur-NeRF~\cite{ma2022deblurnerf} & 14.14 &  .2009 &  .6647 &  20.58 &  \cellcolor{third!35}.5768 &  \cellcolor{second!35}.3443 &  18.37 &  .3247 &  \cellcolor{third!35}.4061 &  \cellcolor{third!35}11.56 &  \cellcolor{third!35}.1704 & \cellcolor{third!35}.6075 &  11.05 &  .1690 & .6425 &  15.14 &  .2884 &  \cellcolor{third!35}.5330 \\
			DP-NeRF~\cite{lee2023dpnerf} &  14.14 &  .2091 &  \cellcolor{third!35}.6540 & \cellcolor{third!35}20.71 &  .5752 &  .3487 &  18.38 &  .3276 & .4127 &  11.21 &  .1482 &  .6247 &  10.88 &  .1536 &  .6543 &  15.06 &  .2827 &  .5389 \\
			RegNeRF~\cite{niemeyer2022regnerf}~(No kernel) & \cellcolor{best!35}14.57 & \cellcolor{best!35}.2523 & .6680 & 17.13 & .4616 & .3808 & 14.23 & .1473 & .7298 & 13.01 & .2408 & .5956 & \cellcolor{best!35}14.04 & \cellcolor{best!35}.3227 & \cellcolor{best!35}.5602 & 14.60 & .2849 & .5869 \\
			RegNeRF~\cite{niemeyer2022regnerf}~(w/\textcolor[rgb]{0.13, 0.55, 0.13}{DP-kernel}) &  14.20 &  \cellcolor{third!35}.2214 &  \cellcolor{best!35}.6412 &  18.88 &  .5136 &  \cellcolor{third!35}.3476 &  13.03 &  .1003 & .6845 &  9.21 &  .0815 &  .6920 &  10.86 &  .1644 &  .6657 &  13.24 &  .2162 &  .6062 \\
			Sparse-DeRF~(w/\textcolor[rgb]{0.25, 0.0, 1.0}{DN-kernel})~-~Ours &  \cellcolor{third!35}14.27 &  .2200 &  .6564 &  20.57 &  .5675 &  .3589 &  \cellcolor{second!35}19.81 & .3330 &  \cellcolor{second!35}.4058 &  \cellcolor{second!35}11.67 &  \cellcolor{second!35}.1827 &  \cellcolor{second!35}.5947 &  \cellcolor{third!35}11.28 &  \cellcolor{second!35}.1800 &  \cellcolor{second!35}.6298 &  \cellcolor{best!35}15.52 &  \cellcolor{third!35}.2966 &  \cellcolor{best!35}.5291 \\
			Sparse-DeRF~(w/\textcolor[rgb]{0.13, 0.55, 0.13}{DP-kernel})~-~Ours &  14.10 &  .2116 &  \cellcolor{second!35}.6413 &  18.97 &  .5206 &  \cellcolor{best!35}.3417 &  \cellcolor{best!35}20.32 &  \cellcolor{third!35}.3488 &  \cellcolor{best!35}.4056 &  \cellcolor{best!35}12.00 &  \cellcolor{best!35}.1943 &  \cellcolor{best!35}.5902 &  \cellcolor{second!35}11.36 &  \cellcolor{third!35}.1769 &  \cellcolor{third!35}.6422 &  \cellcolor{second!35}15.35 &  .2904 &  \cellcolor{second!35}.5242 \\
			\midrule\midrule
			
			\multirow{2}{*}{Real Scene}& \multicolumn{3}{c}{Ball} & \multicolumn{3}{c}{Basket} & \multicolumn{3}{c}{Buick} & \multicolumn{3}{c}{Coffee} & \multicolumn{3}{c}{Decoration} & & \\
			 & PSNR($\uparrow$) & SSIM($\uparrow$) & LPIPS($\downarrow$) & PSNR($\uparrow$) & SSIM($\uparrow$) & LPIPS($\downarrow$) & PSNR($\uparrow$) & SSIM($\uparrow$) & LPIPS($\downarrow$) & PSNR($\uparrow$) & SSIM($\uparrow$) & LPIPS($\downarrow$) & PSNR($\uparrow$) & SSIM($\uparrow$) & LPIPS($\downarrow$) &  & &\\
			\midrule
			Naive NeRF~\cite{mildenhall2021nerf} & 18.45 & .4363 & .5693 & 13.32 & \cellcolor{third!35}.2595 & .6131 & 13.49 & .2620 & .5680 & 17.76 & .5249 & .4807 & 12.25 & .1718 & .6738 &  &  & \\
			MPR~\cite{zamir2021MPR}+NeRF & 18.69 & .4413 & .5659 & 13.09 & .2505 & .6194 & 13.31 & .2574 & .5706 & 18.13 & .5243 & .4869 & \cellcolor{third!35}12.41 & \cellcolor{third!35}.1761 & .6735 &  &  &  \\
			Deblur-NeRF~\cite{ma2022deblurnerf} & 18.80 &  .4325 &  .5521 &  12.86 &  .2414 &  .5947 &  13.11 &  .2446 &  .5609 &  18.66 & .5272 &  .4696 &  12.09 &  .1508 &  .6722 &  &  & \\
			DP-NeRF~\cite{lee2023dpnerf} &  18.22 &  .4200 &  .5611 &  13.24 &  .2263 &  .6062 & 13.43 &  .2480 &  .5571 &  18.15 &  .5191 &  .4876 &  12.34 &  .1672 &  \cellcolor{third!35}.6675 &   &   &   \\
			RegNeRF~\cite{niemeyer2022regnerf}~(No kernel) & \cellcolor{best!35}20.48 & \cellcolor{best!35}.4951 & \cellcolor{third!35}.5398 & \cellcolor{best!35}15.53 & \cellcolor{best!35}.3388 & \cellcolor{best!35}.5403 & \cellcolor{best!35}17.05 & \cellcolor{best!35}.4241 & \cellcolor{best!35}.4361 & \cellcolor{best!35}23.24 & \cellcolor{best!35}.6950 & \cellcolor{best!35}.3505 & 10.94 & .0816 & .7545 \\
			RegNeRF~\cite{niemeyer2022regnerf}~(w/\textcolor[rgb]{0.13, 0.55, 0.13}{DP-kernel}) &  18.58 &  .4297 &  .5565 & 13.49 &  .2429 &  .5985 &  12.51 &  .2199 & .5579 &  15.71 &  .4280 &  .5370 &  11.15 &  .0927 &  .7520 &  &  & \\
			Sparse-DeRF~(w/\textcolor[rgb]{0.25, 0.0, 1.0}{DN-kernel})~-~Ours &  \cellcolor{third!35}20.07 &  \cellcolor{second!35}.4612 &  \cellcolor{second!35}.5290 &  \cellcolor{third!35}13.62 &  \cellcolor{third!35}.2678 &  \cellcolor{third!35}.5835 &  \cellcolor{second!35}14.38 &  \cellcolor{second!35}.3182 &  \cellcolor{second!35}.4913 &  \cellcolor{third!35}19.51 &  \cellcolor{third!35}.6048 &  \cellcolor{third!35}.4146 &  \cellcolor{second!35}13.09 &  \cellcolor{second!35}.2173 &  \cellcolor{second!35}.6256 &   &   &  \\
			Sparse-DeRF~(w/\textcolor[rgb]{0.13, 0.55, 0.13}{DP-kernel})~-~Ours &  \cellcolor{second!35}20.09 &  \cellcolor{third!35}.4583 &  \cellcolor{best!35}.5244 &  \cellcolor{second!35}14.05 &  \cellcolor{second!35}.2890 &  \cellcolor{second!35}.5555 & \cellcolor{third!35}13.78 &  \cellcolor{third!35}.2864 &  \cellcolor{third!35}.5036 &  \cellcolor{second!35}19.67 &  \cellcolor{second!35}.6084 &  \cellcolor{second!35}.4096 &  \cellcolor{best!35}13.11 &  \cellcolor{best!35}.2192 &  \cellcolor{best!35}.6192 &   &   &  \\
			\midrule
			\multirow{2}{*}{Real Scene}& \multicolumn{3}{c}{Girl} & \multicolumn{3}{c}{Heron} & \multicolumn{3}{c}{Parterre} & \multicolumn{3}{c}{Puppet} & \multicolumn{3}{c}{Stair} & \multicolumn{3}{c}{Average} \\
			 & PSNR($\uparrow$) & SSIM($\uparrow$) & LPIPS($\downarrow$) & PSNR($\uparrow$) & SSIM($\uparrow$) & LPIPS($\downarrow$) & PSNR($\uparrow$) & SSIM($\uparrow$) & LPIPS($\downarrow$) & PSNR($\uparrow$) & SSIM($\uparrow$) & LPIPS($\downarrow$) & PSNR($\uparrow$) & SSIM($\uparrow$) & LPIPS($\downarrow$) & PSNR($\uparrow$) & SSIM($\uparrow$) & LPIPS($\downarrow$) \\
			\midrule
			Naive NeRF~\cite{mildenhall2021nerf} & 11.06 & \cellcolor{third!35}.2298 & \cellcolor{third!35}.6304 & 13.69 & \cellcolor{third!35}.1564 & .5683 & 15.20 & .2528 & .6462 & 14.34 & .2937 & .6156 & 14.27 & .0474 & .6385 & 14.38 & .2635 & .6004\\
			MPR~\cite{zamir2021MPR}+NeRF & 10.80 & .1977 & .6424 & 13.51 & .1538 & .5652 & 15.45 & \cellcolor{third!35}.2667 & .6422 & 14.22 & .2844 & .6135 & 14.15 & .0414 & .6392 & 14.38 & .2594 & .6019 \\
			Deblur-NeRF~\cite{ma2022deblurnerf} &  10.94 &  .2031 &  .6436 &  \cellcolor{third!35}13.97 &  .1459 &  .5499 &  15.02 &  .2320 &  .6431 &  14.28 &  .2764 &  .6036 &  14.40 &  .0525 &  .6313 & 14.41 &  .2506 &  .5921 \\
			DP-NeRF~\cite{lee2023dpnerf} &  \cellcolor{third!35}11.14 &  .2151 &  .6485 &  13.63 &  .1473 &  \cellcolor{third!35}.5376 &  14.70 &  .2226 &  .6348 &  14.32 &  .2887 &  .5930 &  14.38 &  .0521 &  \cellcolor{best!35}.6104 &  14.36 &  .2506 &  .5904 \\
			RegNeRF~\cite{niemeyer2022regnerf}~(No kernel) & 8.31 & .0602 & .7545 & 11.43 & .0951 & .6687 & \cellcolor{third!35}15.58 & \cellcolor{second!35}.2709 & \cellcolor{third!35}.6510 & \cellcolor{best!35}16.02 & \cellcolor{best!35}.3548 & \cellcolor{best!35}.5739 & \cellcolor{best!35}16.30 & \cellcolor{best!35}.1816 & .6182 & \cellcolor{third!35}15.49 & \cellcolor{third!35}.2997 & \cellcolor{third!35}.5888 \\
			RegNeRF~\cite{niemeyer2022regnerf}~(w/\textcolor[rgb]{0.13, 0.55, 0.13}{DP-kernel}) &  7.86 &  .0419 &  .7737 &  11.57 &  .0783 &  .6246 &  12.54 &  .1524 & .7091 &  10.67 &  .1331 &  .6604 &  13.49 &  .0170 &  .6770 &  12.76 &  .1836 &  .6447 \\
			Sparse-DeRF~(w/\textcolor[rgb]{0.25, 0.0, 1.0}{DN-kernel})~-~Ours &  \cellcolor{second!35}12.85 &  \cellcolor{second!35}.3418 &  \cellcolor{second!35}.5585 &  \cellcolor{best!35}14.30 &  \cellcolor{best!35}.1706 &  \cellcolor{best!35}.5197 &  \cellcolor{best!35}16.61 &  \cellcolor{best!35}.2903 &  \cellcolor{second!35}.5912 &  \cellcolor{third!35}14.91 &  \cellcolor{second!35}.3141 &  \cellcolor{third!35}.5886 &  \cellcolor{third!35}15.97 &  \cellcolor{third!35}.1255 &  \cellcolor{second!35}.6125 &  \cellcolor{second!35}15.53 &  \cellcolor{second!35}.3112 &  \cellcolor{second!35}.5515 \\
			Sparse-DeRF~(w/\textcolor[rgb]{0.13, 0.55, 0.13}{DP-kernel})~-~Ours &  \cellcolor{best!35}13.11 &  \cellcolor{best!35}.3707 &  \cellcolor{best!35}.5486 &  \cellcolor{second!35}14.17 &  \cellcolor{second!35}.1669 &  \cellcolor{second!35}.5285 &  \cellcolor{second!35}16.19 &  .2527 &  \cellcolor{best!35}.5908 &  \cellcolor{second!35}15.25 &  \cellcolor{best!35}.3277 &  \cellcolor{best!35}.5739 &  \cellcolor{best!35}16.30 &  \cellcolor{second!35}.1343 &  \cellcolor{third!35}.6133 &  \cellcolor{best!35}15.57 &  \cellcolor{best!35}.3114 &  \cellcolor{best!35}.5467 \\
			\midrule
      \end{tabular}
      }
   \label{tab:appendix_quantitative_results_2view}
   \end{center}
\end{table*}

\begin{table*}[t]
   \Large
   \begin{center}
   \caption{Quantitative results for the entire scenes of synthetic and real scenes from \textbf{4-view} settings. Each color shading represents the \colorbox{best!35}{best}, \colorbox{second!35}{second best} and \colorbox{third!35}{third best} result, respectively.}
      \resizebox{2\columnwidth}{!}{
		\centering
		\setlength{\tabcolsep}{1pt}
         \begin{tabular}{c||ccc|ccc|ccc|ccc|ccc||ccc}
			\midrule
			\multirow{2}{*}{Synthetic Scene}& \multicolumn{3}{c}{Factory} & \multicolumn{3}{c}{Cozyroom} & \multicolumn{3}{c}{Pool} & \multicolumn{3}{c}{Tanabata} & \multicolumn{3}{c}{Trolley} & \multicolumn{3}{c}{Average} \\
			 & PSNR($\uparrow$) & SSIM($\uparrow$) & LPIPS($\downarrow$) & PSNR($\uparrow$) & SSIM($\uparrow$) & LPIPS($\downarrow$) & PSNR($\uparrow$) & SSIM($\uparrow$) & LPIPS($\downarrow$) & PSNR($\uparrow$) & SSIM($\uparrow$) & LPIPS($\downarrow$) & PSNR($\uparrow$) & SSIM($\uparrow$) & LPIPS($\downarrow$) & PSNR($\uparrow$) & SSIM($\uparrow$) & LPIPS($\downarrow$) \\
			\midrule
			Naive NeRF~\cite{mildenhall2021nerf} & 16.63 & .3494 & .5861 & 22.75 & .6935 & .2903 & \cellcolor{best!35}25.92 & \cellcolor{second!35}.6419 & \cellcolor{third!35}.2495 & 16.15 & .4408 & .4635 & 18.66 & .5380 & .4104 & \cellcolor{third!35}20.02 & .5327 & .4000 \\
			MPR~\cite{zamir2021MPR}+NeRF & 16.97 & .3570 & .5789 & 22.86 & .7014 & .2895 & \cellcolor{second!35}25.77 & \cellcolor{best!35}.6436 & \cellcolor{second!35}.2465 & 16.78 & .4827 & .4361 & 17.61 & .5056 & .4272 & 20.00 & \cellcolor{third!35}.5381 & .3956 \\
			Deblur-NeRF~\cite{ma2022deblurnerf} &  \cellcolor{third!35}17.26 &  \cellcolor{third!35}.3740 &  \cellcolor{third!35}.5088 &  \cellcolor{third!35}25.51 &  \cellcolor{third!35}.7767 &  \cellcolor{third!35}.1897 &  23.38 &  .5068 &  .2649 &  15.91 &  .4209 &  .4198 &  17.91 &  .5213 &  .3661 &  19.99 &  .5199 &  .3499 \\
			DP-NeRF~\cite{lee2023dpnerf} &  \cellcolor{best!35}19.20 &  \cellcolor{best!35}.5175 &  \cellcolor{best!35}.3813 &  21.50 &  .6242 &  .1980 &  21.85 &  .4235 &  .3117 &  \cellcolor{third!35}17.30 &  \cellcolor{second!35}.5265 &  \cellcolor{second!35}.3381 &  \cellcolor{best!35}19.35 &  \cellcolor{second!35}.5765 &  \cellcolor{best!35}.3084 &  19.84 &  .5336 &  \cellcolor{second!35}.3075 \\
			RegNeRF~\cite{niemeyer2022regnerf}~(No kernel) & 16.32 & .3441 & .5876 & 23.25 & .7155 & .2663 & 16.21 & .2040 & .6451 & \cellcolor{second!35}17.35 & \cellcolor{third!35}.4915 & .4509 & \cellcolor{third!35}18.81 & .5447 & .4020 & 18.39 & .4600 & .4704 \\
			RegNeRF~\cite{niemeyer2022regnerf}~(w/\textcolor[rgb]{0.13, 0.55, 0.13}{DP-kernel}) &  16.71 & .3497 & .4683 & 21.37 & .6471 &  .1802 &  15.17 &  .1740 & .6666 &  10.27 &  .1165 &  .6911 &  18.66 &  .5412 &  .4067 &  16.44 &  .3657 &  .4826 \\
			Sparse-DeRF~(w/\textcolor[rgb]{0.25, 0.0, 1.0}{DN-kernel})~-~Ours &  16.91 &  .3693 &  .5357 &  \cellcolor{second!35}25.79 &  \cellcolor{second!35}.7872 &  \cellcolor{second!35}.1801 &  \cellcolor{third!35}25.48 &  \cellcolor{third!35}.6042 &  \cellcolor{best!35}.2405 &  16.58 &  .4705 &  \cellcolor{third!35}.3793 &  18.10 &  \cellcolor{third!35}.5514 &  \cellcolor{third!35}.3414 &  \cellcolor{second!35}20.57 &  \cellcolor{second!35}.5565 &  \cellcolor{third!35}.3354 \\
			Sparse-DeRF~(w/\textcolor[rgb]{0.13, 0.55, 0.13}{DP-kernel})~-~Ours &  \cellcolor{second!35}18.99 &  \cellcolor{second!35}.4770 &  \cellcolor{second!35}.4034 &  \cellcolor{best!35}26.51 &  \cellcolor{best!35}.8051 &  \cellcolor{best!35}.1627 &  23.33 &  .4888 &  .2733 &  \cellcolor{best!35}17.51 &  \cellcolor{best!35}.5371 &  \cellcolor{best!35}.3356 &  \cellcolor{second!35}18.92 &  \cellcolor{best!35}.5799 &  \cellcolor{second!35}.3126 &  \cellcolor{best!35}21.05 &  \cellcolor{best!35}.5776 &  \cellcolor{best!35}.2975 \\
			\midrule\midrule
			
			\multirow{2}{*}{Real Scene}& \multicolumn{3}{c}{Ball} & \multicolumn{3}{c}{Basket} & \multicolumn{3}{c}{Buick} & \multicolumn{3}{c}{Coffee} & \multicolumn{3}{c}{Decoration} & & \\
			 & PSNR($\uparrow$) & SSIM($\uparrow$) & LPIPS($\downarrow$) & PSNR($\uparrow$) & SSIM($\uparrow$) & LPIPS($\downarrow$) & PSNR($\uparrow$) & SSIM($\uparrow$) & LPIPS($\downarrow$) & PSNR($\uparrow$) & SSIM($\uparrow$) & LPIPS($\downarrow$) & PSNR($\uparrow$) & SSIM($\uparrow$) & LPIPS($\downarrow$) &  & &\\
			\midrule
			Naive NeRF~\cite{mildenhall2021nerf} & 21.61 & .5043 & .4992 & 17.23 & .4384 & .4746 & 19.20 & .5508 & .3579 & 23.21 & .7341 & .3281 & 14.78 & .3299 & .5640 &  &  &  \\
			MPR~\cite{zamir2021MPR}+NeRF & 21.46 & .5046 & .4992 & 16.98 & .4260 & .4766 & \cellcolor{third!35}19.39 & \cellcolor{third!35}.5486 & .3629 & 22.98 & .7267 & .3284 & 14.88 & \cellcolor{third!35}.3460 & .5532 &  &  &  \\
			Deblur-NeRF~\cite{ma2022deblurnerf} &  21.90 &  .5182 &  .4609 &  17.15 &  .4331 & .4316 &  19.34 &  .5278 &  .3407 &  23.10 &  .7302 &  .2873 &  14.94 &  .3201 &  \cellcolor{third!35}.5471 &   &  &   \\
			DP-NeRF~\cite{lee2023dpnerf} &  \cellcolor{second!35}23.20 &  \cellcolor{second!35}.5842 &  \cellcolor{second!35}.3847 &  17.42 &  .4046 &  .4658 &  19.26 &  .5323 &  \cellcolor{third!35}.3374 & 24.05 & .7597 &  \cellcolor{third!35}.2624 &  \cellcolor{third!35}14.98 &  .3018 &  .5618 &   &   &   \\
			RegNeRF~\cite{niemeyer2022regnerf}~(No kernel) & 21.16 & .5100 & .4952 & \cellcolor{third!35}18.12 & \cellcolor{third!35}.4711 & .4569 & \cellcolor{best!35}20.11 & \cellcolor{best!35}.5821 & .3412 & \cellcolor{third!35}26.14 & \cellcolor{third!35}.7837 & .2965 & 11.35 & .1045 & .7030 \\
			RegNeRF~\cite{niemeyer2022regnerf}~(w/\textcolor[rgb]{0.13, 0.55, 0.13}{DP-kernel}) &  19.13 & .4391 & \cellcolor{third!35}.4042 & 17.42 & .4165 &  \cellcolor{second!35}.3746 &  13.31 &  .2697 & .4715 &  16.40 &  .4684 &  .5239 &  11.07 &  .1058 &  .7089 &  &  &  \\
			Sparse-DeRF~(w/\textcolor[rgb]{0.25, 0.0, 1.0}{DN-kernel})~-~Ours &  \cellcolor{third!35}22.28 &  \cellcolor{third!35}.5506 &  .4334 &  \cellcolor{second!35}18.69 &  \cellcolor{second!35}.5058 &  \cellcolor{third!35}.3774 &  19.28 & .5414 &  \cellcolor{second!35}.3371 &  \cellcolor{second!35}26.76 &  \cellcolor{second!35}.8126 &  \cellcolor{second!35}.2419 &  \cellcolor{best!35}17.12 &  \cellcolor{best!35}.4192 &  \cellcolor{best!35}.4761 &   &   &   \\
			Sparse-DeRF~(w/\textcolor[rgb]{0.13, 0.55, 0.13}{DP-kernel})~-~Ours &  \cellcolor{best!35}23.39 &  \cellcolor{best!35}.6010 &  \cellcolor{best!35}.3806 &  \cellcolor{best!35}20.41 &  \cellcolor{best!35}.5451 &  \cellcolor{best!35}.3305 &  \cellcolor{second!35}19.48 &  \cellcolor{second!35}.5531 &  \cellcolor{best!35}.3251 &  \cellcolor{best!35}27.77 &  \cellcolor{best!35}.8364 &  \cellcolor{best!35}.2049 &  \cellcolor{second!35}16.33 &  \cellcolor{second!35}.3914 &  \cellcolor{second!35}.4950 &   &   &   \\
			\midrule
			\multirow{2}{*}{Real Scene}& \multicolumn{3}{c}{Girl} & \multicolumn{3}{c}{Heron} & \multicolumn{3}{c}{Parterre} & \multicolumn{3}{c}{Puppet} & \multicolumn{3}{c}{Stair} & \multicolumn{3}{c}{Average} \\
			 & PSNR($\uparrow$) & SSIM($\uparrow$) & LPIPS($\downarrow$) & PSNR($\uparrow$) & SSIM($\uparrow$) & LPIPS($\downarrow$) & PSNR($\uparrow$) & SSIM($\uparrow$) & LPIPS($\downarrow$) & PSNR($\uparrow$) & SSIM($\uparrow$) & LPIPS($\downarrow$) & PSNR($\uparrow$) & SSIM($\uparrow$) & LPIPS($\downarrow$) & PSNR($\uparrow$) & SSIM($\uparrow$) & LPIPS($\downarrow$) \\
			\midrule
			Naive NeRF~\cite{mildenhall2021nerf} & \cellcolor{third!35}15.91 & \cellcolor{third!35}.5659 & \cellcolor{third!35}.4161 & 18.83 & .4263 & .4086 & \cellcolor{third!35}20.64 & \cellcolor{third!35}.4857 & .4822 & \cellcolor{second!35}18.15 & \cellcolor{second!35}.4619 & .4470 & 20.27 & .3622 & .5033 & \cellcolor{third!35}18.98 & \cellcolor{third!35}.4860 & .4481 \\
			MPR~\cite{zamir2021MPR}+NeRF & 15.60 & .5617 & .4224 & \cellcolor{third!35}18.91 & .4217 & .4173 & \cellcolor{best!35}20.76 & \cellcolor{second!35}.4893 & .4781 & 17.96 & .4548 & .4468 & 19.94 & .3493 & .4993 & 18.89 & .4829 & .4484 \\
			Deblur-NeRF~\cite{ma2022deblurnerf} &  15.62 &  .5351 &  .4206 &  18.85 &  \cellcolor{third!35}.4350 &  .3398 &  19.71 &  .4318 &  \cellcolor{second!35}.4549 &  17.77 &  .4400 &  .4358 & 20.49 & .3901 &  \cellcolor{second!35}.4326 &  18.89 &  .4761&  \cellcolor{third!35}.4151 \\
			DP-NeRF~\cite{lee2023dpnerf} &  15.40 &  .5237 &  .4318 &  18.68 &  .4292 &  \cellcolor{third!35}.3352 &  17.62 &  .2987 &  .4991 &  17.61 &  .4369 &  \cellcolor{third!35}.4313 &  19.44 &  .3107 &  .4657 &  18.77 &  .4582 & .4175 \\
			RegNeRF~\cite{niemeyer2022regnerf}~(No kernel) & 10.24 & .1511 & .7199 & 18.86 & .4235 & .4194 & 18.68 & .4238 & .5163 & \cellcolor{best!35}18.66 &\cellcolor{best!35} .4788 & .4327 & \cellcolor{second!35}21.06 & \cellcolor{second!35}.3975 & .4706 & 18.44 & .4326 & .4852 \\
			RegNeRF~\cite{niemeyer2022regnerf}~(w/\textcolor[rgb]{0.13, 0.55, 0.13}{DP-kernel}) &  9.52 & .1304 & .7119 & 11.56 & .0595 &  .6149 &  13.61 &  .1721 & .7245 &  11.10 &  .1342 &  .6914 &  12.73 &  .0249 &  .6610 &  13.59 &  .2221 &  .5887 \\
			Sparse-DeRF~(w/\textcolor[rgb]{0.25, 0.0, 1.0}{DN-kernel})~-~Ours &  \cellcolor{best!35}16.95 &  \cellcolor{best!35}.5975 &  \cellcolor{best!35}.3823 &  \cellcolor{second!35}19.04 &  \cellcolor{second!35}.4558 &  \cellcolor{second!35}.3315 &  \cellcolor{second!35}20.74 &  \cellcolor{best!35}.4916 &  \cellcolor{best!35}.4323 &  \cellcolor{third!35}18.11 &  \cellcolor{third!35}.4616 &  \cellcolor{best!35}.4202 &  \cellcolor{third!35}20.80 &  \cellcolor{third!35}.3948 &  \cellcolor{third!35}.4386 &  \cellcolor{second!35}19.98 &  \cellcolor{best!35}.5231 &  \cellcolor{second!35}.3871 \\
			Sparse-DeRF~(w/\textcolor[rgb]{0.13, 0.55, 0.13}{DP-kernel})~-~Ours &  \cellcolor{second!35}16.47 &  \cellcolor{second!35}.5855 &  \cellcolor{second!35}.3835 &  \cellcolor{best!35}19.09 &  \cellcolor{best!35}.4624 &  \cellcolor{best!35}.3194 &  18.36 &  .3375 &  \cellcolor{third!35}.4715 &  17.97 & .4612 &  \cellcolor{second!35}.4206 &  \cellcolor{best!35}21.25 &  \cellcolor{best!35}.4045 &  \cellcolor{best!35}.4044 &  \cellcolor{best!35}20.05 &  \cellcolor{second!35}.5178 &  \cellcolor{best!35}.3736 \\
			\midrule
      \end{tabular}
      }
   \label{tab:appendix_quantitative_results_4view}
   \end{center}
\end{table*}

\begin{table*}[t]
   \Large
   \begin{center}
   \caption{Quantitative results of novel view synthesis for the entire scenes of synthetic and real scenes obtained from \textbf{6-view} settings. Each color shading represents the \colorbox{best!35}{best}, \colorbox{second!35}{second best} and \colorbox{third!35}{third best} result, respectively. \textcolor[rgb]{0.13, 0.55, 0.13}{DP-kernel} and \textcolor[rgb]{0.25, 0.0, 1.0}{DN-kernel} denote the kernel of~\cite{lee2023dpnerf} and~\cite{ma2022deblurnerf}.}
      \resizebox{2\columnwidth}{!}{
		\centering
		\setlength{\tabcolsep}{1pt}
         \begin{tabular}{c||ccc|ccc|ccc|ccc|ccc||ccc}
			\midrule
			\multirow{2}{*}{Synthetic Scene}& \multicolumn{3}{c}{Factory} & \multicolumn{3}{c}{Cozyroom} & \multicolumn{3}{c}{Pool} & \multicolumn{3}{c}{Tanabata} & \multicolumn{3}{c}{Trolley} & \multicolumn{3}{c}{Average} \\
			 & PSNR($\uparrow$) & SSIM($\uparrow$) & LPIPS($\downarrow$) & PSNR($\uparrow$) & SSIM($\uparrow$) & LPIPS($\downarrow$) & PSNR($\uparrow$) & SSIM($\uparrow$) & LPIPS($\downarrow$) & PSNR($\uparrow$) & SSIM($\uparrow$) & LPIPS($\downarrow$) & PSNR($\uparrow$) & SSIM($\uparrow$) & LPIPS($\downarrow$) & PSNR($\uparrow$) & SSIM($\uparrow$) & LPIPS($\downarrow$) \\
			\midrule
			Naive NeRF~\cite{mildenhall2021nerf} &17.12 &.3663 & .5663 & 23.12 & .7160 & .2713 & \cellcolor{best!35}28.42 & \cellcolor{best!35}.7463 & .2106 & 19.92 & .5889 & .3866 & 19.68 & .5749 & .3853 & 21.65 & .5985 & .3638 \\
			MPR~\cite{zamir2021MPR}+NeRF & 17.21 & .3754 & .5649 & 23.06 & .7087 & .2749 & \cellcolor{second!35}28.38 & \cellcolor{second!35}.7461 & .2093 & 20.23 & .5966 & .3783 & 19.71 & .5728 & .3870 & 21.72 & .5999 & .3629 \\
			Deblur-NeRF~\cite{ma2022deblurnerf} & 18.77 & .5048 & .3890 & 26.67 & .8214 & .1475 & 27.50 & .7116 & \cellcolor{third!35}.1783 & 21.10 & .6693 & .2517 & \cellcolor{third!35}21.58 & \cellcolor{third!35}.6921 & \cellcolor{third!35}.2263 &  23.12 &  .6798 &  .2386 \\
			DP-NeRF~\cite{lee2023dpnerf} & \cellcolor{best!35}21.63 & \cellcolor{best!35}.6402 & \cellcolor{best!35}.2984 & \cellcolor{second!35}27.63 & \cellcolor{best!35}.8475 & \cellcolor{best!35}.1224 & 25.36 & .6227 & .1861 & 21.27 & .6818 & \cellcolor{third!35}.2023 & \cellcolor{best!35}22.49 & \cellcolor{best!35}.7259 & \cellcolor{best!35}.1899  & \cellcolor{third!35}23.68  &  \cellcolor{second!35}.7036 &  \cellcolor{best!35}.1998 \\
			RegNeRF~\cite{niemeyer2022regnerf}~(No kernel) & 17.04 & .3690 & .5729 & 23.40 & .7205 & .2649 & 18.05 & .3610 & .4870 & 20.33 & .5980 & .3745 & 19.63 & .5760 & .3833 & 19.69 & .5249 & .4165 \\
			RegNeRF~\cite{niemeyer2022regnerf}~(w/\textcolor[rgb]{0.13, 0.55, 0.13}{DP-kernel}) & \cellcolor{third!35}21.03 & \cellcolor{second!35}.6273 & \cellcolor{second!35}.3076 & \cellcolor{best!35}27.73 & \cellcolor{second!35}.8455 & \cellcolor{second!35}.1238 & 18.04 & .3173 & .4987 & \cellcolor{third!35}21.84 & \cellcolor{best!35}.7246 & \cellcolor{second!35}.2022 & 19.36 & .5661 & .3908 & 21.60  &  .6162 &  .3046 \\
			Sparse-DeRF~(w/\textcolor[rgb]{0.25, 0.0, 1.0}{DN-kernel})~-~Ours & 18.33 & .5144 & .4004 & 26.63 & .8184 & .1513 & 28.16 & .7344 & \cellcolor{second!35}.1755 & \cellcolor{second!35}22.06 & \cellcolor{third!35}.6988 & .2245 & 21.43 & .6855 & .2380 & \cellcolor{second!35}23.32 &  \cellcolor{third!35}.6903 &  \cellcolor{third!35}.2379 \\
			Sparse-DeRF~(w/\textcolor[rgb]{0.13, 0.55, 0.13}{DP-kernel})~-~Ours & \cellcolor{second!35}21.29 & \cellcolor{third!35}.6179 & \cellcolor{third!35}.3261 & \cellcolor{third!35}27.34 & \cellcolor{third!35}.8340 & \cellcolor{third!35}.1298 & \cellcolor{third!35}28.19 & \cellcolor{third!35}.7365 & \cellcolor{best!35}.1606 & \cellcolor{best!35}22.30 & \cellcolor{second!35}.7233 & \cellcolor{best!35}.2017 & \cellcolor{second!35}22.21 & \cellcolor{second!35}.7159 & \cellcolor{second!35}.2036  &  \cellcolor{best!35}24.27 &  \cellcolor{best!35}.7255 &  \cellcolor{second!35}.2044 \\
			\midrule\midrule
			
			\multirow{2}{*}{Real Scene}& \multicolumn{3}{c}{Ball} & \multicolumn{3}{c}{Basket} & \multicolumn{3}{c}{Buick} & \multicolumn{3}{c}{Coffee} & \multicolumn{3}{c}{Decoration} & & \\
			 & PSNR($\uparrow$) & SSIM($\uparrow$) & LPIPS($\downarrow$) & PSNR($\uparrow$) & SSIM($\uparrow$) & LPIPS($\downarrow$) & PSNR($\uparrow$) & SSIM($\uparrow$) & LPIPS($\downarrow$) & PSNR($\uparrow$) & SSIM($\uparrow$) & LPIPS($\downarrow$) & PSNR($\uparrow$) & SSIM($\uparrow$) & LPIPS($\downarrow$) &  & &\\
			\midrule
			Naive NeRF~\cite{mildenhall2021nerf} & 22.12 & .5359 & .4818 & 21.39 & .6253 & .3328 & 21.21 & .6197 & .3100 & 22.73 & .7163 & .3397 & \cellcolor{third!35}19.14 & \cellcolor{third!35}.5118 & .4460\\
			MPR~\cite{zamir2021MPR}+NeRF & 22.15 & .5365 & .4745 & 21.39 & .6261 & .3294 & 21.11 & .6192 & .3063 & 22.56 & .7087 & .3435 & 18.31 & .4837 & .4735\\
			Deblur-NeRF~\cite{ma2022deblurnerf} & 24.47 & .6473 & .3342 & \cellcolor{best!35}23.63 & \cellcolor{best!35}.7204 & \cellcolor{best!35}.2108 & 21.45 & .6285 & .2625 & 23.70 & .7604 & .2588 & 17.74 & .4790 & \cellcolor{third!35}.4296 \\
			DP-NeRF~\cite{lee2023dpnerf} & \cellcolor{third!35}24.73 & \cellcolor{second!35}.6651 & \cellcolor{second!35}.3107 & 23.24 & .6643 & .2494 & \cellcolor{second!35}21.65 & \cellcolor{second!35}.6418 & \cellcolor{best!35}.2445 & 25.51 & \cellcolor{third!35}.7949 & \cellcolor{third!35}.2170 & 17.48 & .4725 & .4457 \\
			RegNeRF~\cite{niemeyer2022regnerf}~(No kernel) & 21.98 & .5312 & .4677 & 21.67 & .6129 & .3406 & 21.16 & .6215 & .3051 & \cellcolor{third!35}25.67 & .7716 & .3105 & 12.45 & .1569 & .7061 \\
			RegNeRF~\cite{niemeyer2022regnerf}~(w/\textcolor[rgb]{0.13, 0.55, 0.13}{DP-kernel}) & \cellcolor{best!35}25.02 & \cellcolor{best!35}.6719 & \cellcolor{best!35}.2953 & 22.95 & .6554 & .2501 & \cellcolor{third!35}21.57 & \cellcolor{best!35}.6456 & \cellcolor{second!35}.2462 & 23.38 & .7393 & .2341 & 11.43 & .1247 & .6932 \\
			Sparse-DeRF~(w/\textcolor[rgb]{0.25, 0.0, 1.0}{DN-kernel})~-~Ours & 23.76 & .6184 & .3607 & \cellcolor{second!35}23.39 & \cellcolor{second!35}.7155 & \cellcolor{second!35}.2329 & \cellcolor{best!35}21.66 & \cellcolor{third!35}.6405 & \cellcolor{third!35}.2585 & \cellcolor{second!35}27.84 & \cellcolor{second!35}.8388 & \cellcolor{second!35}.2068 & \cellcolor{best!35}19.98 & \cellcolor{best!35}.5553 & \cellcolor{best!35}.3644\\
			Sparse-DeRF~(w/\textcolor[rgb]{0.13, 0.55, 0.13}{DP-kernel})~-~Ours & \cellcolor{second!35}24.80 & \cellcolor{third!35}.6643 & \cellcolor{third!35}.3200 & \cellcolor{third!35}23.32 & \cellcolor{third!35}.6789 & \cellcolor{third!35}.2455 & 21.27 & .6234 & .2662 & \cellcolor{best!35}28.91 & \cellcolor{best!35}.8525 & \cellcolor{best!35}.1902 & \cellcolor{second!35}19.69 & \cellcolor{second!35}.5440 & \cellcolor{second!35}.3672 \\
			\midrule
			\multirow{2}{*}{Real Scene}& \multicolumn{3}{c}{Girl} & \multicolumn{3}{c}{Heron} & \multicolumn{3}{c}{Parterre} & \multicolumn{3}{c}{Puppet} & \multicolumn{3}{c}{Stair} & \multicolumn{3}{c}{Average} \\
			 & PSNR($\uparrow$) & SSIM($\uparrow$) & LPIPS($\downarrow$) & PSNR($\uparrow$) & SSIM($\uparrow$) & LPIPS($\downarrow$) & PSNR($\uparrow$) & SSIM($\uparrow$) & LPIPS($\downarrow$) & PSNR($\uparrow$) & SSIM($\uparrow$) & LPIPS($\downarrow$) & PSNR($\uparrow$) & SSIM($\uparrow$) & LPIPS($\downarrow$) & PSNR($\uparrow$) & SSIM($\uparrow$) & LPIPS($\downarrow$) \\
			\midrule
			Naive NeRF~\cite{mildenhall2021nerf} & 18.43 & .6500 & .3335 & 18.91 & .4204 & .4307 & 21.82 & .5487 & .4351 & 19.52 & .5242 & .3933 & 21.01 & .3602 & .5111 & 20.63 & .5513 & .4014 \\
			MPR~\cite{zamir2021MPR}+NeRF & \cellcolor{third!35}18.61 & .6580 & .3324 & 18.97 & .4220 & .4287 & 21.78 & .5477 & .4358 & 19.84 & .5357 & .3793 & 21.27 & .3758 & .4070 & 20.60 & .5513 & .4010 \\
			Deblur-NeRF~\cite{ma2022deblurnerf} & 18.09 & .6582 & .3055 & 19.40 & .4709 & .3166 & 21.92 & .5557 & .3872 & 20.58 & .5834 & .3226 & 22.60 & .4993 & \cellcolor{third!35}.3349 & 21.36 & .6003 & .3163 \\
			DP-NeRF~\cite{lee2023dpnerf} & 18.16 & \cellcolor{third!35}.6622 & \cellcolor{third!35}.3020 & \cellcolor{third!35}19.58 & \cellcolor{third!35}.4991 & \cellcolor{second!35}.2905 & \cellcolor{second!35}22.36 & \cellcolor{second!35}.5879 & \cellcolor{best!35}.3350 & \cellcolor{second!35}20.65 & \cellcolor{second!35}.5868 & \cellcolor{best!35}.2988 & \cellcolor{third!35}23.40 & \cellcolor{best!35}.5624 & \cellcolor{best!35}.2986 & \cellcolor{third!35}21.68 & \cellcolor{third!35}.6137 & \cellcolor{second!35}.2992 \\
			RegNeRF~\cite{niemeyer2022regnerf}~(No kernel) & 11.06 & .1966 & .7091 & 18.96 & .4218 & .4292 & 21.83 & .5526 & .4353 & 20.07 & .5309 & .3827 & 21.69 & .3938 & .4963 & 19.65 & .4790 & .4583 \\
			RegNeRF~\cite{niemeyer2022regnerf}~(w/\textcolor[rgb]{0.13, 0.55, 0.13}{DP-kernel}) & 10.38 & .1434 & .7200 & 12.83 & .1019 & .5757 & 21.34 & .5159 & \cellcolor{third!35}.3415 & 20.49 & \cellcolor{third!35}.5799 & \cellcolor{second!35}.3113 & 13.08 & .2613 & .6581 & 18.25 & .4439 & .4326 \\
			Sparse-DeRF~(w/\textcolor[rgb]{0.25, 0.0, 1.0}{DN-kernel})~-~Ours & \cellcolor{second!35}18.86 & \cellcolor{second!35}.6860 & \cellcolor{second!35}.2769 & \cellcolor{best!35}19.63 & \cellcolor{second!35}.5011 & \cellcolor{third!35}.2992 & \cellcolor{third!35}22.15 & \cellcolor{third!35}.5703 & .3787 & \cellcolor{best!35}20.81 & \cellcolor{best!35}.5882 & .3149 & \cellcolor{second!35}23.42 & \cellcolor{third!35}.5340 & .3374 & \cellcolor{second!35}22.15 & \cellcolor{second!35}.6248 & \cellcolor{third!35}.3030 \\
			Sparse-DeRF~(w/\textcolor[rgb]{0.13, 0.55, 0.13}{DP-kernel})~-~Ours & \cellcolor{best!35}18.96 & \cellcolor{best!35}.6887 & \cellcolor{best!35}.2744 & \cellcolor{second!35}19.59 & \cellcolor{best!35}.5051 & \cellcolor{best!35}.2808 & \cellcolor{best!35}22.57 & \cellcolor{best!35}.5994 & \cellcolor{second!35}.3317 & \cellcolor{third!35}20.60 & .5756 & \cellcolor{third!35}.3124 & \cellcolor{best!35}23.48 & \cellcolor{second!35}.5509 & \cellcolor{second!35}.3189 & \cellcolor{best!35}22.32 & \cellcolor{best!35}.6283 & \cellcolor{best!35}.2907 \\
			\midrule
      \end{tabular}
      }
   \label{tab:appendix_quantitative_results_6view}
   \end{center}
\end{table*}

\begin{table*}[t]
   \Large
   \begin{center}
   \caption{Ablation quantitative results for the entire scenes of synthetic and real scenes from \textbf{2-view} settings. Each color shading represents the \colorbox{best!35}{best}, \colorbox{second!35}{second best} and \colorbox{third!35}{third best} result, respectively.}
      \resizebox{2\columnwidth}{!}{
		\centering
		\setlength{\tabcolsep}{1pt}
         \begin{tabular}{c|C{1.1cm}C{1.1cm}C{1.1cm}||ccc|ccc|ccc|ccc|ccc||ccc}
			\midrule		
			\multirow{2}{*}{Kernel Type}& \multicolumn{3}{c}{Synthetic Scene} & \multicolumn{3}{c}{Factory} & \multicolumn{3}{c}{Cozyroom} & \multicolumn{3}{c}{Pool} & \multicolumn{3}{c}{Tanabata} & \multicolumn{3}{c}{Trolley} & \multicolumn{3}{c}{Average} \\
			 & \textit{SS} & \textit{MGS} & \textit{PD} & PSNR($\uparrow$) & SSIM($\uparrow$) & LPIPS($\downarrow$) & PSNR($\uparrow$) & SSIM($\uparrow$) & LPIPS($\downarrow$) & PSNR($\uparrow$) & SSIM($\uparrow$) & LPIPS($\downarrow$) & PSNR($\uparrow$) & SSIM($\uparrow$) & LPIPS($\downarrow$) & PSNR($\uparrow$) & SSIM($\uparrow$) & LPIPS($\downarrow$) & PSNR($\uparrow$) & SSIM($\uparrow$) & LPIPS($\downarrow$) \\
			\midrule
			\multirow{7}{*}{\textcolor[rgb]{0.13, 0.55, 0.13}{DP-kernel}}& & & &  14.14 &  \cellcolor{third!35}.2091 &  .6540 &  \cellcolor{best!35}20.71 &  \cellcolor{best!35}.5752 &  .3487 &  18.38 &  .3276 &  .4127 &  11.21 &  .1482 &  .6247 &  10.88 &  .1536 &  .6543 &  15.06 &  .2827 &  .5389 \\
			& \ding{51} & & & 14.09 & .2008 & .6550 & 19.57 & .5321 & .3544 & 18.49 & .3178 & .4271 & 11.21 & .1552 & .6243 & 11.04 & \cellcolor{second!35}.1801 & \cellcolor{third!35}.6331 & 14.88 & .2772 & .5388 \\
			& & \ding{51} & & \cellcolor{best!35}14.33 & .2088 & \cellcolor{third!35}.6496 & 18.89 & .5032 & .3577 & \cellcolor{third!35}20.30 & \cellcolor{third!35}.3467 & \cellcolor{second!35}.4017 & \cellcolor{second!35}11.69 & \cellcolor{second!35}.1748 & \cellcolor{third!35}.6067 & \cellcolor{best!35}11.78 & \cellcolor{best!35}.1948 & \cellcolor{second!35}.6289 & \cellcolor{second!35}15.40 & \cellcolor{third!35}.2857 & \cellcolor{third!35}.5289 \\
			& & & \ding{51} &  \cellcolor{third!35}14.19 &  .2035 &  .6620 &  \cellcolor{third!35}19.82 &  \cellcolor{third!35}.5404 &  \cellcolor{third!35}.3451 &  18.96 &  .3350 &  .4125 &  11.16 &  .1457 &  .6408 &  11.02 &  .1733 &  .6350 &  15.03 &  .2796 &  .5391 \\
			& \ding{51} & & \ding{51} &  13.95 &  .1963 &  .6629 &  18.81 &  .5010 &  .3637 &  18.57 &  .3404 &  .4148 &  11.00 &  .1501 &  .6259 &  10.25 &  .1357 &  .6730 &  14.52 &  .2647 &  .5481 \\
			& \ding{51} & \ding{51} & &  \cellcolor{second!35}14.24 &  \cellcolor{best!35}.2228 &  \cellcolor{second!35}.6439 &  \cellcolor{second!35}20.16 &  \cellcolor{second!35}.5560 &  \cellcolor{best!35}.3391 &  \cellcolor{best!35}20.37 &  \cellcolor{best!35}.3526 &  \cellcolor{best!35}.4004 &  \cellcolor{third!35}11.55 &  \cellcolor{third!35}.1710 &  \cellcolor{second!35}.5999 &  \cellcolor{third!35}11.16 &  .1744 &  \cellcolor{best!35}.6241 &  \cellcolor{best!35}15.50 &  \cellcolor{best!35}.2954 &  \cellcolor{best!35}.5215 \\
			& \ding{51} & \ding{51} & \ding{51} &  14.10 &  \cellcolor{second!35}.2116 &  \cellcolor{best!35}.6413 &  18.97 &  .5206 &  \cellcolor{second!35}.3417 &  \cellcolor{second!35}20.32 &  \cellcolor{second!35}.3488 &  \cellcolor{third!35}.4056 & \cellcolor{best!35}12.00 &  \cellcolor{best!35}.1943 &  \cellcolor{best!35}.5902 &  \cellcolor{second!35}11.36 &  \cellcolor{third!35}.1769 &  .6422 &  \cellcolor{third!35}15.35 & \cellcolor{second!35}.2904 & \cellcolor{second!35}.5242 \\
			\midrule
			\multirow{7}{*}{\textcolor[rgb]{0.25, 0.0, 1.0}{DN-kernel}}& & & & 14.14 &  .2009 &  .6647 &  \cellcolor{third!35}20.58 &  \cellcolor{third!35}.5768 &  \cellcolor{second!35}.3443 &  18.37 &  .3247 & .4061 &  \cellcolor{second!35}11.56 & .1704 & .6075 &  11.05 &  .1690 & .6425 &  \cellcolor{third!35}15.14 &  .2884 &  .5330 \\
			& \ding{51} & & & \cellcolor{third!35}14.21 & .2177 & .6650 & 20.33 & .5663 & .3591 & 18.74 & .3200 & .4093 & 11.13 & .1547 & .6191 & 10.78 & .1554 & .6566 & 15.04 & .2828 & .5418 \\
			& & \ding{51} & & 14.20 & .2107 & .6686 & \cellcolor{best!35}21.02 & \cellcolor{best!35}.5999 & \cellcolor{best!35}.3326 & 19.64 & \cellcolor{third!35}.3330 & \cellcolor{third!35}.3950 & \cellcolor{third!35}11.55 & \cellcolor{second!35}.1804 & \cellcolor{best!35}.5903 & \cellcolor{second!35}11.48 & \cellcolor{second!35}.1883 & .6351 & 15.12 & \cellcolor{best!35}.3025 & \cellcolor{best!35}.5243 \\
			& & & \ding{51} &  14.11 &  .2096 &  \cellcolor{second!35}.6588 &  18.78 &  .5094 &  .3928 &  \cellcolor{third!35}19.76 &  \cellcolor{best!35}.3464 &  \cellcolor{best!35}.3901 &  10.58 &  .1206 &  .6565 &  10.99 &  .1648 &  \cellcolor{best!35}.6289 &  14.84 &  .2702 &  .5454 \\
			& \ding{51} & & \ding{51} &  \cellcolor{second!35}14.24 &  \cellcolor{best!35}.2222 &  \cellcolor{third!35}.6619 &  \cellcolor{best!35}21.02 &  \cellcolor{second!35}.5918 &  \cellcolor{third!35}.3544 &  18.77 &  .3197 &  .4115 &  11.12 &  .1642 &  .6217 &  10.41 &  .1467 &  .6452 &  15.11 &  .2889 &  .5389 \\
			& \ding{51} & \ding{51} & &  14.14 &  \cellcolor{third!35}.2189 &  .6622 &  20.10 &  .5555 &  .3761 &  \cellcolor{best!35}20.11 &  \cellcolor{second!35}.3372 &  \cellcolor{second!35}.3939 &  11.49 &  \cellcolor{third!35}.1753 &  \cellcolor{third!35}.5975 &  \cellcolor{best!35}11.99 &  \cellcolor{best!35}.2026 &  \cellcolor{third!35}.6299 &  \cellcolor{best!35}15.57 &  \cellcolor{second!35}.2979 &  \cellcolor{third!35}.5319 \\
			& \ding{51} & \ding{51} & \ding{51} &  \cellcolor{best!35}14.27 &  \cellcolor{second!35}.2200 &  \cellcolor{best!35}.6564 &  20.57 &  .5675 &  .3589 &  \cellcolor{second!35}19.81 &  \cellcolor{third!35}.3330 & .4058 & \cellcolor{best!35}11.67 &  \cellcolor{best!35}.1827 &  \cellcolor{second!35}.5947 & \cellcolor{third!35}11.28 & \cellcolor{third!35}.1800 &  \cellcolor{second!35}.6298 &  \cellcolor{second!35}15.52 & \cellcolor{third!35}.2966 &  \cellcolor{second!35}.5291 \\
			\midrule\midrule
			
			& \multicolumn{3}{c}{Real Scene}& \multicolumn{3}{c}{Ball} & \multicolumn{3}{c}{Basket} & \multicolumn{3}{c}{Buick} & \multicolumn{3}{c}{Coffee} & \multicolumn{3}{c}{Decoration} & & \\
			& \textit{SS} & \textit{MGS} & \textit{PD} & PSNR($\uparrow$) & SSIM($\uparrow$) & LPIPS($\downarrow$) & PSNR($\uparrow$) & SSIM($\uparrow$) & LPIPS($\downarrow$) & PSNR($\uparrow$) & SSIM($\uparrow$) & LPIPS($\downarrow$) & PSNR($\uparrow$) & SSIM($\uparrow$) & LPIPS($\downarrow$) & PSNR($\uparrow$) & SSIM($\uparrow$) & LPIPS($\downarrow$) &  & &\\
			\midrule
			\multirow{7}{*}{\textcolor[rgb]{0.13, 0.55, 0.13}{DP-kernel}}& & & &  18.22 &  .4200 &  .5611 &  13.24 &  .2263 &  .6062 &  13.43 &  .2480 &  .5571 &  18.15 &  .5191 &  .4876 &  12.34 &  .1672 & .6675 &   &   &   \\
			& \ding{51} & & & 18.76 & .4329 & .5521 & 12.68 & .2189 & .6301 & 13.05 & .2269 & .5657 & 17.51 & .4792 & .4907 & 12.28 & .1638 & .6619 & & &\\
			& & \ding{51} & & \cellcolor{second!35}19.80 & \cellcolor{second!35}.4544 & \cellcolor{second!35}.5273 & \cellcolor{second!35}14.08 & \cellcolor{third!35}.2706 & \cellcolor{best!35}.5509 & \cellcolor{best!35}13.96 & \cellcolor{best!35}.3028 & \cellcolor{best!35}.4992 & \cellcolor{best!35}19.72 & \cellcolor{best!35}.6115 & \cellcolor{best!35}.4035 & \cellcolor{best!35}13.14 & \cellcolor{best!35}.2198 & \cellcolor{second!35}.6201 & & & \\
			& & & \ding{51} &  18.28 &  .4191 &  .5591 &  12.95 &  .2281 &  .5951 &  13.49 &  .2654 &  .5464 &  17.92 &  .5030 &  .4875 &  12.04 & .1491 &  .6700 & & &  \\
			& \ding{51} & & \ding{51} &  18.50 &  .4341 &  .5537 &  12.92 &  .2407 &  .5938 &  13.64 &  .2564 &  .5427 &  18.09 &  .5263 &  .4689 &  12.42 &  .1669 &  .6689 & & & \\
			& \ding{51} & \ding{51} & &  \cellcolor{second!35}19.80 &  \cellcolor{third!35}.4412 &  \cellcolor{third!35}.5346 &  \cellcolor{best!35}14.28 &  \cellcolor{best!35}.2913 &  \cellcolor{second!35}.5520 &  \cellcolor{second!35}13.82 &  \cellcolor{second!35}.2920 &  \cellcolor{second!35}.5013 &  \cellcolor{third!35}18.81 &  \cellcolor{third!35}.5712 &  \cellcolor{third!35}.4363 &  \cellcolor{third!35}13.04 &  \cellcolor{third!35}.2166 &  \cellcolor{third!35}.6203 & & & \\
			& \ding{51} & \ding{51} & \ding{51} &  \cellcolor{best!35}20.09 & \cellcolor{best!35}.4583 &  \cellcolor{best!35}.5244 &  \cellcolor{third!35}14.05 &  \cellcolor{second!35}.2890 &  \cellcolor{third!35}.5555 &  \cellcolor{third!35}13.78 &  \cellcolor{third!35}.2864 &  \cellcolor{third!35}.5036 &  \cellcolor{second!35}19.67 & \cellcolor{second!35}.6084 &  \cellcolor{second!35}.4096 &  \cellcolor{second!35}13.11 &  \cellcolor{second!35}.2192 &  \cellcolor{best!35}.6192 &   &   &  \\
			\midrule
			\multirow{7}{*}{\textcolor[rgb]{0.25, 0.0, 1.0}{DN-kernel}} & & & &  18.80 &  .4325 &  .5521 &  12.86 &  .2414 &  .5947 &  13.11 &  .2446 &  .5609 & 18.66 &  .5272 & .4696 &  12.09 &  .1508 &  .6722 &  &  & \\
			& \ding{51} & & & 18.88 & .4388 & .5519 & 13.03 & .2516 & .6018 & 13.36 & .2533 & .5523 & 18.04 & .5145 & .4847 & 12.38 & .1736 & .6545 & & &\\
			& & \ding{51} & & \cellcolor{third!35}19.95 & \cellcolor{third!35}.4596 & \cellcolor{third!35}.5394 & \cellcolor{third!35}13.52 & \cellcolor{third!35}.2534 & \cellcolor{third!35}.5871 & \cellcolor{third!35}14.11 & \cellcolor{third!35}.3087 & \cellcolor{third!35}.5051 & \cellcolor{third!35}19.33 & \cellcolor{second!35}.6003 & \cellcolor{best!35}.4133 & \cellcolor{best!35}13.13 & \cellcolor{third!35}.2153 & \cellcolor{best!35}.6222 & & & \\
			& & & \ding{51} &  18.70 &  .4235 &  .5673 &  13.24 &  .2505 &  .5931 &  13.52 &  .2520 &  .5438 &  18.15 &  .5031 &  .4949 &  12.32 &  .1683 &  .6617 & & &  \\
			& \ding{51} & & \ding{51} &  18.36 &  .4140 &  .5667 &  13.14 &  .2482 &  .5975 &  13.40 &  .2549 &  .5461 &  17.90 &  .5098 &  .4857 &  12.37 &  .1634 &  .6658 & & & \\
			& \ding{51} & \ding{51} & &  \cellcolor{second!35}20.06 &  \cellcolor{best!35}.4637 &  \cellcolor{second!35}.5299 &  \cellcolor{second!35}13.56 &  \cellcolor{second!35}.2593 &  \cellcolor{best!35}.5828 &  \cellcolor{best!35}14.51 &  \cellcolor{best!35}.3204 &  \cellcolor{best!35}.4869 &  \cellcolor{second!35}19.40 &  \cellcolor{third!35}.5956 &  \cellcolor{third!35}.4198 &  \cellcolor{third!35}13.07 &  \cellcolor{best!35}.2189 &  \cellcolor{second!35}.6230 & & & \\
			& \ding{51} & \ding{51} & \ding{51} &  \cellcolor{best!35}20.07 &  \cellcolor{second!35}.4612 &  \cellcolor{best!35}.5290 &  \cellcolor{best!35}13.62 &  \cellcolor{best!35}.2678 &  \cellcolor{second!35}.5835 &  \cellcolor{second!35}14.38 &  \cellcolor{second!35}.3182 &  \cellcolor{second!35}.4913 & \cellcolor{best!35}19.51 & \cellcolor{best!35}.6048 & \cellcolor{second!35}.4146 & \cellcolor{second!35}13.09 &  \cellcolor{second!35}.2173 &  \cellcolor{third!35}.6256 &   &   &  \\
			\midrule
			& \multicolumn{3}{c}{Real Scene}& \multicolumn{3}{c}{Girl} & \multicolumn{3}{c}{Heron} & \multicolumn{3}{c}{Parterre} & \multicolumn{3}{c}{Puppet} & \multicolumn{3}{c}{Stair} & \multicolumn{3}{c}{Average} \\
			& \textit{SS} & \textit{MGS} & \textit{PD} & PSNR($\uparrow$) & SSIM($\uparrow$) & LPIPS($\downarrow$) & PSNR($\uparrow$) & SSIM($\uparrow$) & LPIPS($\downarrow$) & PSNR($\uparrow$) & SSIM($\uparrow$) & LPIPS($\downarrow$) & PSNR($\uparrow$) & SSIM($\uparrow$) & LPIPS($\downarrow$) & PSNR($\uparrow$) & SSIM($\uparrow$) & LPIPS($\downarrow$) & PSNR($\uparrow$) & SSIM($\uparrow$) & LPIPS($\downarrow$) \\
			\midrule
			\multirow{7}{*}{\textcolor[rgb]{0.13, 0.55, 0.13}{DP-kernel}}& & & &  11.14 &  .2151 &  .6485 &  13.63 &  .1473 &  .5376 &  14.70 &  .2226 &  .6348 &  14.32 &  .2887 &  .5930 &  14.38 &  .0521 &  \cellcolor{third!35}.6104 &  14.36 &  .2506 &  .5904\\
			& \ding{51} & & & 10.75 & .1816 & .6600 & 14.05 & .1553 & .5467 & 15.17 & .2340 & .6205 & 14.13 & .2770 & .5934 & 14.24 & .0442 & .6120 & 14.26 & .2414 & .5933 \\
			& & \ding{51} & & \cellcolor{second!35}12.89 & \cellcolor{second!35}.3465 & \cellcolor{second!35}.5645 & \cellcolor{best!35}14.48 & \cellcolor{second!35}.1677 & \cellcolor{best!35}.5209 & \cellcolor{second!35}16.20 & \cellcolor{third!35}.2511 & \cellcolor{second!35}.5927 & \cellcolor{third!35}14.48 & \cellcolor{third!35}.2941 & \cellcolor{second!35}.5819 & \cellcolor{third!35}15.81 & \cellcolor{third!35}.1167 & \cellcolor{best!35}.6039 & \cellcolor{second!35}15.46 & \cellcolor{second!35}.3035 & \cellcolor{best!35}.5465 \\
			& & & \ding{51} &  11.09 &  .2039 &  .6398 &  13.52 &  .1381 &  .5553 &  14.73 &  .2355 &  .6403 &  14.37 &  .2918 &  \cellcolor{third!35}.5857 &  14.49 &  .0555 &  .6165 &  14.28 &  .2490 &  .5896 \\
			& \ding{51} & & \ding{51} &  11.19 &  .2311 &  .6275 &  13.86 &  .1524 &  .5365 &  15.44 &  .2498 &  .6124 &  14.03 &  .2771 &  .6020 &  14.40 &  .0518 &  .6231 &  14.00 &  .2587 &  .5830 \\
			& \ding{51} & \ding{51} & &  \cellcolor{third!35}12.78 &  \cellcolor{third!35}.3357 &  \cellcolor{third!35}.5685 &  \cellcolor{second!35}14.45 &  \cellcolor{best!35}.1741 &  \cellcolor{second!35}.5228 &  \cellcolor{best!35}16.36 &  \cellcolor{best!35}.2643 &  \cellcolor{third!35}.5939 &  \cellcolor{second!35}14.56 &  \cellcolor{second!35}.2958 &  .5918 &  \cellcolor{second!35}16.11 &  \cellcolor{second!35}.1267 &  \cellcolor{second!35}.6054 &  \cellcolor{third!35}15.40 &  \cellcolor{third!35}.3009 &  \cellcolor{third!35}.5527 \\
			& \ding{51} & \ding{51} & \ding{51} & \cellcolor{best!35}13.11 &  \cellcolor{best!35}.3707 & \cellcolor{best!35}.5486 & \cellcolor{third!35}14.17 &  \cellcolor{third!35}.1669 &  \cellcolor{third!35}.5285 &  \cellcolor{third!35}16.19 &  \cellcolor{second!35}.2527 & \cellcolor{best!35}.5908 & \cellcolor{best!35}15.25 & \cellcolor{best!35}.3277 & \cellcolor{best!35}.5739 & \cellcolor{best!35}16.30 & \cellcolor{best!35}.1343 & .6133 &  \cellcolor{best!35}15.57 &  \cellcolor{best!35}.3114 &  \cellcolor{second!35}.5467 \\
			\midrule
			\multirow{7}{*}{\textcolor[rgb]{0.25, 0.0, 1.0}{DN-kernel}}& & & &  10.94 &  .2031 &  .6436 &  13.97 &  .1459 &  .5499 &  15.02 &  .2320 &  .6431 &  14.28 &  .2764 &  .6036 &  14.40 &  .0525 &  .6313 &  14.41 &  .2506 &  .5921 \\
			& \ding{51} & & & 10.43 & .1696 & .6654 & 13.10 & .1336 & .5652 & 15.09 & .2075 & .6536 & 14.66 & .3000 & .5950 & 14.36 & .0528 & .6122 & 14.33 & .2495 & .5937 \\
			& & \ding{51} & & \cellcolor{third!35}12.55 & \cellcolor{third!35}.3184 & \cellcolor{third!35}.5736 & \cellcolor{second!35}14.51 & \cellcolor{best!35}.1850 & \cellcolor{third!35}.5217 & \cellcolor{best!35}16.65 & \cellcolor{third!35}.2830 & \cellcolor{third!35}.5953 & 14.68 & .2944 & \cellcolor{third!35}.5924 & \cellcolor{third!35}15.93 & \cellcolor{third!35}.1189 & \cellcolor{second!35}.6081 & \cellcolor{third!35}15.44 & \cellcolor{third!35}.3037 & \cellcolor{third!35}.5558 \\
			& & & \ding{51} &  11.19 &  .2214 &  .6475 &  13.95 &  .1421 &  .5508 &  14.30 &  .2189 &  .6739 &  \cellcolor{best!35}14.99 &  \cellcolor{second!35}.3117 &  \cellcolor{second!35}.5894 &  14.35 &  .0574 &  .6164 &  14.47 &  .2549 &  .5939 \\
			& \ding{51} & & \ding{51} &  11.55 &  .2299 &  .6195 &  13.93 &  .1543 &  .5382 &  14.92 &  .2291 &  .6399 &  \cellcolor{third!35}14.74 &  \cellcolor{third!35}.3042 &  .6001 &  14.43 &  .0556 &  .6211 &  14.47 &  .2563 &  .5881 \\
			& \ding{51} & \ding{51} & &  \cellcolor{second!35}12.70 &  \cellcolor{best!35}.3438 &  \cellcolor{second!35}.5627 &  \cellcolor{best!35}14.57 &  \cellcolor{second!35}.1779 &  \cellcolor{second!35}.5200 &  \cellcolor{third!35}16.51 &  \cellcolor{second!35}.2872 &  \cellcolor{best!35}.5825 &  14.24 &  .2848 &  .6097 &  \cellcolor{best!35}16.02 &  \cellcolor{second!35}.1216 &  \cellcolor{best!35}.6018 &  \cellcolor{second!35}15.46 &  \cellcolor{second!35}.3073 &  \cellcolor{second!35}.5519 \\
			& \ding{51} & \ding{51} & \ding{51} &  \cellcolor{best!35}12.85 & \cellcolor{second!35}.3418 &  \cellcolor{best!35}.5585 &  \cellcolor{third!35}14.30 &  \cellcolor{third!35}.1706 &  \cellcolor{best!35}.5197 & \cellcolor{second!35}16.61 &  \cellcolor{best!35}.2903 & \cellcolor{second!35}.5912 &  \cellcolor{second!35}14.91 &  \cellcolor{best!35}.3141 &  \cellcolor{best!35}.5886 &  \cellcolor{second!35}15.97 &  \cellcolor{best!35}.1255 &  \cellcolor{third!35}.6125 &  \cellcolor{best!35}15.53 &  \cellcolor{best!35}.3112 &  \cellcolor{best!35}.5515 \\
			\midrule
      \end{tabular}
      }
   \label{tab:appendix_ablation_quant_results_2view}
   \end{center}
\end{table*}

\begin{table*}[t]
   \Large
   \begin{center}
   \caption{Ablation quantitative results for the entire scenes of synthetic and real scenes obtained from \textbf{4-view} settings. Each color shading represents the \colorbox{best!35}{best}, \colorbox{second!35}{second best} and \colorbox{third!35}{third best} result, respectively.}
      \resizebox{2\columnwidth}{!}{
		\centering
		\setlength{\tabcolsep}{1pt}
         \begin{tabular}{c|C{1.1cm}C{1.1cm}C{1.1cm}||ccc|ccc|ccc|ccc|ccc||ccc}
			\midrule		
			\multirow{2}{*}{Kernel Type}& \multicolumn{3}{c}{Synthetic Scene} & \multicolumn{3}{c}{Factory} & \multicolumn{3}{c}{Cozyroom} & \multicolumn{3}{c}{Pool} & \multicolumn{3}{c}{Tanabata} & \multicolumn{3}{c}{Trolley} & \multicolumn{3}{c}{Average} \\
			 & \textit{SS} & \textit{MGS} & \textit{PD} & PSNR($\uparrow$) & SSIM($\uparrow$) & LPIPS($\downarrow$) & PSNR($\uparrow$) & SSIM($\uparrow$) & LPIPS($\downarrow$) & PSNR($\uparrow$) & SSIM($\uparrow$) & LPIPS($\downarrow$) & PSNR($\uparrow$) & SSIM($\uparrow$) & LPIPS($\downarrow$) & PSNR($\uparrow$) & SSIM($\uparrow$) & LPIPS($\downarrow$) & PSNR($\uparrow$) & SSIM($\uparrow$) & LPIPS($\downarrow$) \\
			\midrule
			\multirow{7}{*}{\textcolor[rgb]{0.13, 0.55, 0.13}{DP-kernel}}& & & &  \cellcolor{best!35}19.20 &  \cellcolor{second!35}.5175 &  \cellcolor{best!35}.3813 &  21.50 &  .6242 &  .1980 &  21.85 &  .4235 &  .3117 &  \cellcolor{third!35}17.30 &  \cellcolor{third!35}.5265 &  \cellcolor{third!35}.3381 &  \cellcolor{best!35}19.35 &  \cellcolor{second!35}.5765 &  \cellcolor{best!35}.3084 &  19.84 &  .5336 &  .3075 \\
			& \ding{51} & & & \cellcolor{third!35}19.01 & .4914 & .4082 & 21.69 & .6356 & .1889 & 22.62 & .4520 & .3013 & 16.61 & .4789 & .3771 & 17.82 & .5385 & .3475 & 19.55 & .5193 & .3246 \\
			& & \ding{51} & & 18.22 & .4638 & .4144 & 22.48 & .6603 & .1762 & \cellcolor{third!35}23.11 & \cellcolor{third!35}.4709 & \cellcolor{third!35}.2791 & \cellcolor{best!35}18.18 & \cellcolor{best!35}.5517 & \cellcolor{best!35}.3124 & 18.83 & .5657 & .3299 & \cellcolor{third!35}20.16 & \cellcolor{third!35}.5425 & \cellcolor{third!35}.3024 \\
			& & & \ding{51} &  18.79 &  \cellcolor{third!35}.4972 &  \cellcolor{third!35}.3974 &  22.59 &  .6760 &  .2314 &  22.41 &  .4348 &  .3011 &  16.88 &  .4913 &  .3606 &  17.59 &  .5194 &  .3652 &  19.65 &  .5237&  .3311 \\
			& \ding{51} & & \ding{51} &  \cellcolor{second!35}19.09 &  \cellcolor{best!35}.5253 &  \cellcolor{second!35}.3889 &  \cellcolor{second!35}25.83 &  \cellcolor{second!35}.7796 &  \cellcolor{third!35}.1709 &  \cellcolor{best!35}24.13 &  \cellcolor{best!35}.5229 &  \cellcolor{best!35}.2474 &  16.95 &  .5017 &  .3501 &  18.56 &  \cellcolor{third!35}.5764 &  \cellcolor{third!35}.3167 &  \cellcolor{second!35}20.91 &  \cellcolor{best!35}.5812 &  \cellcolor{best!35}.2948 \\
			& \ding{51} & \ding{51} & &  17.62 &  .4227 &  .4615 &  \cellcolor{third!35}23.43 &  \cellcolor{third!35}.7266 &  \cellcolor{second!35}.1688 &  22.71 &  .4454 &  .2850 &  15.64 &  .4150 &  .4308 &  \cellcolor{third!35}18.84 &  .5722 &  .3209 &  19.65 &  .5164 &  .3334 \\
			& \ding{51} & \ding{51} & \ding{51} &  18.99 &  .4770 &  .4034 &  \cellcolor{best!35}26.51 &  \cellcolor{best!35}.8051 &  \cellcolor{best!35}.1627 &  \cellcolor{second!35}23.33 &  \cellcolor{second!35}.4888 &  \cellcolor{second!35}.2733 & \cellcolor{second!35}17.51 &  \cellcolor{second!35}.5371 &  \cellcolor{second!35}.3356 &  \cellcolor{second!35}18.92 &  \cellcolor{best!35}.5799 & \cellcolor{second!35}.3126 &  \cellcolor{best!35}21.05 &  \cellcolor{second!35}.5776 &  \cellcolor{second!35}.2975\\
			\midrule
			\multirow{7}{*}{\textcolor[rgb]{0.25, 0.0, 1.0}{DN-kernel}}& & & &  \cellcolor{best!35}17.26 &  \cellcolor{second!35}.3740 &  \cellcolor{best!35}.5088 &  25.51 &  .7767 &  .1897 & \cellcolor{third!35}23.38 &  \cellcolor{third!35}.5068 &  \cellcolor{second!35}.2649 &  15.91 &  .4209 &  .4198 &  17.91 &  .5213 &  .3661 &  \cellcolor{third!35}19.99 &  .5199 &  \cellcolor{third!35}.3499 \\
			& \ding{51} & & & 16.89 & .3611 & \cellcolor{third!35}.5353 & \cellcolor{best!35}25.82 & \cellcolor{best!35}.7876 & \cellcolor{best!35}.1770 & 22.92 & .4820 & .2809 & 15.52 & .4381 & .4173 & 18.07 & .5317 & .3684 & 19.89 & .5201 & .3558 \\
			& & \ding{51} & & 16.33 & .3387 & .5636 & \cellcolor{third!35}25.55 & \cellcolor{third!35}.7813 & \cellcolor{third!35}.1874 & \cellcolor{second!35}24.28 & \cellcolor{second!35}.5544 & \cellcolor{second!35}.2649 & \cellcolor{third!35}16.60 & .4635 & \cellcolor{second!35}.3781 & 17.96 & .5095 & .3870 & 19.44 & \cellcolor{second!35}.5295 & .3562 \\
			& & & \ding{51} &  16.49 &  .3444 &  .5691 &  25.40 &  .7674 &  .1982 &  21.87 &  .4194 &  .3203 &  16.48 &  \cellcolor{second!35}.4798 &  .3830 &  \cellcolor{second!35}18.38 &  \cellcolor{third!35}.5412 &  \cellcolor{second!35}.3445 &  19.72 &  .5104 &  .3630 \\
			& \ding{51} & & \ding{51} &  \cellcolor{second!35}17.11 &  \cellcolor{best!35}.3766 &  \cellcolor{second!35}.5195 &  25.43 &  .7770 &  .1891 &  22.41 &  .4585 &  .2864 &  \cellcolor{second!35}16.63 &  \cellcolor{third!35}.4792 &  .3826 &  \cellcolor{best!35}18.57 &  \cellcolor{best!35}.5559 &  \cellcolor{third!35}.3476 &  \cellcolor{second!35}20.03 &  \cellcolor{third!35}.5294 &  \cellcolor{second!35}.3450 \\
			& \ding{51} & \ding{51} & &  16.22 &  .3389 &  .5637 &  25.12 &  .7566 &  .1950 & 23.00 &  .4792 &  .2810 &  \cellcolor{best!35}17.37 &  \cellcolor{best!35}.5088 &  \cellcolor{best!35}.3661 &  18.08 &  .5275 &  .3677 &  19.96 & .5222 &  .3547 \\
			& \ding{51} & \ding{51} & \ding{51} &  \cellcolor{third!35}16.91 &  \cellcolor{third!35}.3693 &  .5357 &  \cellcolor{second!35}25.79 &  \cellcolor{second!35}.7872 &  \cellcolor{second!35}.1801 &  \cellcolor{best!35}25.48 &  \cellcolor{best!35}.6042 &  \cellcolor{best!35}.2405 &  16.58 &  .4705 &  \cellcolor{third!35}.3793 &  \cellcolor{third!35}18.10 &  \cellcolor{second!35}.5514 &  \cellcolor{best!35}.3414 &  \cellcolor{best!35}20.57 &  \cellcolor{best!35}.5565 &  \cellcolor{best!35}.3354 \\
			\midrule\midrule
			
			& \multicolumn{3}{c}{Real Scene}& \multicolumn{3}{c}{Ball} & \multicolumn{3}{c}{Basket} & \multicolumn{3}{c}{Buick} & \multicolumn{3}{c}{Coffee} & \multicolumn{3}{c}{Decoration} & & \\
			& \textit{SS} & \textit{MGS} & \textit{PD} & PSNR($\uparrow$) & SSIM($\uparrow$) & LPIPS($\downarrow$) & PSNR($\uparrow$) & SSIM($\uparrow$) & LPIPS($\downarrow$) & PSNR($\uparrow$) & SSIM($\uparrow$) & LPIPS($\downarrow$) & PSNR($\uparrow$) & SSIM($\uparrow$) & LPIPS($\downarrow$) & PSNR($\uparrow$) & SSIM($\uparrow$) & LPIPS($\downarrow$) &  & &\\
			\midrule
			\multirow{7}{*}{\textcolor[rgb]{0.13, 0.55, 0.13}{DP-kernel}}& & & &  23.20 &  .5842 &  .3847 &  17.42 &  .4046 &  .4658 &  19.26 &  .5323 &  .3374 &  24.05 &  .7597 &  .2624 & 14.98 &  .3018 &  .5618 &   &   &   \\
			& \ding{51} & & & \cellcolor{best!35}23.55 & \cellcolor{best!35}.6053 & \cellcolor{second!35}.3685 & 18.32 & .4442 & .4292 & \cellcolor{third!35}19.35 & .5340 & .3339 & 23.17 & .7442 & .2670 & 16.14 & .3739 & .5082 & & & \\
			& & \ding{51} & & 22.52 & .5522 & \cellcolor{third!35}.3689 & \cellcolor{second!35}20.31 & \cellcolor{second!35}.5388 & \cellcolor{second!35}.3364 & 19.16 & .5203 & .3430 & \cellcolor{third!35}25.62 & .7884 & \cellcolor{second!35}.2082 & \cellcolor{best!35}17.33 & \cellcolor{best!35}.4081 & \cellcolor{best!35}.4777 & & & \\
			& & & \ding{51} &  23.18 &  .5817 &  .3878 &  17.63 &  .4194 &  .4475 &  \cellcolor{best!35}19.53 &  \cellcolor{third!35}.5441 &  \cellcolor{third!35}.3325 &  24.85 &  \cellcolor{third!35}.7885 &  .2467 &  14.99 &  .3407 &  .5267 & & &  \\
			& \ding{51} & & \ding{51} &  \cellcolor{third!35}23.30 &  \cellcolor{third!35}.5893 &  .3904 &  18.49 &  .4615 &  .3926 &  19.14 &  .5385 &  .3408 &  24.05 &  .7716 &  .2494 &  15.25 &  .3458 &  .5274 & & & \\
			& \ding{51} & \ding{51} & &  22.36 &  .5504 &  \cellcolor{best!35}.3612 &  \cellcolor{third!35}19.61 &  \cellcolor{third!35}.5134 &  \cellcolor{third!35}.3661 &  19.27 &  \cellcolor{second!35}.5458 &  \cellcolor{second!35}.3316 &  \cellcolor{second!35}27.55 &  \cellcolor{second!35}.8167 &  \cellcolor{third!35}.2242 &  \cellcolor{second!35}16.53 &  \cellcolor{third!35}.3902 &  \cellcolor{second!35}.4867 & & & \\
			& \ding{51} & \ding{51} & \ding{51} &  \cellcolor{second!35}23.39 & \cellcolor{second!35}.6010 &  .3806 &  \cellcolor{best!35}20.41 &  \cellcolor{best!35}.5451 &  \cellcolor{best!35}.3305 &  \cellcolor{second!35}19.48 &  \cellcolor{best!35}.5531 &  \cellcolor{best!35}.3251 &  \cellcolor{best!35}27.77 &  \cellcolor{best!35}.8364 &  \cellcolor{best!35}.2049 & \cellcolor{third!35}16.33 &  \cellcolor{second!35}.3914 &  \cellcolor{third!35}.4950 &   &   &   \\
			\midrule
			\multirow{7}{*}{\textcolor[rgb]{0.25, 0.0, 1.0}{DN-kernel}} & & & &  21.90 &  .5182 &  .4609 &  17.15 &  .4331 &  .4316 &  \cellcolor{third!35}19.34 &  .5278 &  .3407 &  23.10 &  .7302 &  .2873 &  14.94 &  .3201 &  .5471 &   &  &   \\
			& \ding{51} & & & 21.94 & .5220 & .4571 & 17.38 & .4464 & .4298 & 19.26 & .5401 & .3418 & 23.03 & .7264 & .2830 & 14.82 & .3293 & .5363 & & &\\
			& & \ding{51} & & 21.85 & .5168 & .4514 & \cellcolor{second!35}17.87 & \cellcolor{third!35}.4641 & \cellcolor{third!35}.4014 & 19.23 & .5289 & .3467 & \cellcolor{second!35}26.91 & \cellcolor{third!35}.8082 & \cellcolor{third!35}.2473 & \cellcolor{third!35}16.69 & \cellcolor{second!35}.4031 & \cellcolor{second!35}.4859 & & & \\
			& & & \ding{51} &  21.85 &  \cellcolor{third!35}.5313 &  \cellcolor{second!35}.4392 &  16.97 &  .4301 &  .4228 &  \cellcolor{best!35}19.45 &  \cellcolor{best!35}.5443 &  .3403 &  23.78 &  .7679 &  .2754 &  15.18 &  .3549 &  .5112 & & &  \\
			& \ding{51} & & \ding{51} &  \cellcolor{third!35}22.06 &  \cellcolor{second!35}.5319 &  .4481 &  16.75 &  .4308 &  .4333 &  \cellcolor{second!35}19.34 &  \cellcolor{second!35}.5421 &  \cellcolor{second!35}.3373 &  23.44 &  .7421 &  .2943 &  14.86 &  .3429 &  .5299 & & & \\
			& \ding{51} & \ding{51} & &  \cellcolor{second!35}22.09 &  .5299 &  \cellcolor{third!35}.4477 &  \cellcolor{third!35}17.81 &  \cellcolor{second!35}.4854 &  \cellcolor{second!35}.3812 &  19.31 &  .5400 &  \cellcolor{third!35}.3384 &  \cellcolor{best!35}27.40 &  \cellcolor{best!35}.8263 &  \cellcolor{best!35}.2353 &  \cellcolor{second!35}16.93 &  \cellcolor{third!35}.4023 &  \cellcolor{third!35}.4879 & & & \\
			& \ding{51} & \ding{51} & \ding{51} &  \cellcolor{best!35}22.28 &  \cellcolor{best!35}.5506 &  \cellcolor{best!35}.4334 & \cellcolor{best!35}18.69 &  \cellcolor{best!35}.5058 &  \cellcolor{best!35}.3774 &  19.28 & \cellcolor{third!35}.5414 &  \cellcolor{best!35}.3371 & \cellcolor{third!35}26.76 &  \cellcolor{second!35}.8126 &  \cellcolor{second!35}.2419 &  \cellcolor{best!35}17.12 &  \cellcolor{best!35}.4192 &  \cellcolor{best!35}.4761 &   &   &   \\
			\midrule
			& \multicolumn{3}{c}{Real Scene}& \multicolumn{3}{c}{Girl} & \multicolumn{3}{c}{Heron} & \multicolumn{3}{c}{Parterre} & \multicolumn{3}{c}{Puppet} & \multicolumn{3}{c}{Stair} & \multicolumn{3}{c}{Average} \\
			& \textit{SS} & \textit{MGS} & \textit{PD} & PSNR($\uparrow$) & SSIM($\uparrow$) & LPIPS($\downarrow$) & PSNR($\uparrow$) & SSIM($\uparrow$) & LPIPS($\downarrow$) & PSNR($\uparrow$) & SSIM($\uparrow$) & LPIPS($\downarrow$) & PSNR($\uparrow$) & SSIM($\uparrow$) & LPIPS($\downarrow$) & PSNR($\uparrow$) & SSIM($\uparrow$) & LPIPS($\downarrow$) & PSNR($\uparrow$) & SSIM($\uparrow$) & LPIPS($\downarrow$) \\
			\midrule
			\multirow{7}{*}{\textcolor[rgb]{0.13, 0.55, 0.13}{DP-kernel}}& & & &  15.40 &  .5237 &  .4318 &  18.68 &  .4292 &  .3352 &  17.62 &  .2987 &  .4991 &  17.61 &  .4369 &  .4313 &  19.44 &  .3107 &  .4657 &  18.77 &  .4582 &  .4175 \\
			& \ding{51} & & & 15.67 & .5342 & .4274 & 18.44 & .4145 & .3322 & 17.37 & .2900 & .4935 & \cellcolor{best!35}18.49 & \cellcolor{best!35}.4765 & \cellcolor{best!35}.3982 & 19.85 & .3433 & .4503 & 19.04 & .4760 & .4008 \\
			& & \ding{51} & & \cellcolor{best!35}17.03 & \cellcolor{best!35}.6045 & \cellcolor{best!35}.3636 & 18.90 & .4526 & \cellcolor{second!35}.3185 & \cellcolor{best!35}18.38 & \cellcolor{third!35}.3323 & \cellcolor{best!35}.4575 & 17.89 & .4408 & .4333 & \cellcolor{third!35}20.33 & .3416 & \cellcolor{third!35}.4297 & \cellcolor{second!35}19.75 & \cellcolor{third!35}.4980 & \cellcolor{second!35}.3737 \\
			& & & \ding{51} &  14.80 &  .4950 &  .4538 &  18.83 &  .4336 &  .3308 &  17.81 &  .3089 &  .4915 &  \cellcolor{second!35}18.12 &  \cellcolor{third!35}.4505 &  \cellcolor{second!35}.4205 &  20.23 &  \cellcolor{second!35}.3615 &  .4422 &  19.00 &  .4724 &  .4080 \\
			& \ding{51} & & \ding{51} &  15.57 &  .5297 &  .4349 &  \cellcolor{second!35}19.04 &  \cellcolor{second!35}.4578 &  .3278 &  17.88 &  .3094 &  .4887 &  17.84 &  .4461 &  .4329 &  19.47 &  .3102 &  .4522 &  19.00 &  .4760 &  .4037 \\
			& \ding{51} & \ding{51} & &  \cellcolor{second!35}16.78 &  \cellcolor{second!35}.5919 &  \cellcolor{second!35}.3806 &  \cellcolor{third!35}18.99 &  \cellcolor{third!35}.4575 &  \cellcolor{best!35}.3166 &  \cellcolor{third!35}18.28 &  \cellcolor{second!35}.3333 &  \cellcolor{second!35}.4630 &  17.63 &  .4272 &  .4243 &  \cellcolor{second!35}20.36 &  \cellcolor{third!35}.3594 &  \cellcolor{second!35}.4201 &  \cellcolor{third!35}19.74 &  \cellcolor{second!35}.4986 &  \cellcolor{third!35}.3774 \\
			& \ding{51} & \ding{51} & \ding{51} &  \cellcolor{third!35}16.47 & \cellcolor{third!35}.5855 &  \cellcolor{third!35}.3835 & \cellcolor{best!35}19.09 &  \cellcolor{best!35}.4624 &  \cellcolor{third!35}.3194 &  \cellcolor{second!35}18.36 &  \cellcolor{best!35}.3375 &  \cellcolor{third!35}.4715 & \cellcolor{third!35}17.97 &  \cellcolor{second!35}.4612 & \cellcolor{third!35}.4206 &  \cellcolor{best!35}21.25 & \cellcolor{best!35}.4045 & \cellcolor{best!35}.4044 &  \cellcolor{best!35}20.05 &  \cellcolor{best!35}.5178 &  \cellcolor{best!35}.3736 \\
			\midrule
			\multirow{7}{*}{\textcolor[rgb]{0.25, 0.0, 1.0}{DN-kernel}}& & & &  15.62 &  .5351 &  .4206 &  18.85 &  .4350 &  .3398 &  19.71 &  .4318 &  .4549 &  17.77 &  .4400 &  .4358 &  20.49 &  .3901 & \cellcolor{third!35}.4326 &  18.89 &  .4761 & .4151 \\
			& \ding{51} & & & 15.04 & .5147 & .4312 & 18.75 & .4291 & .3585 & \cellcolor{second!35}20.73 & \cellcolor{third!35}.4841 & \cellcolor{second!35}.4379 & \cellcolor{third!35}18.15 & \cellcolor{second!35}.4647 & \cellcolor{third!35}.4248 & 20.45 & .3825 & \cellcolor{second!35}.4249 & 18.96 & .4839 & .4125 \\
			& & \ding{51} & & \cellcolor{best!35}17.13 & \cellcolor{best!35}.6103 & \cellcolor{best!35}.3623 & 18.96 & \cellcolor{second!35}.4527 & \cellcolor{third!35}.3366 & 20.47 & .4735 & .4464 & \cellcolor{best!35}18.39 & \cellcolor{best!35}.4694 & \cellcolor{best!35}.4156 & \cellcolor{second!35}20.94 & \cellcolor{third!35}.3914 & .4432 & \cellcolor{third!35}19.84 & \cellcolor{third!35}.5118 & \cellcolor{third!35}.3937 \\
			& & & \ding{51} &  15.17 &  .5215 &  .4383 &  \cellcolor{third!35}18.97 &  .4367 &  .3480 &  20.57 &  .4833 &  .4463 &  \cellcolor{second!35}18.24 &  \cellcolor{third!35}.4620 &  .4252 &  20.42 &  .3752 &  .4578 &  19.06 &  .4907 &  .4105 \\
			& \ding{51} & & \ding{51} &  15.48 &  .5272 &  .4282 &  \cellcolor{second!35}19.00 &  .4430 &  .3458 &  20.37 &  .4801 &  .4401 &  18.08 &  .4577 &  .4276 &  20.05 &  .3635 &  .4601 &  18.94 &  .4861 &  .4145 \\
			& \ding{51} & \ding{51} & &  \cellcolor{second!35}17.07 &  \cellcolor{second!35}.6081 &  \cellcolor{second!35}.3706 &  18.95 &  \cellcolor{third!35}.4514 &  \cellcolor{best!35}.3254 &  \cellcolor{third!35}20.64 &  \cellcolor{second!35}.4854 &  \cellcolor{third!35}.4385 &  18.03 &  .4510 &  .4298 &  \cellcolor{best!35}21.05 &  \cellcolor{best!35}.4086 &  \cellcolor{best!35}.4182 &  \cellcolor{second!35}19.93 &  \cellcolor{second!35}.5188 &  \cellcolor{second!35}.3873 \\
			& \ding{51} & \ding{51} & \ding{51} & \cellcolor{third!35}16.95 &  \cellcolor{third!35}.5975 & \cellcolor{third!35}.3823 &  \cellcolor{best!35}19.04 &  \cellcolor{best!35}.4558 &  \cellcolor{second!35}.3315 &  \cellcolor{best!35}20.74 &  \cellcolor{best!35}.4916 &  \cellcolor{best!35}.4323 & 18.11 &  .4616 &  \cellcolor{second!35}.4202 &  \cellcolor{third!35}20.80 & \cellcolor{second!35}.3948 &  .4386 &  \cellcolor{best!35}19.98 &  \cellcolor{best!35}.5231 & \cellcolor{best!35}.3871 \\
			\midrule
      \end{tabular}
      }
   \label{tab:appendix_ablation_quant_results_4view}
   \end{center}
\end{table*}

\begin{table*}[t]
   \Large
   \begin{center}
   \caption{Ablation quantitative results for the entire scenes of synthetic and real scenes obtained from \textbf{6-view} settings. Each color shading represents the \colorbox{best!35}{best}, \colorbox{second!35}{second best} and \colorbox{third!35}{third best} result, respectively.}
      \resizebox{2\columnwidth}{!}{
		\centering
		\setlength{\tabcolsep}{1pt}
         \begin{tabular}{c|C{1.1cm}C{1.1cm}C{1.1cm}||ccc|ccc|ccc|ccc|ccc||ccc}
			\midrule		
			\multirow{2}{*}{Kernel Type}& \multicolumn{3}{c}{Synthetic Scene} & \multicolumn{3}{c}{Factory} & \multicolumn{3}{c}{Cozyroom} & \multicolumn{3}{c}{Pool} & \multicolumn{3}{c}{Tanabata} & \multicolumn{3}{c}{Trolley} & \multicolumn{3}{c}{Average} \\
			 & \textit{SS} & \textit{MGS} & \textit{PD} & PSNR($\uparrow$) & SSIM($\uparrow$) & LPIPS($\downarrow$) & PSNR($\uparrow$) & SSIM($\uparrow$) & LPIPS($\downarrow$) & PSNR($\uparrow$) & SSIM($\uparrow$) & LPIPS($\downarrow$) & PSNR($\uparrow$) & SSIM($\uparrow$) & LPIPS($\downarrow$) & PSNR($\uparrow$) & SSIM($\uparrow$) & LPIPS($\downarrow$) & PSNR($\uparrow$) & SSIM($\uparrow$) & LPIPS($\downarrow$) \\
			\midrule
			\multirow{7}{*}{\textcolor[rgb]{0.13, 0.55, 0.13}{DP-kernel}}& & & & \cellcolor{second!35}21.63 & \cellcolor{second!35}.6402 & \cellcolor{second!35}.2984 & \cellcolor{best!35}27.63 & \cellcolor{best!35}.8475 & \cellcolor{best!35}.1224 & 25.36 & .6227 & .1861 & 21.27 & .6818 & .2023 & \cellcolor{best!35}22.49 & \cellcolor{best!35}.7259 & \cellcolor{best!35}.1899 & 23.68 & .7036 & \cellcolor{second!35}.1998 \\
			& \ding{51} & & & \cellcolor{best!35}21.84 & \cellcolor{best!35}.6556 & \cellcolor{best!35}.2947 & \cellcolor{third!35}27.48 & \cellcolor{second!35}.8419 & \cellcolor{third!35}.1272 & 26.73 & .6891 & .1687 & \cellcolor{best!35}22.57 & \cellcolor{third!35}.7289 & \cellcolor{best!35}.1950 & 22.15 & \cellcolor{third!35}.7160 & .2015 & \cellcolor{third!35}24.15 & \cellcolor{second!35}.7263 & \cellcolor{best!35}.1974 \\
			& & \ding{51} & & 19.30 & .5628 & .3553 & \cellcolor{second!35}27.56 & \cellcolor{third!35}.8408 & \cellcolor{second!35}.1262 & \cellcolor{third!35}28.10 & \cellcolor{second!35}.7374 & \cellcolor{best!35}.1544 & 21.83 & .7128 & .2074 & \cellcolor{second!35}22.34 & \cellcolor{second!35}.7197 & \cellcolor{second!35}.1928 & 23.83 & .7147 & .2072 \\
			& & & \ding{51} & 20.52 & .5994 & .3224 & 27.29 & .8322 & .1349 & 27.07 & .7002 & .1719 & \cellcolor{second!35}22.50 & \cellcolor{best!35}.7306 & \cellcolor{second!35}.1958 & 22.07 & .7127 & \cellcolor{third!35}.2011 & 23.89 & .7150 & .2052 \\
			& \ding{51} & & \ding{51} & 21.27 & .6104 & \cellcolor{third!35}.3191 & 27.39 & .8376 & .1301 & \cellcolor{best!35}28.54 & \cellcolor{best!35}.7518 & .1608 & \cellcolor{third!35}22.32 & \cellcolor{second!35}.7290 & .1966 & 21.87 & .7109 & .2037 & \cellcolor{best!35}24.28 & \cellcolor{best!35}.7279 & \cellcolor{third!35}.2021 \\
			& \ding{51} & \ding{51} & &  18.43 & .5407 & .3738 & 27.44 & .8399 & .1313 & 27.89 & .7321 & \cellcolor{second!35}.1596 & 22.02 & .7197 & .2071 & 21.87 & .7118 & .2046 & 23.53 & .7088 & .2153 \\
			& \ding{51} & \ding{51} & \ding{51} & \cellcolor{third!35}21.29 & \cellcolor{third!35}.6179 & .3261 & 27.34 & .8340 & .1298 & \cellcolor{second!35}28.19 & \cellcolor{third!35}.7365 & \cellcolor{third!35}.1606 & 22.30 & .7233 & \cellcolor{third!35}.2017 & \cellcolor{third!35}22.21 & .7159 & .2036 & \cellcolor{second!35}24.27 & \cellcolor{third!35}.7255 & .2044 \\
			\midrule
			\multirow{7}{*}{\textcolor[rgb]{0.25, 0.0, 1.0}{DN-kernel}}& & & & \cellcolor{third!35}18.77 & .5048 & \cellcolor{third!35}.3890 & \cellcolor{third!35}26.67 & \cellcolor{third!35}.8214 & \cellcolor{second!35}.1475 & 27.50 & .7116 & .1783 & 21.10 & .6693 & .2517 & 21.58 & \cellcolor{best!35}.6921 & \cellcolor{best!35}.2263 & 23.12 & .6798 & \cellcolor{third!35}.2386 \\
			& \ding{51} & & & 17.87 & .4554 & .4370 & \cellcolor{best!35}26.76 & \cellcolor{best!35}.8239 & \cellcolor{best!35}.1462 & \cellcolor{third!35}28.12 & .7316 & \cellcolor{third!35}.1735 & \cellcolor{third!35}21.69 & \cellcolor{second!35}.6979 & \cellcolor{second!35}.2282 & \cellcolor{third!35}21.61 & .6887 & \cellcolor{second!35}.2294 & \cellcolor{second!35}23.59 & .6795 & .2429 \\
			& & \ding{51} & & 18.06 & .4849 & .4104 & 26.54 & .8159 & .1486 & 28.09 & \cellcolor{best!35}.7392 & \cellcolor{best!35}.1725 & 20.97 & .6866 & .2394 & \cellcolor{best!35}21.71 & \cellcolor{third!35}.6895 & .2320 & \cellcolor{best!35}24.33 & .6832 & .2406 \\
			& & & \ding{51} & \cellcolor{second!35}19.22 & \cellcolor{second!35}.5397 & \cellcolor{second!35}.3756 & 26.52 & .8184 & .1514 & 27.72 & .7242 & .1828 & 21.67 & .6961 & \cellcolor{third!35}.2306 & \cellcolor{second!35}21.64 & \cellcolor{second!35}.6920 & \cellcolor{third!35}.2300 & 23.35 & \cellcolor{best!35}.6941 & \cellcolor{second!35}.2341 \\
			& \ding{51} & & \ding{51} & \cellcolor{best!35}19.77 & \cellcolor{best!35}.5461 & \cellcolor{best!35}.3692 & 26.63 & .8192 & .1480 & \cellcolor{best!35}28.17 & \cellcolor{second!35}.7353 & \cellcolor{second!35}.1733 & 21.08 & .6878 & .2367 & 21.39 & .6821 & .2347 & \cellcolor{third!35}23.41 & \cellcolor{best!35}.6941 & \cellcolor{best!35}.2324 \\
			& \ding{51} & \ding{51} & &  17.50 & .4349 & .4613 & \cellcolor{second!35}26.72 & \cellcolor{second!35}.8226 & .1491 & 27.40 & .7046 & .1848 & \cellcolor{second!35}21.73 & \cellcolor{third!35}.6962 & .2369 & 21.20 & .6688 & .2477 & 22.91 & .6654 & .2560 \\
			& \ding{51} & \ding{51} & \ding{51} & 18.33 & \cellcolor{third!35}.5144 & .4004 & 26.63 & .8184 & .1513 & \cellcolor{second!35}28.16 & \cellcolor{third!35}.7344 & .1755 & \cellcolor{best!35}22.06 & \cellcolor{best!35}.6988 & \cellcolor{best!35}.2245 & 21.43 & .6855 & .2380 & 23.32 & \cellcolor{third!35}.6903 & .2379 \\
			\midrule\midrule
			
			& \multicolumn{3}{c}{Real Scene}& \multicolumn{3}{c}{Ball} & \multicolumn{3}{c}{Basket} & \multicolumn{3}{c}{Buick} & \multicolumn{3}{c}{Coffee} & \multicolumn{3}{c}{Decoration} & & \\
			& \textit{SS} & \textit{MGS} & \textit{PD} & PSNR($\uparrow$) & SSIM($\uparrow$) & LPIPS($\downarrow$) & PSNR($\uparrow$) & SSIM($\uparrow$) & LPIPS($\downarrow$) & PSNR($\uparrow$) & SSIM($\uparrow$) & LPIPS($\downarrow$) & PSNR($\uparrow$) & SSIM($\uparrow$) & LPIPS($\downarrow$) & PSNR($\uparrow$) & SSIM($\uparrow$) & LPIPS($\downarrow$) &  & &\\
			\midrule
			\multirow{7}{*}{\textcolor[rgb]{0.13, 0.55, 0.13}{DP-kernel}}& & & & 24.73 & .6651 & .3107 & \cellcolor{second!35}23.24 & .6643 & .2494 & \cellcolor{best!35}21.65 & \cellcolor{best!35}.6418 & \cellcolor{best!35}.2445 & 25.52 & .7949 & .2170 & 17.48 & .4725 & .4457 \\
			& \ding{51} & & & \cellcolor{second!35}24.93 & \cellcolor{best!35}.6697 & \cellcolor{second!35}.3054 & 22.91 & .6688 & \cellcolor{best!35}.2453 & \cellcolor{second!35}21.48 & \cellcolor{second!35}.6400 & \cellcolor{second!35}.2580 & 28.12 & .8224 & .1987 & 18.34 & .4934 & .4226 \\
			& & \ding{51} & & 24.21 & .6412 & \cellcolor{third!35}.3094 & 22.82 & \cellcolor{third!35}.6631 & .2508 & 21.17 & .6251 & .2681 & \cellcolor{third!35}28.39 & \cellcolor{second!35}.8448 & \cellcolor{best!35}.1722 & \cellcolor{second!35}19.82 & \cellcolor{second!35}.5485 & \cellcolor{third!35}.3697 \\
			& & & \ding{51} & \cellcolor{third!35}24.92 & \cellcolor{third!35}.6661 & .3111 & 22.27 & .6326 & .2860 & 21.30 & .6234 & .2710 & 26.62 & .7973 & .2259 & 18.07 & .4926 & .4230 \\
			& \ding{51} & & \ding{51} & 24.86 & .6598 & .3181 & 21.29 & .6094 & .3065 & \cellcolor{third!35}21.40 & \cellcolor{third!35}.6359 & \cellcolor{third!35}.2584 & \cellcolor{second!35}28.58 & .8332 & .1910 & 17.78 & .4869 & .4352 \\
			& \ding{51} & \ding{51} & & \cellcolor{best!35}25.01 & \cellcolor{second!35}.6691 & \cellcolor{best!35}.3007 & \cellcolor{third!35}23.19 & \cellcolor{second!35}.6706 & \cellcolor{third!35}.2481 & 21.12 & .6212 & .2662 & 28.23 & \cellcolor{third!35}.8410 & \cellcolor{second!35}.1798 & \cellcolor{best!35}19.88 & \cellcolor{best!35}.5550 & \cellcolor{best!35}.3606 \\
			& \ding{51} & \ding{51} & \ding{51} & 24.80 & .6643 & .3200 & \cellcolor{best!35}23.32 & \cellcolor{best!35}.6789 & \cellcolor{second!35}.2455 & 21.27 & .6234 & .2662 & \cellcolor{best!35}28.91 & \cellcolor{best!35}.8525 & \cellcolor{third!35}.1902 & \cellcolor{third!35}19.69 & \cellcolor{third!35}.5440 & \cellcolor{second!35}.3672 \\
			\midrule
			\multirow{7}{*}{\textcolor[rgb]{0.25, 0.0, 1.0}{DN-kernel}} & & & & \cellcolor{second!35}24.47 & \cellcolor{second!35}.6473 & \cellcolor{best!35}.3342 & \cellcolor{second!35}23.63 & \cellcolor{second!35}.7204 & \cellcolor{best!35}.2108 & \cellcolor{third!35}21.45 & .6285 & \cellcolor{third!35}.2625 & 23.70 & .7604 & .2588 & 17.74 & .4790 & .4296 \\
			& \ding{51} & & & 23.02 & .5757 & .3964 & 22.96 & .7128 & \cellcolor{third!35}.2230 & \cellcolor{best!35}21.72 & \cellcolor{second!35}.6389 & \cellcolor{second!35}.2595 & \cellcolor{third!35}26.90 & .8004 & .2354 & 17.21 & .4774 & .4346 \\
			& & \ding{51} & & \cellcolor{third!35}24.15 & \cellcolor{third!35}.6343 & .3465 & 22.59 & .6894 & .2365 & 21.30 & .6185 & .2678 & 26.41 & .7947 & \cellcolor{third!35}.2300 & \cellcolor{third!35}19.38 & \cellcolor{third!35}.5272 & \cellcolor{third!35}.3908 \\
			& & & \ding{51} & 23.68 & .6175 & .3572 & 22.90 & .7035 & .2255 & 20.95 & .6172 & .2716 & \cellcolor{second!35}27.18 & \cellcolor{third!35}.8083 & 2336 & 18.53 & .5030 & .4084 \\
			& \ding{51} & & \ding{51} & \cellcolor{best!35}24.60 & \cellcolor{best!35}.6530 & \cellcolor{second!35}.3439 & 22.96 & .7132 & .2296 & 21.43 & \cellcolor{third!35}.6360 & .2642 & 26.13 & \cellcolor{second!35}.8178 & \cellcolor{second!35}.2159 & 18.07 & .4784 & .4420 \\
			& \ding{51} & \ding{51} & & 23.93 & .6332 & \cellcolor{third!35}.3464 & \cellcolor{best!35}23.89 & \cellcolor{best!35}.7342 & \cellcolor{second!35}.2188 & 21.11 & .6247 & .2747 & 26.88 & .8074 & .2403 & \cellcolor{second!35}19.53 & \cellcolor{second!35}.5383 & \cellcolor{second!35}.3855 \\
			& \ding{51} & \ding{51} & \ding{51} & 23.76  & .6184 & .3607 & \cellcolor{third!35}23.39 & \cellcolor{third!35}.7155 & .2329 & \cellcolor{second!35}21.66 & \cellcolor{best!35}.6405 & \cellcolor{best!35}.2585 & \cellcolor{best!35}27.84 & \cellcolor{best!35}.8388 & \cellcolor{best!35}.2068 & \cellcolor{best!35}19.98 & \cellcolor{best!35}.5553 & \cellcolor{best!35}.3644 \\
			\midrule
			& \multicolumn{3}{c}{Real Scene}& \multicolumn{3}{c}{Girl} & \multicolumn{3}{c}{Heron} & \multicolumn{3}{c}{Parterre} & \multicolumn{3}{c}{Puppet} & \multicolumn{3}{c}{Stair} & \multicolumn{3}{c}{Average} \\
			& \textit{SS} & \textit{MGS} & \textit{PD} & PSNR($\uparrow$) & SSIM($\uparrow$) & LPIPS($\downarrow$) & PSNR($\uparrow$) & SSIM($\uparrow$) & LPIPS($\downarrow$) & PSNR($\uparrow$) & SSIM($\uparrow$) & LPIPS($\downarrow$) & PSNR($\uparrow$) & SSIM($\uparrow$) & LPIPS($\downarrow$) & PSNR($\uparrow$) & SSIM($\uparrow$) & LPIPS($\downarrow$) & PSNR($\uparrow$) & SSIM($\uparrow$) & LPIPS($\downarrow$) \\
			\midrule
			\multirow{7}{*}{\textcolor[rgb]{0.13, 0.55, 0.13}{DP-kernel}}& & & & 18.16 & .6622 & .3020 & \cellcolor{third!35}19.58 & \cellcolor{third!35}.4991 & .2905 & 22.36 & .5879 & .3350 & \cellcolor{best!35}20.65 & \cellcolor{best!35}.5868 & \cellcolor{best!35}.2988 & 23.40 & \cellcolor{second!35}.5624 & \cellcolor{best!35}.2986 & 21.68 & .6137 & .2992 \\
			& \ding{51} & & & \cellcolor{third!35}18.79 & .6771 & .2890 & 19.56 & .4967 & .2903 & \cellcolor{third!35}22.53 & \cellcolor{third!35}.5972 & \cellcolor{third!35}.3318 & 20.41 & .5765 & .3101 & \cellcolor{best!35}23.96 & \cellcolor{best!35}.5723 & \cellcolor{second!35}.2997 & \cellcolor{third!35}22.10 & \cellcolor{third!35}.6214 & .2951 \\
			& & \ding{51} & & 18.77 & \cellcolor{third!35}.6833 & \cellcolor{second!35}.2703 & \cellcolor{third!35}19.58 & \cellcolor{second!35}.4998 & \cellcolor{second!35}.2856 & 22.27 & .5801 & .3440 & \cellcolor{third!35}20.48 & \cellcolor{third!35}.5790 & \cellcolor{third!35}.3043 & 23.23 & .5468 & .3149 & 22.07 & .6212 & \cellcolor{second!35}.2889 \\
			& & & \ding{51} & 18.50 & .6650 & .2956 & \cellcolor{best!35}19.59 & .4895 & .3003 & 22.45 & .5926 & .3344 & 20.29 & .5687 & .3230 & 23.33 & .5545 & .3120 & 21.73 & .6082 & .3082 \\
			& \ding{51} & & \ding{51} &  18.36 & .6628 & .2989 & 19.57 & .4920 & .3006 & \cellcolor{best!35}22.69 & \cellcolor{best!35}.6015 & \cellcolor{best!35}.3296 & 20.38 & .5774 & .3211 & \cellcolor{third!35}23.47 & .5462 & .3244 & 21.84 & .6105 & .3084\\
			& \ding{51} & \ding{51} & & \cellcolor{best!35}19.20 & \cellcolor{second!35}.6872 & \cellcolor{best!35}.2695 & 19.45 & .4947 & \cellcolor{third!35}.2871 & 22.33 & .5833 & .3438 & 20.46 & \cellcolor{second!35}.5800 & \cellcolor{second!35}.3035 & 23.34 & \cellcolor{third!35}.5550 & \cellcolor{third!35}.3082 & \cellcolor{second!35}22.22 & \cellcolor{second!35}.6257 & \cellcolor{best!35}.2868 \\
			& \ding{51} & \ding{51} & \ding{51} & \cellcolor{second!35}18.96 & \cellcolor{best!35}.6887 & \cellcolor{third!35}.2744 & \cellcolor{best!35}19.59 & \cellcolor{best!35}.5051 & \cellcolor{best!35}.2808 & \cellcolor{second!35}22.57 & \cellcolor{second!35}.5994 & \cellcolor{second!35}.3317 & \cellcolor{second!35}20.60 & .5756 & .3124 & \cellcolor{second!35}23.48 & .5509 & .3189 & \cellcolor{best!35}22.32 & \cellcolor{best!35}.6283 & \cellcolor{third!35}.2907 \\
			\midrule
			\multirow{7}{*}{\textcolor[rgb]{0.25, 0.0, 1.0}{DN-kernel}}& & & & 18.09 & .6582 & .3055 & 19.40 & .4709 & .3166 & 21.92 & .5557 & .3872 & \cellcolor{third!35}20.58 & \cellcolor{second!35}.5834 & .3226 & 22.60 & .4993 & .3349 & 21.36 & .6003 & .3163 \\
			& \ding{51} & & & 18.19 & .6615 & .3009 & 19.42 & .4879 & \cellcolor{second!35}.3052 & \cellcolor{third!35}22.18 & \cellcolor{third!35}.5758 & \cellcolor{third!35}.3733 & 20.49 & \cellcolor{third!35}.5823 & \cellcolor{third!35}.3167 & \cellcolor{second!35}23.18 & \cellcolor{third!35}.5312 & \cellcolor{third!35}.3330 & 21.53 & .6044 & .3178 \\
			& & \ding{51} & & \cellcolor{best!35}18.93 & \cellcolor{second!35}.6856 & \cellcolor{best!35}.2748 & 19.58 & .4904 & .3122 & 21.91 & .5570 & .3853 & 20.44 & .5810 & .3232 & 22.90 & .5231 & \cellcolor{best!35}.3294 & \cellcolor{third!35}21.76 & .6101 & \cellcolor{second!35}.3097 \\
			& & & \ding{51} & 18.18 & .6538 & .3005 & \cellcolor{second!35}19.65 & \cellcolor{second!35}.4962 & \cellcolor{third!35}.3083 & \cellcolor{best!35}22.34 & \cellcolor{best!35}.5834 & \cellcolor{best!35}.3608 & 20.41 & .5785 & .3305 & 23.08 & .5257 & .3458 & 21.69 & .6087 & .3142 \\
			& \ding{51} & & \ding{51} & \cellcolor{third!35}18.74 & \cellcolor{third!35}.6782 & \cellcolor{third!35}.2855 & \cellcolor{best!35}19.74 & \cellcolor{third!35}.4948 & .3153 & \cellcolor{second!35}22.25 & \cellcolor{second!35}.5796 & \cellcolor{second!35}.3658 & 20.45 & .5771 & .3183 & 22.63 & .5009 & .3637 & 21.70 & \cellcolor{third!35}.6129 & .3144 \\
			& \ding{51} & \ding{51} & & 18.53 & .6646 & .2923 & 19.45 & .4859 & .3182 & 21.62 & .5510 & .3909 & \cellcolor{second!35}20.71 & .5819 & \cellcolor{best!35}.3126 & \cellcolor{second!35}23.18 & \cellcolor{second!35}.5332 & \cellcolor{second!35}.3325 & \cellcolor{second!35}21.88 & \cellcolor{second!35}.6154 & \cellcolor{third!35}.3112 \\
			& \ding{51} & \ding{51} & \ding{51} & \cellcolor{second!35}18.86 & \cellcolor{best!35}.6860 & \cellcolor{second!35}.2769 & \cellcolor{third!35}19.63 & \cellcolor{best!35}.5011 & \cellcolor{best!35}.2992 & 22.15 & .5703 & .3787 & \cellcolor{best!35}20.81 & \cellcolor{best!35}.5882 & \cellcolor{second!35}.3149 & \cellcolor{best!35}23.42 & \cellcolor{best!35}.5340 & .3374 & \cellcolor{best!35}22.15 & \cellcolor{best!35}.6248 & \cellcolor{best!35}.3030 \\
			\midrule
      \end{tabular}
      }
   \label{tab:appendix_ablation_quant_results_6view}
   \end{center}
\end{table*}

\begin{table*}[t]
   \Large
   \begin{center}
   \caption{Quantitative results of the RegNeRF~\cite{niemeyer2022regnerf} with and without the blur kernel from \textbf{2-view} to \textbf{10-view} settings. For comparison, the blur kernel of the DP-NeRF~\cite{lee2023dpnerf} is employed, which is denoted as \textcolor[rgb]{0.13, 0.55, 0.13}{DP-kernel}. Color shading represents the \colorbox{best!35}{better} result.}
      \resizebox{2\columnwidth}{!}{
		\centering
		\setlength{\tabcolsep}{1pt}
\begin{tabular}{C{3cm}|C{3cm}||C{3cm}|C{3cm}|C{3cm}|C{3cm}|C{3cm}|C{3cm}|C{3cm}|C{3cm}|C{3cm}|C{3cm}|}
			\midrule
			\multicolumn{2}{c}{\textbf{Decoration} Scene} & \multicolumn{9}{c}{\# of views} \\
			\midrule
			\textbf{PSNR}($\uparrow$) & Blur Kernel & 2 & 3 & 4 & 5 & 6 & 7 & 8 & 9 & 10 \\
			\midrule
			RegNeRF~\cite{niemeyer2022regnerf} & $\times$ & \cellcolor{best!35}11.20 & \cellcolor{best!35}11.28 & \cellcolor{best!35}11.43 & \cellcolor{best!35}12.11 & \cellcolor{best!35}12.45 & \cellcolor{best!35}14.61 & \cellcolor{best!35}16.69 & 18.41 & 20.21 \\
			RegNeRF~\cite{niemeyer2022regnerf} & \textcolor[rgb]{0.13, 0.55, 0.13}{DP-kernel} & 11.15 & 10.84 & 11.07 & 11.54 & 11.43 & 13.50 & 16.16 & \cellcolor{best!35}18.56 & \cellcolor{best!35}21.75\\
			\midrule\midrule
			\textbf{SSIM}($\uparrow$) & Blur Kernel & 2 & 3 & 4 & 5 & 6 & 7 & 8 & 9 & 10  \\
			\midrule
			RegNeRF~\cite{niemeyer2022regnerf} & $\times$ & \cellcolor{best!35}.1064 & \cellcolor{best!35}.1262 & \cellcolor{best!35}.1139 & \cellcolor{best!35}.1589 & \cellcolor{best!35}.1569 & \cellcolor{best!35}.2876 & \cellcolor{best!35}.3956 & .4880 & .5508 \\
			RegNeRF~\cite{niemeyer2022regnerf} & \textcolor[rgb]{0.13, 0.55, 0.13}{DP-kernel} & .0927 & .1115 & .1058 & .1228 & .1247 & .2034 & .3645 & \cellcolor{best!35}.4986 & \cellcolor{best!35}.6526\\
			\midrule
			\textbf{LPIPS}($\downarrow$) & Blur Kernel & 2 & 3 & 4 & 5 & 6 & 7 & 8 & 9 & 10  \\
			\midrule
			RegNeRF~\cite{niemeyer2022regnerf} & $\times$ & \cellcolor{best!35}.7475 & \cellcolor{best!35}.7059 & \cellcolor{best!35}.6995 & .7078 & .7061 & \cellcolor{best!35}.6258 & .5588 & .4818 & .4322 \\
			RegNeRF~\cite{niemeyer2022regnerf} & \textcolor[rgb]{0.13, 0.55, 0.13}{DP-kernel} & .7520 & .7191 & .7089 & \cellcolor{best!35}.6992 & \cellcolor{best!35}.6932 & .6547 & \cellcolor{best!35}.5277 & \cellcolor{best!35}.4178 & \cellcolor{best!35}.2804\\
			\midrule
      \end{tabular}
      }
   \label{tab:appendix_regnerf_ablation_2view_to_10_view_decoration}
   \end{center}
\end{table*}

\begin{table*}[t]
   \Large
   \begin{center}
   \caption{Entire experimental results for the complex optimization issue in both synthetic and real scenes from 2-view settings. Each color shading represents the \colorbox{best!35}{best} and \colorbox{second!35}{second} results for each experimental setting, respectively.}
      \vspace{-0.2cm}
      \resizebox{2\columnwidth}{!}{
		\centering
		\setlength{\tabcolsep}{1pt}
         \begin{tabular}{C{4cm}||C{3cm}||ccc|ccc|ccc|ccc|ccc||ccc}
			\midrule
			\textbf{Synthetic Scene} & \multirow{2}{*}{Blur Kernel} & \multicolumn{3}{c}{Factory} & \multicolumn{3}{c}{Cozyroom} & \multicolumn{3}{c}{Pool} & \multicolumn{3}{c}{Tanabata} & \multicolumn{3}{c}{Trolley} & \multicolumn{3}{c}{Average} \\
			Model & & PSNR($\uparrow$) & SSIM($\uparrow$) & LPIPS($\downarrow$) & PSNR($\uparrow$) & SSIM($\uparrow$) & LPIPS($\downarrow$) & PSNR($\uparrow$) & SSIM($\uparrow$) & LPIPS($\downarrow$) & PSNR($\uparrow$) & SSIM($\uparrow$) & LPIPS($\downarrow$) & PSNR($\uparrow$) & SSIM($\uparrow$) & LPIPS($\downarrow$) & PSNR($\uparrow$) & SSIM($\uparrow$) & LPIPS($\downarrow$) \\
			\midrule
			RegNeRF~\cite{niemeyer2022regnerf} & $\times$ &  \cellcolor{best!35}14.57 &  \cellcolor{best!35}.2523 & .6680 & 17.13 &  .4616 &  .3808 & \cellcolor{second!35}14.23 & \cellcolor{second!35}.1473 & .7298 &  \cellcolor{best!35}13.01 &  \cellcolor{best!35}.2408 & \cellcolor{second!35}.5956 &  \cellcolor{best!35}14.04 &  \cellcolor{best!35}.3227 &  \cellcolor{best!35}.5602 & \cellcolor{second!35}14.60 & \cellcolor{second!35}.2849 & \cellcolor{second!35}.5869 \\
			RegNeRF~\cite{niemeyer2022regnerf} & \textcolor[rgb]{0.13, 0.55, 0.13}{DP-kernel} &  \cellcolor{second!35}14.20 &  \cellcolor{second!35}.2214 &   \cellcolor{best!35}.6412 &  \cellcolor{second!35}18.88 & \cellcolor{second!35}.5136 & \cellcolor{second!35}.3476 &  13.03 &  .1003 &  \cellcolor{second!35}.6845 &  9.21 &  .0815 & .6920 &  10.86 &  .1644 &  .6657 &  13.24 &  .2162 &  .6062 \\
\end{tabular}
      }
   \label{tab:appendix_regnerf_ablation_2view}
   \end{center}
\end{table*}

\begin{table*}[t]
   \Large
   \begin{center}
   \caption{Entire experimental results for the complex optimization issue in both synthetic and real scenes from 4-view settings. Each color shading represents the \colorbox{best!35}{best} and \colorbox{second!35}{second} results for each experimental setting, respectively.}
      \resizebox{2\columnwidth}{!}{
		\centering
		\setlength{\tabcolsep}{1pt}
         \begin{tabular}{C{4cm}||C{3cm}||ccc|ccc|ccc|ccc|ccc||ccc}
			\midrule
			\textbf{Synthetic Scene} & \multirow{2}{*}{Blur Kernel} & \multicolumn{3}{c}{Factory} & \multicolumn{3}{c}{Cozyroom} & \multicolumn{3}{c}{Pool} & \multicolumn{3}{c}{Tanabata} & \multicolumn{3}{c}{Trolley} & \multicolumn{3}{c}{Average} \\
			Model & & PSNR($\uparrow$) & SSIM($\uparrow$) & LPIPS($\downarrow$) & PSNR($\uparrow$) & SSIM($\uparrow$) & LPIPS($\downarrow$) & PSNR($\uparrow$) & SSIM($\uparrow$) & LPIPS($\downarrow$) & PSNR($\uparrow$) & SSIM($\uparrow$) & LPIPS($\downarrow$) & PSNR($\uparrow$) & SSIM($\uparrow$) & LPIPS($\downarrow$) & PSNR($\uparrow$) & SSIM($\uparrow$) & LPIPS($\downarrow$) \\
			\midrule
			RegNeRF~\cite{niemeyer2022regnerf} & $\times$ & 16.32 & .3441 & .5876 &  \cellcolor{second!35}23.25 & \cellcolor{second!35}.7155 & .2663 & \cellcolor{second!35}16.21 & \cellcolor{second!35}.2040 & \cellcolor{second!35}.6451 & \cellcolor{second!35}17.35 & \cellcolor{second!35}.4915 & \cellcolor{second!35}.4509 & \cellcolor{second!35}18.81 & \cellcolor{second!35}.5447 & \cellcolor{second!35}.4020 & \cellcolor{second!35}18.39 & \cellcolor{second!35}.4600 & \cellcolor{second!35}.4704 \\
			RegNeRF~\cite{niemeyer2022regnerf} & \textcolor[rgb]{0.13, 0.55, 0.13}{DP-kernel} &  \cellcolor{second!35}16.71 &  \cellcolor{second!35}.3497 &  \cellcolor{second!35}.4683 &  21.37 &  .6471 &  \cellcolor{second!35}.1802 &  15.17 &  .1740 &  .6666 &  10.27 &  .1165 & .6911 & 18.66 & .5412 &  .4067 &  16.44 &  .3657 &  .4826 \\
\end{tabular}
      }
   \label{tab:appendix_regnerf_ablation_4view}
   \end{center}
\end{table*}

\begin{table*}[t]
   \Large
   \begin{center}
   \caption{Entire experimental results for the complex optimization issue in both synthetic and real scenes from 6-view settings. Each color shading represents the \colorbox{best!35}{best} and \colorbox{second!35}{second} results for each experimental setting, respectively.}
      \resizebox{2\columnwidth}{!}{
		\centering
		\setlength{\tabcolsep}{1pt}
         \begin{tabular}{C{4cm}||C{3cm}||ccc|ccc|ccc|ccc|ccc||ccc}
			\midrule
			\textbf{Synthetic Scene} & \multirow{2}{*}{Blur Kernel} & \multicolumn{3}{c}{Factory} & \multicolumn{3}{c}{Cozyroom} & \multicolumn{3}{c}{Pool} & \multicolumn{3}{c}{Tanabata} & \multicolumn{3}{c}{Trolley} & \multicolumn{3}{c}{Average} \\
			Model & & PSNR($\uparrow$) & SSIM($\uparrow$) & LPIPS($\downarrow$) & PSNR($\uparrow$) & SSIM($\uparrow$) & LPIPS($\downarrow$) & PSNR($\uparrow$) & SSIM($\uparrow$) & LPIPS($\downarrow$) & PSNR($\uparrow$) & SSIM($\uparrow$) & LPIPS($\downarrow$) & PSNR($\uparrow$) & SSIM($\uparrow$) & LPIPS($\downarrow$) & PSNR($\uparrow$) & SSIM($\uparrow$) & LPIPS($\downarrow$) \\
			\midrule
			RegNeRF~\cite{niemeyer2022regnerf} & $\times$ & 17.04 & .3690 & .5729 & 23.40 & .7205 & .2649 & \cellcolor{second!35}18.05 & \cellcolor{second!35}.3610 & \cellcolor{second!35}.4870 & 20.33 & .5980 & .3745 & \cellcolor{second!35}19.63 & \cellcolor{second!35}.5760 & \cellcolor{second!35}.3833 & 19.69 & .5249 & .4165 \\
			RegNeRF~\cite{niemeyer2022regnerf} & \textcolor[rgb]{0.13, 0.55, 0.13}{DP-kernel} & \cellcolor{second!35}21.03 & \cellcolor{best!35}.6273 & \cellcolor{best!35}.3076 & \cellcolor{best!35}27.73 & \cellcolor{best!35}.8455 & \cellcolor{best!35}.1238 & 18.04 & .3173 & .4987 & \cellcolor{second!35}21.84 & \cellcolor{second!35}.7246 & \cellcolor{second!35}.2022 & 19.36 & .5661 & .3908 & \cellcolor{second!35}21.60 & \cellcolor{second!35}.6162 & \cellcolor{second!35}.3046 \\
\end{tabular}
      }
   \label{tab:appendix_regnerf_ablation_6view}
   \end{center}
\end{table*}

\begin{table*}[t]
   \Large
   \begin{center}
   \caption{Quantitative comparison between naive gradient scaling of~\cite{philip2023floatersnomore} and our MGS for entire scenes of synthetic and real scenes from \textbf{2-view} settings. The experimental results are presented according to the two types of kernel we utilize in this paper. Color shading represents the \colorbox{best!35}{better} result.}
      \resizebox{2\columnwidth}{!}{
		\centering
		\setlength{\tabcolsep}{1pt}
         \begin{tabular}{C{3cm}||C{5.2cm}||ccc|ccc|ccc|ccc|ccc||ccc}
			\midrule
			\textbf{Synthetic Scene} & \multirow{2}{*}{Gradient Scaling} & \multicolumn{3}{c}{Factory} & \multicolumn{3}{c}{Cozyroom} & \multicolumn{3}{c}{Pool} & \multicolumn{3}{c}{Tanabata} & \multicolumn{3}{c}{Trolley} & \multicolumn{3}{c}{Average} \\
			Kernel Type & & PSNR($\uparrow$) & SSIM($\uparrow$) & LPIPS($\downarrow$) & PSNR($\uparrow$) & SSIM($\uparrow$) & LPIPS($\downarrow$) & PSNR($\uparrow$) & SSIM($\uparrow$) & LPIPS($\downarrow$) & PSNR($\uparrow$) & SSIM($\uparrow$) & LPIPS($\downarrow$) & PSNR($\uparrow$) & SSIM($\uparrow$) & LPIPS($\downarrow$) & PSNR($\uparrow$) & SSIM($\uparrow$) & LPIPS($\downarrow$) \\
			\midrule
			\textcolor[rgb]{0.25, 0.0, 1.0}{DN-kernel} & $+$~Naive~\cite{philip2023floatersnomore} & 11.93 & .1340 & .6835 & 17.82 & .4573 & .4827 & 18.89 & .3131 & .4237 & 11.45 & .1603 & .6107 & 11.12 & .1769 & \cellcolor{best!35}.6207 & 14.70 & .2483 & .5643 \\
			\textcolor[rgb]{0.25, 0.0, 1.0}{DN-kernel} & $+$~MGS & \cellcolor{best!35}14.20 & \cellcolor{best!35}.2107 & \cellcolor{best!35}.6686 & \cellcolor{best!35}21.02 & \cellcolor{best!35}.5999 & \cellcolor{best!35}.3326 & \cellcolor{best!35}20.30 & \cellcolor{best!35}.3467 & \cellcolor{best!35}.4017 & \cellcolor{best!35}11.55 & \cellcolor{best!35}.1804 & \cellcolor{best!35}.5903 & \cellcolor{best!35}11.48 & \cellcolor{best!35}.1883 & .6351 & \cellcolor{best!35}15.12 & \cellcolor{best!35}.3025 & \cellcolor{best!35}.5243 \\
			\midrule
			\textcolor[rgb]{0.13, 0.55, 0.13}{DP-kernel} & $+$~Naive~\cite{philip2023floatersnomore} & 11.98 & .1378 & .6799 & 16.54 & .3988 & .5360 & 19.08 & .3135 & .4275 & 11.30 & .1653 & \cellcolor{best!35}.5992 & 11.14 & .1731 & .6291 & 14.01 & .2377 & .5743 \\
\end{tabular}
      }
   \label{tab:appendix_gradient_scaling_comparison_2view}
   \end{center}
\end{table*}

\begin{table*}[t]
   \Large
   \begin{center}
   \caption{Quantitative comparison between naive gradient scaling of~\cite{philip2023floatersnomore} and our MGS for entire scenes of synthetic and real scenes from \textbf{4-view} settings. The experimental results are presented according to the two types of kernel we utilize in this paper. Color shading represents the \colorbox{best!35}{better} result.}
      \resizebox{2\columnwidth}{!}{
		\centering
		\setlength{\tabcolsep}{1pt}
         \begin{tabular}{C{3cm}||C{5.2cm}||ccc|ccc|ccc|ccc|ccc||ccc}
         	\midrule
			\textbf{Synthetic Scene} & \multirow{2}{*}{Gradient Scaling} & \multicolumn{3}{c}{Factory} & \multicolumn{3}{c}{Cozyroom} & \multicolumn{3}{c}{Pool} & \multicolumn{3}{c}{Tanabata} & \multicolumn{3}{c}{Trolley} & \multicolumn{3}{c}{Average} \\
			Kernel Type & & PSNR($\uparrow$) & SSIM($\uparrow$) & LPIPS($\downarrow$) & PSNR($\uparrow$) & SSIM($\uparrow$) & LPIPS($\downarrow$) & PSNR($\uparrow$) & SSIM($\uparrow$) & LPIPS($\downarrow$) & PSNR($\uparrow$) & SSIM($\uparrow$) & LPIPS($\downarrow$) & PSNR($\uparrow$) & SSIM($\uparrow$) & LPIPS($\downarrow$) & PSNR($\uparrow$) & SSIM($\uparrow$) & LPIPS($\downarrow$) \\
			\midrule
			\textcolor[rgb]{0.25, 0.0, 1.0}{DN-kernel} & $+$~Naive~\cite{philip2023floatersnomore} & 12.80 & .1949 & .6631 & 23.41 & .7202 & .2517 & 23.54 & .5069 & \cellcolor{best!35}.2508 & 14.22 & .3275 & .5065 & 15.41 & .4109 & .4487 & 18.69 & .4321 & .4242 \\
			\textcolor[rgb]{0.25, 0.0, 1.0}{DN-kernel} & $+$~MGS & \cellcolor{best!35}16.33 & \cellcolor{best!35}.3387 & \cellcolor{best!35}.5636 & \cellcolor{best!35}25.55 & \cellcolor{best!35}.7813 & \cellcolor{best!35}.1874 & \cellcolor{best!35}24.28 & \cellcolor{best!35}.5544 & .2649 & \cellcolor{best!35}16.60 & \cellcolor{best!35}.4635 & \cellcolor{best!35}.3781 & \cellcolor{best!35}17.96 & \cellcolor{best!35}.5095 & \cellcolor{best!35}.3870 & \cellcolor{best!35}19.44 & \cellcolor{best!35}.5295 & \cellcolor{best!35}.3562 \\
			\midrule
			\textcolor[rgb]{0.13, 0.55, 0.13}{DP-kernel} & $+$~Naive~\cite{philip2023floatersnomore} & 11.89 & .1392 & .6831 & 22.26 & .6676 & .2354 & 22.36 & \cellcolor{best!35}.4721 & \cellcolor{best!35}.2616 & 14.14 & .3024 & .4953 & 15.62 & .3850 & .4351 & 17.25 & .3933 & .4221 \\
\end{tabular}
      }
   \label{tab:appendix_gradient_scaling_comparison_4view}
   \end{center}
\end{table*}

\begin{table*}[t]
   \Large
   \begin{center}
   \caption{Quantitative comparison between naive gradient scaling of~\cite{philip2023floatersnomore} and our MGS for entire scenes of synthetic and real scenes from \textbf{6-view} settings. The experimental results are presented according to the two types of kernel we utilize in this paper. Color shading represents the \colorbox{best!35}{better} result.}
      \resizebox{2\columnwidth}{!}{
		\centering
		\setlength{\tabcolsep}{1pt}
         \begin{tabular}{C{3cm}||C{5.2cm}||ccc|ccc|ccc|ccc|ccc||ccc}
         	\midrule
			\textbf{Synthetic Scene} & \multirow{2}{*}{Gradient Scaling} & \multicolumn{3}{c}{Factory} & \multicolumn{3}{c}{Cozyroom} & \multicolumn{3}{c}{Pool} & \multicolumn{3}{c}{Tanabata} & \multicolumn{3}{c}{Trolley} & \multicolumn{3}{c}{Average} \\
			Kernel Type & & PSNR($\uparrow$) & SSIM($\uparrow$) & LPIPS($\downarrow$) & PSNR($\uparrow$) & SSIM($\uparrow$) & LPIPS($\downarrow$) & PSNR($\uparrow$) & SSIM($\uparrow$) & LPIPS($\downarrow$) & PSNR($\uparrow$) & SSIM($\uparrow$) & LPIPS($\downarrow$) & PSNR($\uparrow$) & SSIM($\uparrow$) & LPIPS($\downarrow$) & PSNR($\uparrow$) & SSIM($\uparrow$) & LPIPS($\downarrow$) \\
			\midrule
			\textcolor[rgb]{0.25, 0.0, 1.0}{DN-kernel} & $+$~Naive~\cite{philip2023floatersnomore} & 14.70 & .2560 & .6091 & 26.13 & .8080 & .1658 & 27.24 & .7167 & .1810 & 20.29 & .6379 & .2816 & 18.48 & .5824 & .3208 & 22.00 & .6002 & .3117 \\
			\textcolor[rgb]{0.25, 0.0, 1.0}{DN-kernel} & $+$~MGS & \cellcolor{best!35}18.06 & \cellcolor{best!35}.4849 & \cellcolor{best!35}.4104 & \cellcolor{best!35}26.54 & \cellcolor{best!35}.8159 & \cellcolor{best!35}.1486 & \cellcolor{best!35}28.09 & \cellcolor{best!35}.7392 & \cellcolor{best!35}.1725 & \cellcolor{best!35}20.97 & \cellcolor{best!35}.6866 & \cellcolor{best!35}.2394 & \cellcolor{best!35}21.71 & \cellcolor{best!35}.6895 & \cellcolor{best!35}.2320 & \cellcolor{best!35}22.40 & \cellcolor{best!35}.6832 & \cellcolor{best!35}.2406 \\
			\midrule
			\textcolor[rgb]{0.13, 0.55, 0.13}{DP-kernel} & $+$~Naive~\cite{philip2023floatersnomore} & 14.47 & .2649 & .5744 & 26.71 & .8233 & .1456 & 27.90 & .7330 & .1598 & 20.04 & .6253 & .2528 & 18.63 & .6020 & .2865 & 21.55 & .6097 & .2838 \\
\end{tabular}
      }
   \label{tab:appendix_gradient_scaling_comparison_6view}
   \end{center}
\end{table*}

\begin{table*}[t]
   \Large
   \begin{center}
   \caption{Quantitative comparison results for entire scenes from a 2-view setting that include MGS experiments on the \textbf{world coordinate system}. Color shading represents the \colorbox{best!35}{best} and \colorbox{second!35}{second} result.}
      \resizebox{2\columnwidth}{!}{
		\centering
		\setlength{\tabcolsep}{1pt}
         \begin{tabular}{C{3cm}||C{5.2cm}||ccc|ccc|ccc|ccc|ccc||ccc}
         	\midrule
			\textbf{Synthetic Scene} & \multirow{2}{*}{Gradient Scaling} & \multicolumn{3}{c}{Factory} & \multicolumn{3}{c}{Cozyroom} & \multicolumn{3}{c}{Pool} & \multicolumn{3}{c}{Tanabata} & \multicolumn{3}{c}{Trolley} & \multicolumn{3}{c}{Average} \\
			Kernel Type & & PSNR($\uparrow$) & SSIM($\uparrow$) & LPIPS($\downarrow$) & PSNR($\uparrow$) & SSIM($\uparrow$) & LPIPS($\downarrow$) & PSNR($\uparrow$) & SSIM($\uparrow$) & LPIPS($\downarrow$) & PSNR($\uparrow$) & SSIM($\uparrow$) & LPIPS($\downarrow$) & PSNR($\uparrow$) & SSIM($\uparrow$) & LPIPS($\downarrow$) & PSNR($\uparrow$) & SSIM($\uparrow$) & LPIPS($\downarrow$) \\
			\midrule
			\textcolor[rgb]{0.13, 0.55, 0.13}{DP-kernel} & $\times$ & \cellcolor{second!35}15.17 & \cellcolor{best!35}.2710 & .6140 & \cellcolor{second!35}13.48 & \cellcolor{second!35}.2441 & .6758 & 15.24 & \cellcolor{second!35}.1592 & .8082 & \cellcolor{best!35}15.00 & \cellcolor{second!35}.3469 & \cellcolor{second!35}.4822 & \cellcolor{best!35}13.43 & \cellcolor{best!35}.2859 & .5489 & \cellcolor{second!35}14.46 & \cellcolor{best!35}.2614 & .6258 \\
			\textcolor[rgb]{0.13, 0.55, 0.13}{DP-kernel} & $+$~Naive~\cite{philip2023floatersnomore} &  15.10 & .2599 & \cellcolor{second!35}.6074 & \cellcolor{best!35}13.70 & \cellcolor{best!35}.2598 & \cellcolor{best!35}.6662 & \cellcolor{second!35}15.41 & \cellcolor{best!35}.1729 & \cellcolor{second!35}.7816 & 14.85 & .3403 & .4823 & 13.00 & .2566 & .5643 & 14.41 & \cellcolor{second!35}.2579 & \cellcolor{second!35}.6204\\
\end{tabular}
      }
   \label{tab:appendix_mgs_world_exps_2view}
   \end{center}
\end{table*}

\begin{table*}[t]
   \Large
   \begin{center}
   \caption{Quantitative comparison results for entire scenes from a 4-view setting that include MGS experiments on the \textbf{world coordinate system}. Color shading represents the \colorbox{best!35}{best} and \colorbox{second!35}{second} result.}
      \resizebox{2\columnwidth}{!}{
		\centering
		\setlength{\tabcolsep}{1pt}
         \begin{tabular}{C{3cm}||C{5.2cm}||ccc|ccc|ccc|ccc|ccc||ccc}
         	\midrule
			\textbf{Synthetic Scene} & \multirow{2}{*}{Gradient Scaling} & \multicolumn{3}{c}{Factory} & \multicolumn{3}{c}{Cozyroom} & \multicolumn{3}{c}{Pool} & \multicolumn{3}{c}{Tanabata} & \multicolumn{3}{c}{Trolley} & \multicolumn{3}{c}{Average} \\
			Kernel Type & & PSNR($\uparrow$) & SSIM($\uparrow$) & LPIPS($\downarrow$) & PSNR($\uparrow$) & SSIM($\uparrow$) & LPIPS($\downarrow$) & PSNR($\uparrow$) & SSIM($\uparrow$) & LPIPS($\downarrow$) & PSNR($\uparrow$) & SSIM($\uparrow$) & LPIPS($\downarrow$) & PSNR($\uparrow$) & SSIM($\uparrow$) & LPIPS($\downarrow$) & PSNR($\uparrow$) & SSIM($\uparrow$) & LPIPS($\downarrow$) \\
			\midrule
			\textcolor[rgb]{0.13, 0.55, 0.13}{DP-kernel} & $\times$ & 16.61 & .3544 & \cellcolor{second!35}.5434 & \cellcolor{second!35}25.00 & \cellcolor{second!35}.7541 & \cellcolor{best!35}.1967 & \cellcolor{second!35}15.31 & \cellcolor{second!35}.1878 & .8088 & \cellcolor{best!35}17.57 & \cellcolor{best!35}.4957 & \cellcolor{best!35}.3911 & \cellcolor{second!35}17.81 & \cellcolor{second!35}.4852 & \cellcolor{best!35}.4055 & \cellcolor{second!35}18.46 & \cellcolor{best!35}.4554 & \cellcolor{second!35}.4691\\
			\textcolor[rgb]{0.13, 0.55, 0.13}{DP-kernel} & $+$~Naive~\cite{philip2023floatersnomore} & \cellcolor{second!35}16.70 & \cellcolor{second!35}.3570 & .5459 & \cellcolor{best!35}25.11 & \cellcolor{best!35}.7578 & \cellcolor{second!35}.2058 & 15.15 & .1751 & \cellcolor{second!35}.7878 & \cellcolor{second!35}17.50 & \cellcolor{second!35}.4898 & \cellcolor{second!35}.3982 & 17.54 & .4732 & .4153 & 18.40 & \cellcolor{second!35}.4506 & .4706\\
\end{tabular}
      }
   \label{tab:appendix_mgs_world_exps_4view}
   \end{center}
\end{table*}

\begin{table*}[t]
   \Large
   \begin{center}
   \caption{Quantitative comparison results for entire scenes from a 6-view setting that include MGS experiments on the \textbf{world coordinate system}. Color shading represents the \colorbox{best!35}{best} and \colorbox{second!35}{second} result.}
      \resizebox{2\columnwidth}{!}{
		\centering
		\setlength{\tabcolsep}{1pt}
         \begin{tabular}{C{3cm}||C{5.2cm}||ccc|ccc|ccc|ccc|ccc||ccc}
         	\midrule
			\textbf{Synthetic Scene} & \multirow{2}{*}{Gradient Scaling} & \multicolumn{3}{c}{Factory} & \multicolumn{3}{c}{Cozyroom} & \multicolumn{3}{c}{Pool} & \multicolumn{3}{c}{Tanabata} & \multicolumn{3}{c}{Trolley} & \multicolumn{3}{c}{Average} \\
			Kernel Type & & PSNR($\uparrow$) & SSIM($\uparrow$) & LPIPS($\downarrow$) & PSNR($\uparrow$) & SSIM($\uparrow$) & LPIPS($\downarrow$) & PSNR($\uparrow$) & SSIM($\uparrow$) & LPIPS($\downarrow$) & PSNR($\uparrow$) & SSIM($\uparrow$) & LPIPS($\downarrow$) & PSNR($\uparrow$) & SSIM($\uparrow$) & LPIPS($\downarrow$) & PSNR($\uparrow$) & SSIM($\uparrow$) & LPIPS($\downarrow$) \\
			\midrule
			\textcolor[rgb]{0.13, 0.55, 0.13}{DP-kernel} & $\times$ & \cellcolor{best!35}17.40 & \cellcolor{best!35}.3972 & \cellcolor{second!35}.4993 & \cellcolor{best!35}24.76 & \cellcolor{best!35}.7471 & \cellcolor{best!35}1878 & 16.10 & .2033 & .7592 & \cellcolor{second!35}18.35 & \cellcolor{second!35}.5266 & .3951 & 18.60 & .5370 & \cellcolor{second!35}.3628 & \cellcolor{second!35}19.04 & \cellcolor{second!35}.4822 & \cellcolor{second!35}.4408\\
			\textcolor[rgb]{0.13, 0.55, 0.13}{DP-kernel} & $+$~Naive~\cite{philip2023floatersnomore} & \cellcolor{second!35}17.36 & .3929 & .5039 & \cellcolor{second!35}24.27 & \cellcolor{second!35}.7370 & .2001 & \cellcolor{second!35}16.33 & \cellcolor{second!35}.2201 & \cellcolor{second!35}.7567 & \cellcolor{best!35}18.67 & \cellcolor{best!35}.5289 & \cellcolor{best!35}.3888 & \cellcolor{second!35}18.42 & \cellcolor{second!35}.5363 & \cellcolor{second!35}.3628 & 19.01 & \cellcolor{best!35}.4830 & .4425 \\
\end{tabular}
      }
   \label{tab:appendix_mgs_world_exps_6view}
   \end{center}
\end{table*}